\documentclass[]{bytedance_seed}

\usepackage[toc,page,header]{appendix}

\usepackage[T1]{fontenc}
\usepackage[utf8]{inputenc} 

\usepackage{graphicx}
\usepackage{wrapfig}
\usepackage[dvipsnames,table]{xcolor}

\usepackage{amsmath,amssymb}
\usepackage{float}
\usepackage{booktabs}
\usepackage{tabularx}
\usepackage{multirow}
\usepackage{subcaption}
\usepackage{longtable}
\usepackage[most]{tcolorbox}
\tcbuselibrary{breakable}
\usepackage{titletoc}
\usepackage{xspace}
\usepackage{soul}

\usepackage{hyperref}
\hypersetup{colorlinks=true, linkcolor=blue, citecolor=blue}
\usepackage{minitoc}
\usepackage{inconsolata}
\usepackage{url}
\usepackage[linesnumbered,ruled,vlined]{algorithm2e}
\usepackage{multirow}
\usepackage{booktabs}
\usepackage{caption}
\usepackage{subcaption}
\usepackage{xspace}
\usepackage{amsmath}
\usepackage{amssymb}
\usepackage{newtxmath}
\usepackage{bbm}
\usepackage{graphicx}
\usepackage{tabularx}
\usepackage{bbding}
\usepackage{pifont}
\usepackage{diagbox}
\usepackage{makecell}
\usepackage{colortbl}
\usepackage{bbm}
\usepackage{color}
\usepackage{hhline}
\usepackage{arydshln}
\usepackage{soul}
\usepackage{float}
\usepackage[most]{tcolorbox}
\usepackage{enumitem}
\hypersetup{
    colorlinks=true,
    linkcolor=blue,
    citecolor=blue,
}

\newcommand{\paratitle}[1]{\vspace{1.5ex}\noindent\textbf{#1}}

\newcommand{\ignore}[1]{}

\newcommand{\mcpicon}[1]{%
  \IfFileExists{icons/#1}{%
    \raisebox{-0.2\height}{\includegraphics[height=1.2em]{icons/#1}}\,%
  }{%
    \IfFileExists{icons/#1.pdf}{%
      \raisebox{-0.2\height}{\includegraphics[height=1.2em]{icons/#1.pdf}}\,%
    }{%
      \IfFileExists{icons/#1.png}{%
        \raisebox{-0.2\height}{\includegraphics[height=1.2em]{icons/#1.png}}\,%
      }{%
        \IfFileExists{icons/#1.jpg}{%
          \raisebox{-0.2\height}{\includegraphics[height=1.2em]{icons/#1.jpg}}\,%
        }{%
          \IfFileExists{figures/icons/#1}{%
            \raisebox{-0.2\height}{\includegraphics[height=1.2em]{figures/icons/#1}}\,%
          }{%
            \IfFileExists{figures/icons/#1.pdf}{%
              \raisebox{-0.2\height}{\includegraphics[height=1.2em]{figures/icons/#1.pdf}}\,%
            }{%
              \IfFileExists{figures/icons/#1.png}{%
                \raisebox{-0.2\height}{\includegraphics[height=1.2em]{figures/icons/#1.png}}\,%
              }{%
                \IfFileExists{figures/icons/#1.jpg}{%
                  \raisebox{-0.2\height}{\includegraphics[height=1.2em]{figures/icons/#1.jpg}}\,%
                }{%
                }%
              }%
            }%
          }%
        }%
      }%
    }%
  }%
}

\definecolor{clrClaude}{HTML}{D97706}
\newcommand{\modelicon}[2]{%
  \IfFileExists{figures/icons/#2}{%
    \raisebox{-0.2\height}{\includegraphics[height=#1]{figures/icons/#2}}\,%
  }{}%
}

\definecolor{takeaway}{RGB}{165, 209, 216}
\definecolor{takeawayTitle}{RGB}{57, 89, 163}

\definecolor{bdDeep}{HTML}{315AB4}
\definecolor{bdBright}{HTML}{4ABFB5}
\newcommand{\elegantAgentWorld}{%
\textcolor{bdDeep}{A}%
\textcolor{bdDeep!90!bdBright}{g}%
\textcolor{bdDeep!80!bdBright}{e}%
\textcolor{bdDeep!70!bdBright}{n}%
\textcolor{bdDeep!60!bdBright}{t}%
\textcolor{bdDeep!50!bdBright}{-}%
\textcolor{bdDeep!40!bdBright}{W}%
\textcolor{bdDeep!30!bdBright}{o}%
\textcolor{bdDeep!20!bdBright}{r}%
\textcolor{bdDeep!10!bdBright}{l}%
\textcolor{bdBright}{d}%
}

\usepackage[most]{tcolorbox}
\tcbuselibrary{listings,breakable,skins}
\usepackage{booktabs,tabularx,longtable,multirow,array}
\usepackage{enumitem}
\usepackage{listings}
\usepackage[dvipsnames,svgnames,table]{xcolor}
\usepackage[most]{tcolorbox}
\tcbuselibrary{listings,breakable,skins}
\usepackage{booktabs,tabularx,longtable,multirow,array}
\usepackage{enumitem}
\usepackage{listings}
\usepackage[dvipsnames,svgnames,table]{xcolor}
\usepackage{fontawesome}
\usepackage{amssymb}

\definecolor{CaseBlue}{HTML}{2471A3}
\definecolor{CaseBlueBg}{HTML}{EBF5FB}
\definecolor{CaseGreen}{HTML}{229954}
\definecolor{CaseGreenBg}{HTML}{E8F8F0}
\definecolor{CaseOrange}{HTML}{CA6F1E}
\definecolor{CaseOrangeBg}{HTML}{FEF5E7}
\definecolor{PromptPurple}{HTML}{6C3483}
\definecolor{PromptBlueBg}{HTML}{EBF5FB}
\definecolor{RoleUser}{HTML}{1A5276}
\definecolor{RoleAssistant}{HTML}{1E8449}
\definecolor{RoleTool}{HTML}{B9770E}
\definecolor{ActionBlue}{HTML}{2E86C1}
\definecolor{ThinkGray}{HTML}{707B7C}

\lstdefinestyle{promptstyle}{%
  basicstyle=\ttfamily\scriptsize,
  breaklines=true,
  frame=none,
  columns=fullflexible,
  keepspaces=true,
  backgroundcolor=\color{white},
}

\title{\elegantAgentWorld: Scaling Real-World Environment Synthesis for Evolving General Agent Intelligence}

\author{Renmin University of China}
\author{  ByteDance Seed}
\affiliation{See \hyperref[sec:contribution]{Contributions} section for a full author list.}

\abstract{
Large language models are increasingly expected to serve as general-purpose agents that interact with external, stateful tool environments. The Model Context Protocol (MCP) and broader agent skills offer a unified interface for connecting agents with scalable real-world services, but training robust agents remains limited by the lack of realistic environments and principled mechanisms for life-long learning. In this paper, we present \textbf{Agent-World}, a self-evolving training arena for advancing general agent intelligence through scalable environments. Agent-World has two main components: \textbf{(1) Agentic Environment-Task Discovery}, which autonomously explores topic-aligned databases and executable tool ecosystems from thousands of real-world environment themes and synthesizes verifiable tasks with controllable difficulty; and \textbf{(2) Continuous Self-Evolving Agent Training}, which combines multi-environment reinforcement learning with a self-evolving agent arena that automatically identifies capability gaps through dynamic task synthesis and drives targeted learning, enabling the co-evolution of agent policies and environments. Across 23 challenging agent benchmarks, Agent-World-8B and 14B consistently outperforms strong proprietary models and environment scaling baselines. Further analyses reveal scaling trends in relation to environment diversity and self-evolution rounds, offering insights for building general agent intelligence.}

\date{\today}
\correspondence{Guanting Dong at \email{dongguanting@ruc.edu.cn}, Zhicheng Dou at \email{dou@ruc.edu.cn}}
\checkdata[Project Page]{\url{https://agent-tars-world.github.io/-/}}

\begin{document}
\maketitle

\begin{figure}[t]
    \centering
    \includegraphics[width=1\linewidth]{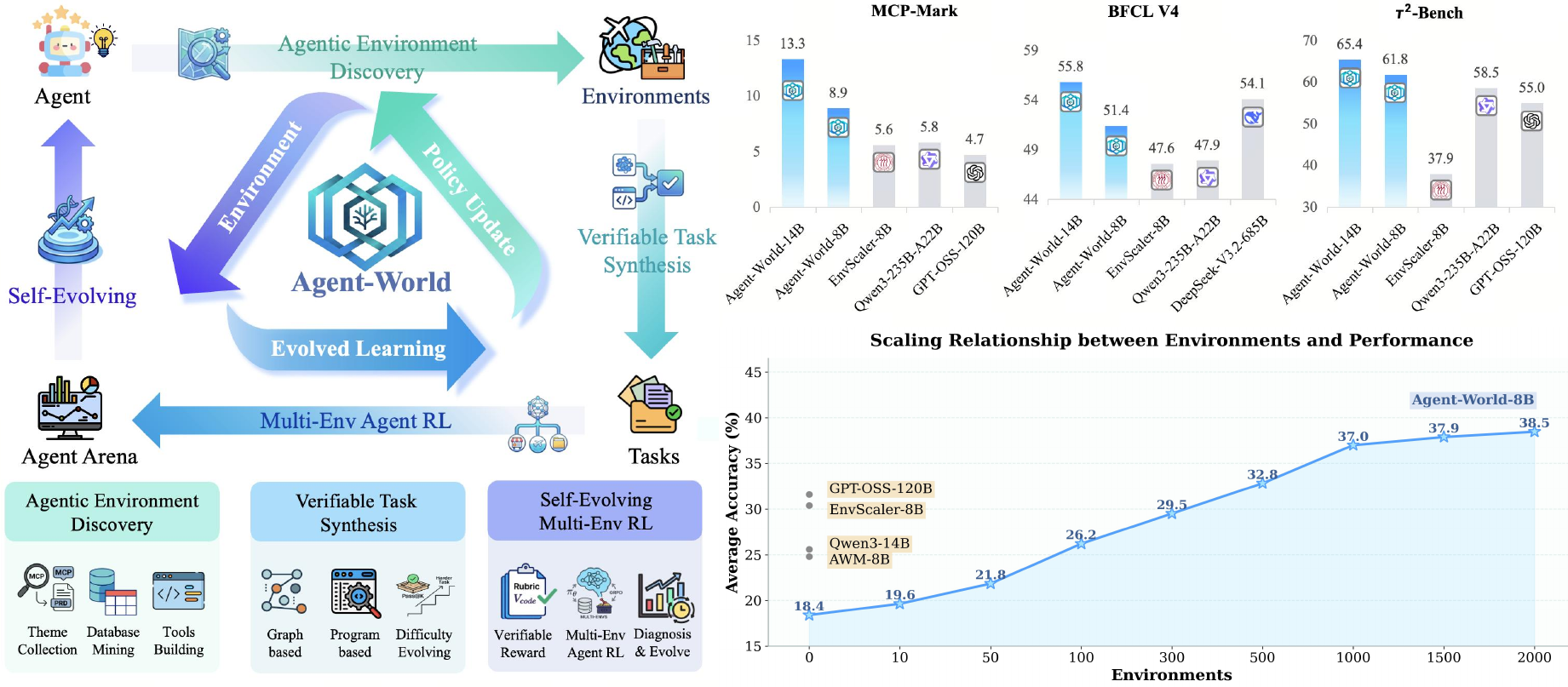}
    \caption{\textbf{Overview of Agent-World (left) and downstream general agent performance (right).} The environment-scaling analysis reports the average score across representative subdomains of MCP-Mark, BFCL V4, and $\tau^2$-Bench.}
    \label{fig:intro}
\end{figure}


\ifdefined\SHENPI
\section{Introduction}
\label{sec:intro}

In recent years, large language models (LLMs) have delivered remarkable progress across a wide range of language understanding and decision-making tasks~\citep{2409_openai_o1,deepseek-r1,team2025kimi,qwen3}. As their capability frontier continues to expand, expectations for LLMs are shifting from chat-oriented text generation toward general-purpose agent assistants. Ideally, such agents should seamlessly integrate real-world interaction with verbal reasoning, and continuously learn from experience to improve themselves, much like human intelligence~\citep{shinn2023reflexion,park2023generativeagentsinteractivesimulacra,zhong2024agieval,feng2024large}. Realizing these agentic capabilities requires not only training LLMs in dynamic environments, but also equipping them with executable tools. On this basis, agents can take actions and observe timely feedback from the environment, forming a \textit{“Generation–Execution–Feedback”} interaction loop~\citep{yao2022react,yao2025taubench,patil2025the,OSWorld,lu-etal-2025-toolsandbox}.

With the rise of agentic reinforcement learning (Agent RL), several agent systems built on static tool environments have demonstrated strong practical value, especially in deep information-seeking and software engineering~\citep{zhang2025landscape,dong2025arpo,webthinker,WebDancer,search-r1,yang2025swesmith}. However, open-world tool environments are inherently compositional and stateful. For instance, in a flight-booking workflow, an agent should follow a valid action order (check inventory $\rightarrow$ execute booking $\rightarrow$ update the calendar), while each action also modifies the underlying environment state. Consequently, agents must orchestrate multi-tool usage flows while tracking state transitions induced by their agent-environment interactions. Prior work centered on stateless or single-tool settings is therefore insufficient for realistic
applications~\citep{search-r1,retool,torl}. This limitation has motivated growing interest in building general agents around standards
such as the Model Context Protocol (MCP)~\citep{mcp_spec_2025} and broader agent skills~\citep{jiang2026xskillcontinuallearningexperience,yu2026polyskilllearninggeneralizableskills}, as well as harness engineering~\citep{pan2026naturallanguageagentharnesses,lou2026autoharnessimprovingllmagents,lopopolo2026harness_engineering}. In this setting, an ideal agent serves as a unified orchestrator that can invoke scalable real-world tools, track state changes in real time, and seamlessly integrate large-scale agentic services into automated workflows~\citep{hou2025model,luo2025mcp,li2025toolathlon}.

Importantly, a key requirement for such general-purpose agents is access to diverse and realistic interactive environments~\citep{huang2025scaling,andrews2025scaling}. However, manually constructing such environments is expensive and difficult to scale, which has driven research toward two main directions: \textbf{(i) Simulated environments} use LLMs as implicit textual world models to produce environment feedback for agent training~\citep{xiao2026webworldlargescaleworldmodel,li2025wordworldlargelanguage,guo2025genenvdifficultyalignedcoevolutionllm,feng2025webworldmodels,li2025simulatingenvironmentsreasoningmodels,wen2025realtimereasoningagentsevolving}. While highly scalable, such simulators are vulnerable to hallucinations and often deviate from real-world dynamics. In contrast, \textbf{(ii) Realistic environments} combine executable tools with real databases, providing stronger grounding for complex interactions~\citep{liu2025deepseek,wu2026autowebworldsynthesizinginfiniteverifiable,tu2026scaleenvscalingenvironmentsynthesis,zhang2026infinitewebscalablewebenvironment,bai2026webgymscalingtrainingenvironments,zhang2025autoenvautomatedenvironmentsmeasuring,zala2024envgengeneratingadaptingenvironments,xi2026toolgymopenworldtoolusingenvironment,xie2024osworldbenchmarkingmultimodalagents,prabhakar2025apigenmtagenticpipelinemultiturn,song2025agentdataprotocolunifying,zhai2025agentevolverefficientselfevolvingagent,cai2025autoforgeautomatedenvironmentsynthesis,sun2026agentskillerscalinggeneralistagent}. Benchmarks such as $\tau^2$-Bench and MCP-Universe have moved evaluation closer to frontier agent applications through stateful environments~\citep{patil2025the,yao2025taubench,chen-etal-2025-acebench}. More recently, several studies have taken initial steps toward synthesizing programmatic environments and tasks for agent training~\citep{fang2025towards,song2026envscaler,tu2026scaleenvscalingenvironmentsynthesis,chen2026divescalingdiversityagentic}.
Unfortunately, their reliance on single-round training makes it difficult for agents to acquire robust, transferable interaction logic in broad environment spaces.

Consequently, although these approaches improve the efficiency of environment construction, two key bottlenecks remain unresolved:
A key bottleneck today is not only environment scale, but continual improvement. Although prior work has made progress in scalable simulated and realistic environments~\citep{huang2025scaling,andrews2025scaling,xiao2026webworldlargescaleworldmodel,li2025wordworldlargelanguage,guo2025genenvdifficultyalignedcoevolutionllm,feng2025webworldmodels,li2025simulatingenvironmentsreasoningmodels,wen2025realtimereasoningagentsevolving,tu2026scaleenvscalingenvironmentsynthesis,song2026envscaler}, most pipelines still emphasize one-pass training and lack a persistent mechanism that diagnoses policy weaknesses and feeds them into the next training round. This makes robust transfer to unseen failure modes difficult.

In this paper, we propose \textbf{Agent-World}, a general-purpose agent training arena that unifies scalable environment synthesis with continuous self-evolving training. As shown in Figure~\ref{fig:intro}, Agent-World follows a two-stage design that forms a closed-loop training process.

\textbf{(1) Continuous Self-Evolving Agent Training.} We train agents via multi-environment reinforcement learning over ``agent--tool--database'' interaction rollouts, using executable rewards for state-aware supervision. Notably, our environment ecosystem naturally serves as a self-evolving arena for envolving agents. Built on these scalable environments, the arena can iteratively synthesize new tasks, automatically identify capability gaps in trained agents, and drive targeted learning, thereby forming a co-evolution loop between agent policies and environments.

\textbf{(2) Agentic Environment-Task Discovery.} To provide high-quality training environments for the self-evolving loop, we use a deep-research agent autonomously mines topic-aligned real-world databases and executable tool interfaces from real-world MCP themes. Then we introduce graph-based task synthesis for generating high-quality agentic tasks.

We conduct comprehensive evaluations on \textbf{23 benchmarks} covering agentic tool use, software engineering, deep research, and general reasoning. As shown in Figure~\ref{fig:intro}, Agent-World consistently outperforms strong proprietary models and competitive baselines. Our analysis further reveals clear scaling relationships among the number of synthesized environments, self-evolution rounds, and downstream agent performance, providing empirical insights into the development of more general agent intelligence.

In summary, our main contributions are as follows:
\begin{itemize}[leftmargin=1em]
\item We introduce \textbf{Agent-World}, a general-purpose agent training arena that unifies scalable environment synthesis with a continuous self-evolving training mechanism, forming a co-evolution loop between agent policies and environments.

\item We propose \textbf{Continuous Self-Evolving Agent Training}, which integrates multi-environment agentic RL with a self-evolving arena to automatically diagnose agent weaknesses and drive targeted learning in a closed training loop.

\item We propose \textbf{Agentic Environment-Task Discovery}, which mines realistic executable environments from real-world environment themes and synthesizes diverse verifiable tasks with controllable difficulty.

\item Experiments across 23 challenging agent benchmarks demonstrate the superior performance of Agent-World. Further analysis reveals scaling relationships among environment diversity, evolution rounds, and agent performance.
\end{itemize}

\section{Preliminary: Agentic Interaction with Multi-Environments}
\label{sec:preliminary}

We model multi-turn agentic interaction with external environments as a Partially Observable Markov Decision Process (POMDP)~\citep{cassandra1998survey},
represented by the tuple $(U,S,A,O,P)$.

\paragraph{\textbf{Intent space ($U$).}}
Let $q\in U$ denote the user's latent intent. The assistant progressively infers $q$ from the accumulated interaction history and environment feedback
to choose appropriate actions.

\paragraph{\textbf{State space ($S$).}}
We factor the global state into an environment state and a dialogue state:
$S = S_E \times S_H$.
At turn $t$, the full state is $s_t=(s_t^E,s_t^H)\in S$.
The environment state $s^E\in S_E$ captures the external world the assistant can query or modify (e.g., databases, files, services),
while the dialogue state $s^H\in S_H$ summarizes conversational context (e.g., dialogue history, constraints, user preferences).

\paragraph{\textbf{Databases and tools (environment carrier and executable interface).}}
To connect the POMDP with multiple environments,
we explicitly parameterize each environment by a pair
$e = (\mathcal{D}, \mathcal{F})$, where $\mathcal{D}$ denotes an environment database and $\mathcal{F}$ denotes a toolset.
Concretely, the database $\mathcal{D}$ is a primary carrier (storage) of the environment state $s^E$---it contains the structured records and/or files
that constitute the mutable external world.
The toolset $\mathcal{F}=\{f_k\}$ provides executable interfaces to interact with $s^E$:
each tool $f\in\mathcal{F}$ can be seen as a callable operator that reads and optionally writes the database,
thereby inducing environment state transitions.

\paragraph{\textbf{Action space ($A$).}}
The assistant chooses between tool-use actions and language-response actions: $A = A_{\text{tool}} \cup A_{\text{resp}}$. For $a_t\in A_{\text{tool}}$, the assistant invokes a tool with structured arguments (e.g., a function name with JSON parameters) to query/modify the environment;
for $a_t\in A_{\text{resp}}$, it emits a natural-language message (including intermediate responses or the final answer).

\paragraph{\textbf{Observation space ($O$).}}
At each turn $t$, the assistant observes $o_t\in O$ and then takes an action $a_t$.
We define $O = O_E \cup O_H$, where $O_E$ contains structured tool observations returned by tool execution (e.g., query results, logs, error codes),
and $O_H$ contains dialogue-side observations (e.g., user utterances, system prompts, or an explicit termination signal in offline training).
Importantly, the environment state $s^E$ is not directly observed and must be inferred indirectly from tool observations in $O_E$.

\paragraph{\textbf{State dynamics ($P$).}}
The transition model $P: S\times A \rightarrow \Pi(S\times O)$ specifies how the system evolves after an action.
Given $(s_t,a_t)$, the process transitions to $s_{t+1}$ and emits the next observation $o_{t+1}$:

\begin{itemize}[leftmargin=1em]
\item If $a_t\in A_{\text{tool}}$, a tool $f\in\mathcal{F}$ is executed against the database $\mathcal{D}$ (i.e., the carrier of $s_t^E$).
This execution may update the environment state $s_{t+1}^E$ via reads/writes on $\mathcal{D}$ and produces a structured observation
$o_{t+1}^E\in O_E$.
The dialogue state $s_{t+1}^H$ is updated by appending the new tool interaction.
\item If $a_t\in A_{\text{resp}}$, the assistant updates the dialogue state (i.e., $s_{t+1}^H$) by emitting a response.
In interactive settings this may lead to a new user observation $o_{t+1}^H\in O_H$; in offline training it typically yields a termination signal.
The environment state remains unchanged for that turn: $s_{t+1}^E = s_t^E$.
\end{itemize}

\section{Methodology}
\label{sec:method}

We propose \textbf{Agent-World}, a general training arena for MCP agents that couples continuous self-evolving training with scalable environment synthesis. The Agent-World pipeline consists of two stages:

(1) \textbf{Continuous Self-Evolving Agent Training} (Sec.~\ref{sec:self-evolving-training}). We train agents with multi-environment ``agent--tool--database'' interaction rollouts and executable rewards. Crucially, our environment ecosystem doubles as an agentic diagnostic arena that continuously identifies agent weaknesses and drives targeted environment-task expansion, forming a closed-loop self-evolutionary training cycle.

(2) \textbf{Agentic Environment-Task Discovery} (Sec.~\ref{section:method-env-discovery}). To provide high-quality training environments for the self-evolving loop, a deep-research agent autonomously mines topic-aligned real-world databases from over 2K themes and compiles executable toolsets, forming a scalable environment ecosystem. Then we introduce graph-based task synthesis for generating high-quality agentic tasks.

Below, we describe each component in detail.

\subsection{Continuous Self-Evolving Agent Training}\label{sec:self-evolving-training}

Given a scalable environment ecosystem $\mathcal{E}=\{e_i=(\mathcal{D}_i,\mathcal{F}_i)\}$ where each environment pairs a database with an executable toolset (Sec.~\ref{sec:preliminary}; construction detailed in Sec.~\ref{section:method-env-discovery}), we perform multi-environment agent reinforcement learning with ``agent--tool--database'' interaction rollouts, using executable verification for state checking and reward assignment. Crucially, $\mathcal{E}$ doubles as an \emph{agentic diagnostic arena}: the trained agent is continuously deployed into unseen environments to automatically locate its weaknesses, which then guides targeted expansion of environments and tasks. This diagnosis-driven expansion feeds back into the next round of training, forming a closed-loop, online self-evolutionary cycle that progressively strengthens the agent's generalization capability.

\subsubsection{Multi-Environment Agent Reinforcement Learning}\label{sec:me-rl}

Unlike static tool-calling settings, our training implements a closed-loop ``agent--tool--database'' interaction: the LLM policy $\pi_\theta$ generates reasoning and tool-call decisions, the tool runtime executes environment-specific tools $\mathcal{F}_e$ in a sandbox while maintaining environment-side states, and the database $\mathcal{D}_e$ serves as the read/write substrate that provides a verifiable, updatable structured data backbone.

Formally, given a task $x$ and environment $e=(\mathcal{D}_e,\mathcal{F}_e)$, the policy $\pi_\theta$ samples actions based on instruction and history $h_t=(o_0,a_0,\dots,o_t)$ following Section~\ref{sec:preliminary}. Tool actions execute $f\in\mathcal{F}_e$ on $\mathcal{D}_e$ and return structured observations $o_{t+1}^{E}\in O_E$; response actions terminate the trace. This yields an executable trajectory $\tau=(o_0,a_0,\ldots,o_T,a_T)$. Following GRPO~\citep{grpo}, we sample $N$ trajectories per task, with tasks in each global batch paired with independent environments to realize cross-environment rollouts.

\begin{figure*}[!t]
\centering
\includegraphics[width=\linewidth]{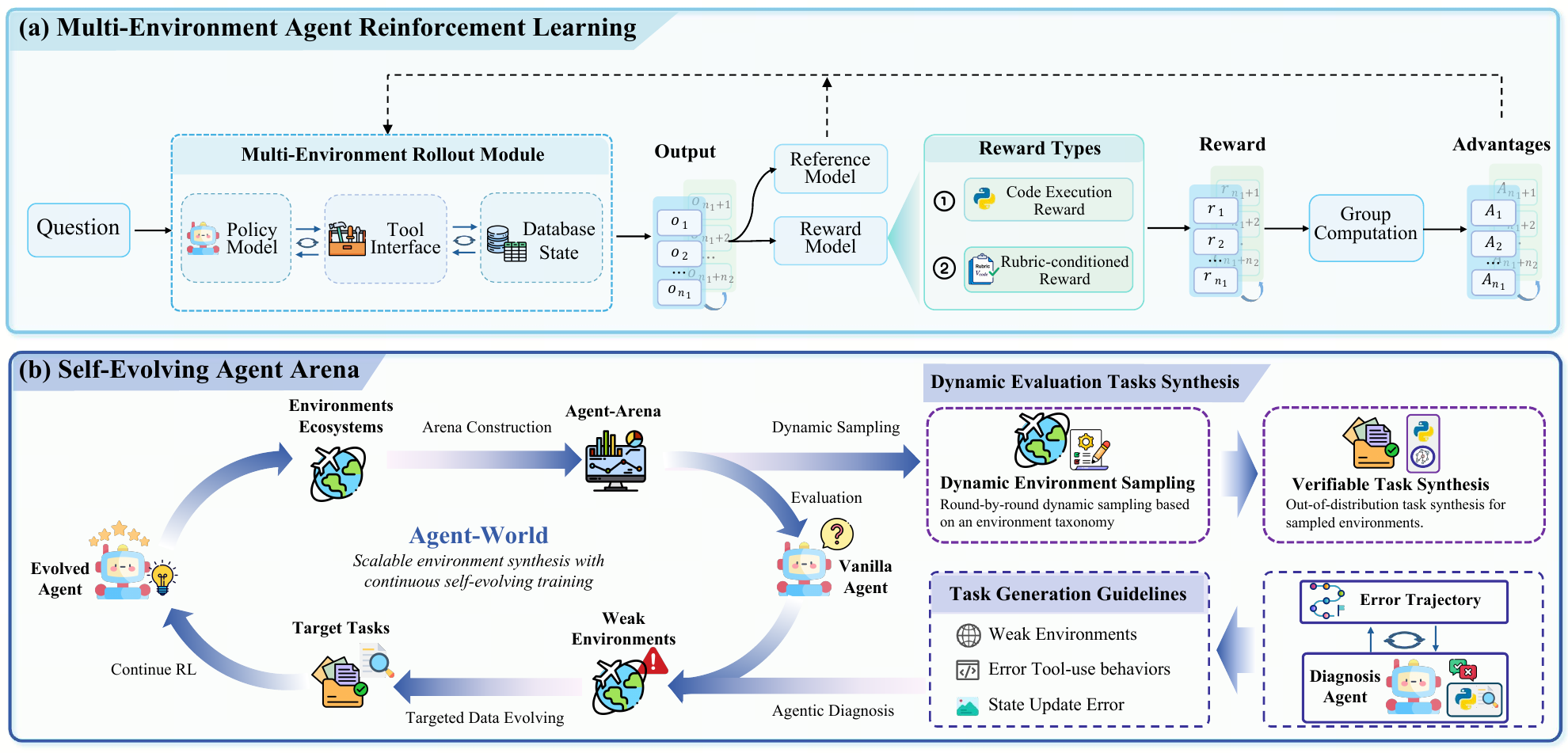}

\caption{
\textbf{The Overall Framework of Continuous Self-Evolving Agent Training.} The agent is trained with multi-environment RL under executable rewards (top), evaluated in a dynamic arena, diagnosed for capability gaps, and improved through targeted environment-task expansion (bottom).
}
\label{fig:training}
\end{figure*}

\textbf{Structured Verifiable Reward.} Distinct from prior static tool-RL settings~\citep{retool,dong2025toolstar}, reward assignment for environment agents must account for factors beyond answer correctness, including environment state and format compliance. In this version, we use \textbf{graph-based tasks} ($\mathcal{X}_{\text{graph}}$), each equipped with a structured rubric $R=\{r_j\}_{j=1}^{n}$ evaluated by a rubric-conditioned LLM judge. The trajectory-level reward is: $r(\tau)=\mathbb{I}\Big[\frac{1}{n}\sum_{j=1}^{n}\mathbb{I}\big[\mathrm{Judge}(r_j,\hat a)\big]\ge 0.8\Big],\quad x\in \mathcal{X}_{\text{graph}}.$

\textbf{Policy Update.} We adopt Group Relative Policy Optimization (GRPO) to maximize the verifiable returns. For each query $q$, we draw $G$ trajectories from $\pi_{\theta_{\text{old}}}(\cdot\mid q)$, compute token-level advantages $\hat{A}_{i,t}$, and update $\pi_\theta$ via:
\begin{equation}
\begin{split}
J_{\text{GRPO}}(\theta) = \mathbb{E}_{(q,a) \sim D, \{y_i\}_{i=1}^G \sim \pi_{\theta_{\text{old}}}(\cdot \mid q)} \Bigg[ &\frac{1}{G} \sum_{i=1}^G \frac{1}{|y_i|} \sum_{t=1}^{|y_i|} \min \Big( r_{i,t}(\theta) \hat{A}_{i,t}, \\
& \text{clip} \left( r_{i,t}(\theta), 1 - \epsilon, 1 + \epsilon \right) \hat{A}_{i,t} \Big) - \beta D_{\text{KL}}(\pi_{\theta} \parallel \pi_{\text{ref}}) \Bigg].
\end{split}
\label{eq:grpo}
\end{equation}
where $\epsilon$ and $\beta$ are hyperparameters and $\hat{A}_{i,t}$ is the normalized advantage of the $i$-th rollout within the group.

\begin{algorithm}[t]
\caption{\textbf{Self-Evolving Agent Arena Loop}}
\label{alg:self-evolving-arena}
\KwIn{Environment ecosystem $\mathcal{E}$; arena subset $\mathcal{E}_{\text{arena}}\subset\mathcal{E}$; initial policy $\pi_{\theta^{(0)}}$; number of rounds $R$}
\KwOut{Improved policy $\pi_{\theta^{(R)}}$}
\For{$r=0,\dots,R-1$}{
    \tcp{\textbf{Phase 1: Dynamic Evaluation Task Synthesis}}
    \ForEach{$e_i=(\mathcal{D}_i,\mathcal{F}_i)\in\mathcal{E}_{\textup{arena}}$}{
        Synthesize fresh verifiable tasks $\mathcal{X}^{(r)}_{\textup{arena}}(e_i)$ with executable rubrics or validators (Sec.~\ref{section:method-task-synthesis})\;
    }
    Evaluate $\pi_{\theta^{(r)}}$ on $\mathcal{X}^{(r)}_{\textup{arena}}=\bigcup_{i}\mathcal{X}^{(r)}_{\textup{arena}}(e_i)$ under closed-loop rollouts (Sec.~\ref{sec:me-rl})\;

    \tcp{\textbf{Phase 2: Agentic diagnosis}}
    Feed per-task failure traces, error distribution statistics, and environment metadata to diagnosis agent $\delta$\;
    $\delta$ outputs weak environments $\mathcal{W}^{(r)}\!\subseteq\!\mathcal{E}_{\textup{arena}}$ and task-generation guidelines $\mathcal{G}^{(r)}_{\textup{guide}}$\;

    \tcp{\textbf{Phase 3: Agent-Environment Co-Evolution.}}
    \ForEach{$e\in\mathcal{W}^{(r)}$}{
        Optionally complexify database: $\mathcal{D}_e\leftarrow\phi(\mathcal{D}_e,\cdot)$ via Eq.~(2)\;
        Generate targeted tasks $\mathcal{X}^{(r)}_{\textup{target}}(e)$ conditioned on $\mathcal{G}^{(r)}_{\textup{guide}}(e)$\;
    }
    Continue RL from $\pi_{\theta^{(r)}}$ on $\mathcal{X}^{(r)}_{\textup{target}}=\bigcup_{e\in\mathcal{W}^{(r)}}\!\mathcal{X}^{(r)}_{\textup{target}}(e)$ to obtain $\pi_{\theta^{(r+1)}}$\;
}
\Return{$\pi_{\theta^{(R)}}$}\;
\end{algorithm}

\subsubsection{Self-Evolving Agent Arena}\label{sec:self-evolving-arena}

\textbf{Motivation.}
Our scalable environment ecosystem $\mathcal{E}=\{(\mathcal{D}(m),\mathcal{F}(m))\mid m\in\mathcal{M}\}$ serves not only as a training source but also as an agentic diagnostic arena. Beyond synthesizing training data, we aim to continuously identify weaknesses of the current agent policy and then expand environments and tasks in a targeted manner to close those gaps. This yields a self-reinforcing loop in which evaluation, diagnosis, and data generation evolve together with the agent.

\textbf{Arena Construction.}
Based on the hierarchical environment taxonomy (Sec.~\ref{section:method-env-discovery}), we construct an evaluation arena by stratified sampling.
Specifically, for each first-tier category $c\in\mathcal{C}$, we randomly select $K$ environments ($K=5$) and merge them into the arena set
$\mathcal{E}_{\text{arena}}=\{e_i\}_{i=1}^{|\mathcal{E}_{\text{arena}}|}$, where each environment is $e_i=(\mathcal{D}_i,\mathcal{F}_i)$.
This design ensures broad coverage over different environment types while keeping evaluation cost controllable.

\textbf{Dynamic Evaluation Task Synthesis.}
For each arena environment $e_i$, we follow Section~\ref{section:method-task-synthesis} and synthesize a fresh batch of verifiable tasks and validators at each iteration.
Concretely, at iteration $r$ we instantiate a task set
$\mathcal{X}^{(r)}_{\text{arena}}(e_i)$ consisting of both graph-based tasks and programmatic tasks, each paired with an executable rubric $R$ or verification code $V_{\text{code}}$.
Importantly, both the sampled environments and the synthesized tasks are dynamic across rounds, preventing overfitting to a static benchmark and enabling continual diagnosis.

\textbf{Agentic Diagnosis.}
Given a trained agent policy $\pi_{\theta^{(r)}}$, we evaluate it on $\mathcal{X}^{(r)}_{\text{arena}}$ under the same closed-loop \texttt{agent-tool-database} execution protocol (Sec.~\ref{sec:me-rl}).
We then introduce a diagnosis agent $\delta$ equipped with a Python interpreter and search tools, which takes as input:
(i) per-task failure traces (tool logs, intermediate observations, and validator feedback),
(ii) error distribution statistics by environment, broken down by taxonomy categories, and (iii) environment metadata (tool schemas/database descriptions).

The diagnosis agent outputs (a) a ranked set of weak environments
$\mathcal{W}^{(r)}\subseteq \mathcal{E}_{\text{arena}}$ and (b) environment-specific task-generation guidelines
$\mathcal{G}^{(r)}_{\text{guide}}(e)$ that describe the missing capabilities (e.g., erroneous tool use or state-update mistakes).

\textbf{Agent-Environment Co-Evolution.}
Conditioned on $\mathcal{W}^{(r)}$ and $\mathcal{G}^{(r)}_{\text{guide}}$, we re-run the verifiable task synthesis pipeline (Sec.~\ref{section:method-task-synthesis}) to generate a targeted training set
$\mathcal{X}^{(r)}_{\text{target}}$,
optionally accompanied by environment expansion via database complexification (Eq.~(2)) when the weakness is due to insufficient state diversity.
Starting from $\pi_{\theta^{(r)}}$, we then perform multi-environment agent RL (Sec.~\ref{sec:me-rl}) on the augmented data to obtain an improved policy $\pi_{\theta^{(r+1)}}$.
Iterating the above steps yields a self-evolving agent-arena loop:
\[
\pi_{\theta^{(r)}} \xrightarrow{\text{evaluate}}
\mathcal{W}^{(r)} \xrightarrow{\text{diagnose+target}}
\mathcal{X}^{(r)}_{\text{target}} \xrightarrow{\text{continue RL}}
\pi_{\theta^{(r+1)}}.
\]

This arena-driven loop turns scalable environments into an automated curriculum engine, continuously driving targeted learning and enabling the co-evolution of agent policies and environments.

\subsection{Agentic Environment-Task Discovery}\label{section:method-env-discovery}

\begin{figure*}[!t]
\centering
\includegraphics[width=\linewidth]{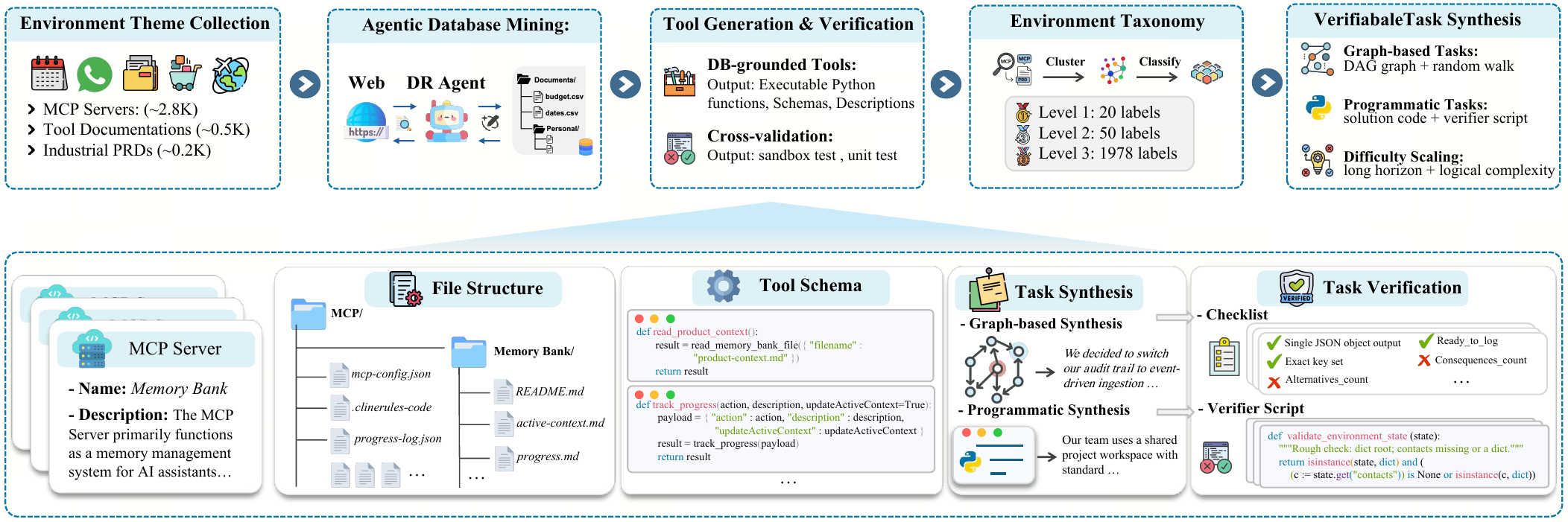}
\caption{
\textbf{The Pipeline of Agentic Environment-Task Discovery.} We start from real-world environment themes, mine topic-aligned databases from the web, generate and verify executable tool interfaces, and synthesize verifiable tasks with controllable difficulty.
}
\label{fig:env-task-discovery}
\end{figure*}

The self-evolving training loop described above relies on a scalable, realistic environment ecosystem. In this section, we detail how to automatically construct such an ecosystem from real-world sources.

\textbf{Environment Theme Collection.} We gather environment themes from two sources: (1)~real-world MCP server specifications from Smithery\footnote{\url{https://smithery.ai/servers}} ($m\in\mathcal{M}_1$), and (2)~tool-definition documents extracted from open-source datasets and inversely mapped to topics via an LLM ($m\in\mathcal{M}_2$), forming the seed topic set $\mathcal{M}=\mathcal{M}_1\cup\mathcal{M}_2$.

\textbf{Agentic Database Mining.} Unlike prior work relying on LLM-synthesized databases~\citep{li2025wordworldlargelanguage,guo2025genenv,li2025simulating}, we build a deep-research agent (with Search, Browser, Code Compiler, and OS tools) that autonomously mines topic-aligned structured data from the web: $\mathcal{D}(m)=\mathcal{G}(m;\pi_\theta,\mathcal{T})$. A database complexification process $\phi$ then iteratively enriches each database over $N$ rounds: $\mathcal{D}^{(n+1)}(m)=\phi(\mathcal{D}^{(n)}(m),m,\mathcal{T})$.

\textbf{Tool Generation and Verification.} A tool-design agent $\psi$ produces candidate tools and unit tests $(\hat{\mathcal{F}},\hat{\mathcal{C}})=\psi(m,\mathcal{D}^{(N)}(m);\pi_\theta)$. We retain only tools that compile and achieve pass rate $\mathrm{Acc}>0.5$, yielding the environment ecosystem $\mathcal{E} = \{\mathcal{D}(m), \mathcal{F}(m)\mid m\in\mathcal{M}\}$.

\textbf{Environment Taxonomy Construction.} We apply hierarchical clustering~\citep{ward1963hierarchical} over 2K+ themes to obtain 50 second-tier labels, which three annotators merge into 20 first-tier types. The resulting taxonomy provides a foundation for stratified arena construction (Sec.~\ref{sec:self-evolving-arena}).

\subsubsection{Graph-Based Task Synthesis}\label{section:method-task-synthesis}

A valid tool execution sequence naturally defines the data flows needed to answer a user query. Motivated by prior graph-based environment-scaling methods~\citep{song2026envscaler,tu2026scaleenvscalingenvironmentsynthesis}, we introduce Graph-Based Task Synthesis as follow.

\textbf{Tool Graph Construction.} For each environment $e = (\mathcal{D}, \mathcal{F})$, we construct a fully connected, weighted directed graph $G = (V, E)$, where each node corresponds to a tool and each edge encodes call dependencies. Edges are classified into three types by an LLM: (1) strong dependency (output of $f_i$ is a required input to $f_j$), (2) weak dependency (input can optionally come from $f_i$), and (3) independent (no parameter-level dependency, ensuring full connectivity).

We generate a tool-call sequence $\tau$ by performing a weight-biased random walk on $G$, instantiate parameters using preceding outputs or database values, and have an LLM refine the chain into an executable sequence $\tau^*$. Given $\tau^*$, we execute it step-by-step in a Python sandbox, and an LLM drafts a task description $q_{final}$ (without exposing tool names or schemas) along with a ground-truth answer $a^*$ and structured evaluation rubrics $R$~\citep{funreason-mt}. To ensure stability, a ReAct agent solves each task 5 times; only tasks with at least 2 consistent passes are retained.

\textbf{Difficulty Scaling.} We scale difficulty by increasing the random walk length, raising the sampling probability of weak/independent edges, and rewriting task descriptions to obscure tool names and execution logic, forcing agents to deduce workflows from abstract goals.

\begin{figure}[t]
    \centering
    \includegraphics[width=\linewidth]{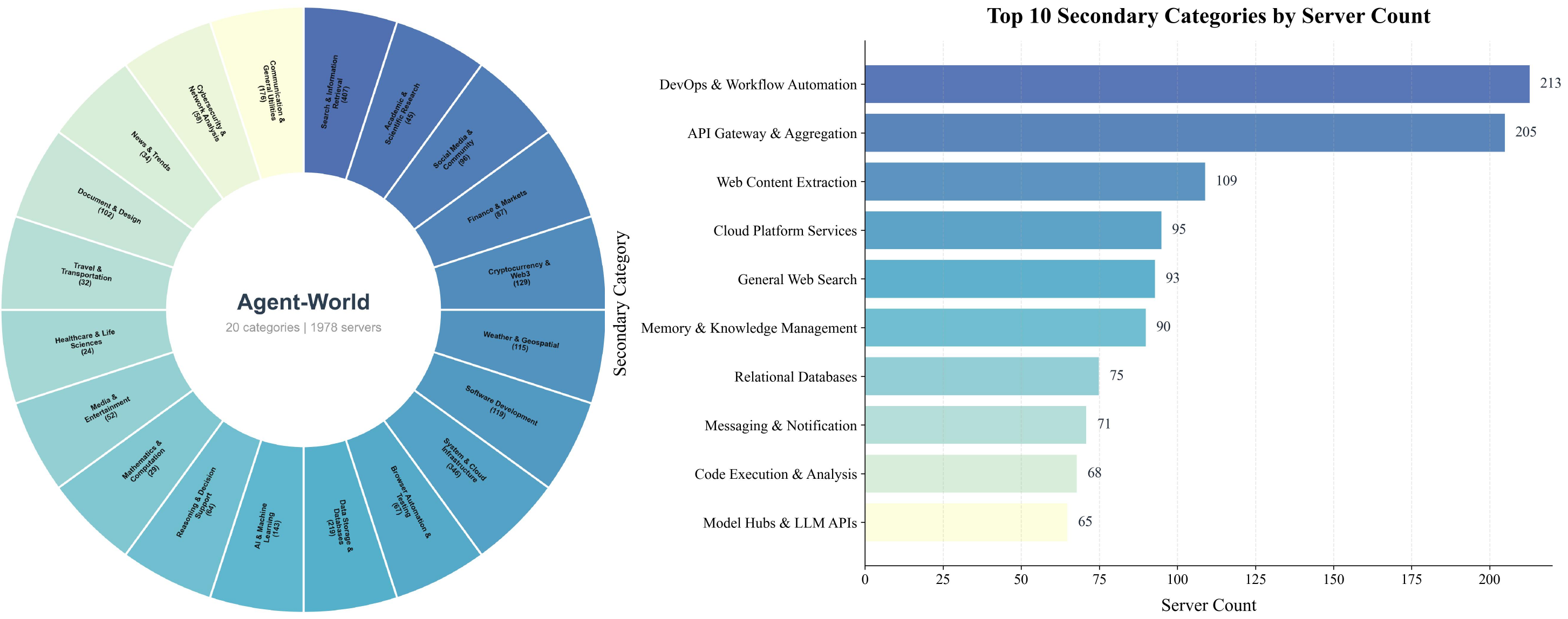}
  \caption{\textbf{Hierarchical environment taxonomy of Agent-World.}
\textbf{Left:} distribution of the 20 first-tier categories with their server counts.
\textbf{Right:} top-10 second-tier categories ranked by server count.}
    \vspace{-1em}
    \label{fig:mcp_taxonomy_method}
\end{figure}

\else
\section{Introduction}
\label{sec:intro}

In recent years, large language models (LLMs) have delivered remarkable progress across a wide range of language understanding and decision-making tasks~\citep{2409_openai_o1,deepseek-r1,team2025kimi,qwen3,seed2026seed18modelcardgeneralized}. As their capability frontier continues to expand, expectations for LLMs are shifting from chat-oriented text generation toward general-purpose agent assistants~\citep{claweval,GAIA,barres2025tau2benchevaluatingconversationalagents,merrill2026terminalbenchbenchmarkingagentshard,li2026skillsbenchbenchmarkingagentskills,he2026openclaw}. Ideally, such agents should seamlessly integrate real-world interaction with verbal reasoning, and continuously learn from experience to improve themselves, much like human intelligence~\citep{shinn2023reflexion,park2023generativeagentsinteractivesimulacra,zhong2024agieval,feng2024large}. Realizing these agentic capabilities requires not only training LLMs in dynamic environments, but also equipping them with executable tools. On this basis, agents can take actions and observe timely feedback from the environment, forming a \textit{“Generation–Execution–Feedback”} interaction loop~\citep{yao2022react,yao2025taubench,patil2025the,OSWorld,lu-etal-2025-toolsandbox}.

With the rise of agentic reinforcement learning (Agent RL), several agent systems built on static tool environments have demonstrated strong practical value, especially in deep information-seeking and software engineering~\citep{zhang2026landscapeagenticreinforcementlearning,sun2026sweworldbuildingsoftwareengineering,dong2025arpo,aepo,webthinker,WebDancer,search-r1,yang2025swesmith}. However, open-world tool environments are inherently compositional and stateful. For instance, in a flight-booking workflow, an agent should follow a valid action order (check inventory $\rightarrow$ execute booking $\rightarrow$ update the calendar), while each action also modifies the underlying environment state. Consequently, agents must orchestrate multi-tool usage flows while tracking state transitions induced by their agent-environment interactions. Prior work centered on stateless or single-tool settings is therefore insufficient for realistic
applications~\citep{search-r1,retool,torl}. This limitation has motivated growing interest in building general agents around standards
such as the Model Context Protocol (MCP)~\citep{mcp_spec_2025,mcpatlas,li2025toolathlon} and broader agent skills~\citep{jiang2026xskillcontinuallearningexperience,yu2026polyskilllearninggeneralizableskills,skillnet}, as well as harness engineering~\citep{harness_survey,pan2026naturallanguageagentharnesses,lou2026autoharnessimprovingllmagents,lopopolo2026harness_engineering}. In this setting, an ideal agent serves as a unified orchestrator that can invoke scalable real-world tools, track state changes in real time, and seamlessly integrate large-scale agentic services into automated workflows~\citep{hou2025model,luo2025mcp,li2025toolathlon}.

Importantly, a key requirement for such general-purpose agents is access to diverse and realistic interactive environments~\citep{huang2025scaling,andrews2025scaling}. However, manually crafting such environments is expensive and difficult to scale, which has driven research toward two main directions: \textbf{(i) Simulated environments} use LLMs as implicit textual world models to produce environment feedback for agent training~\citep{xiao2026webworldlargescaleworldmodel,li2025wordworldlargelanguage,guo2025genenvdifficultyalignedcoevolutionllm,feng2025webworldmodels,li2025simulatingenvironmentsreasoningmodels,wen2025realtimereasoningagentsevolving}. While highly scalable, such simulators are vulnerable to hallucinations and often deviate from real-world dynamics. In contrast, \textbf{(ii) Realistic environments} combine executable tools with real databases, providing stronger grounding for complex interactions~\citep{liu2025deepseek,wu2026autowebworldsynthesizinginfiniteverifiable,tu2026scaleenvscalingenvironmentsynthesis,zhang2026infinitewebscalablewebenvironment,bai2026webgymscalingtrainingenvironments,zhang2025autoenvautomatedenvironmentsmeasuring,zala2024envgengeneratingadaptingenvironments,xi2026toolgymopenworldtoolusingenvironment,xie2024osworldbenchmarkingmultimodalagents,prabhakar2025apigenmtagenticpipelinemultiturn,song2025agentdataprotocolunifying,zhai2025agentevolverefficientselfevolvingagent,cai2025autoforgeautomatedenvironmentsynthesis,sun2026agentskillerscalinggeneralistagent}. Benchmarks such as $\tau^2$-Bench and ClawEval have moved evaluation closer to frontier agent applications through stateful environments~\citep{patil2025the,yao2025taubench,chen-etal-2025-acebench,claweval}. More recently, several studies have taken initial steps toward synthesizing programmatic environments and tasks for agent training~\citep{fang2025towards,song2026envscaler,tu2026scaleenvscalingenvironmentsynthesis,chen2026divescalingdiversityagentic,webworld}.
Unfortunately, their reliance on single-round training makes it difficult for agents to acquire robust, transferable interaction logic in broad environment spaces.

Consequently, although these approaches improve the efficiency of environment construction, two key bottlenecks remain unresolved:

\begin{itemize}[leftmargin=1em]
\item \textbf{Scalable realism and complex environment synthesis:} Existing environments are often purely LLM-generated or derived from limited open-source toolchains, which often mismatch real-world interaction logic. Moreover, synthetic environments are often limited in complexity, restricting the training of agents on long-horizon, state-intensive tasks.

\item \textbf{Continuous self-evolving training mechanisms:} Although realistic environments can naturally serve as effective training arenas, existing work has primarily emphasized environment construction and scaling, while lacking principled mechanisms that use such scalable environments to diagnose agent weaknesses and drive continual self-improvement.

\end{itemize}

In this paper, we propose \textbf{Agent-World}, a general-purpose agent training arena that unifies scalable real-world environment synthesis with continuous self-evolving training. As shown in Figure~\ref{fig:intro}, Agent-World follows a two-stage design that forms a closed-loop training process.

\textbf{(1) Agentic Environment-Task Discovery.} We collect thousands of real-world environment themes and build a deep-research pipeline that autonomously mines topic-aligned databases and executable toolsets from the web, forming a scalable and realistic environment ecosystem (\textbf{including 1978 environments and 19822 tools}). On top of these environments, we synthesize high-quality agent tasks through both graph-based and programmatic generation, and further expand task difficulty with executable verification.

\textbf{(2) Continuous Self-Evolving Agent Training.} We train agents via multi-environment reinforcement learning over ``agent--tool--database'' interaction rollouts, using executable rewards for state-aware supervision. Notably, our environment ecosystem naturally serves as a self-evolving arena for evolving agents. Built on these scalable environments, the arena can iteratively synthesize new tasks, automatically identify capability gaps in trained agents, and drive targeted learning, thereby forming a co-evolution loop between agent policies and environments.

We conduct comprehensive evaluations on \textbf{23 benchmarks} covering \textit{agentic tool-use}, \textit{advanced AI assistant}, \textit{software engineering}, \textit{deep research}, and \textit{general reasoning}. As shown in Figure~\ref{fig:intro}, Agent-World-8B and 14B consistently outperforms strong foundation models and competitive baselines. Our analysis further reveals clear scaling relationships among the number of synthesized environments, self-evolution rounds, and downstream agent performance, providing empirical insights into the development of more general agent intelligence.

In summary, our main contributions are as follows:
\begin{itemize}[leftmargin=1em]
\item We introduce \textbf{Agent-World}, a general-purpose agent training arena that unifies scalable real-world environment synthesis with a continuous self-evolving training mechanism, forming a co-evolution loop between agent policies and environments.

\item We propose \textbf{Agentic Environment-Task Discovery}, which mines realistic executable environments from real-world environment themes and synthesizes diverse verifiable tasks with controllable difficulty.

\item We propose \textbf{Continuous Self-Evolving Agent Training}, which integrates multi-environment agentic RL with a self-evolving arena to automatically diagnose agent weaknesses and drive targeted learning in a closed training loop.

\item Experiments across \textbf{23 challenging agent benchmarks} demonstrate the superior performance of Agent-World. Further analysis reveals scaling relationships among environment diversity, evolution rounds, and agent performance.
\end{itemize}

\section{Preliminary: Agentic Interaction with Multi-Environments}
\label{sec:preliminary}

Following AgentSkiller~\citep{sun2026agentskiller}, we model multi-turn agentic interaction with external environments as a Partially Observable Markov Decision Process (POMDP)~\citep{cassandra1998survey}, represented by the tuple $(U,S,A,O,P)$.

\paragraph{\textbf{Intent space ($U$).}}
Let $q\in U$ denote the user's latent intent. The assistant progressively infers $q$ from the accumulated interaction history and environment feedback to choose appropriate actions.

\paragraph{\textbf{State space ($S$).}}
We factor the global state into an environment state and a dialogue state:
$S = S_E \times S_H$.
At turn $t$, the full state is $s_t=(s_t^E,s_t^H)\in S$.
The environment state $s^E\in S_E$ captures the external world the assistant can query or modify (e.g., databases, files, services),
while the dialogue state $s^H\in S_H$ summarizes conversational context (e.g., dialogue history, constraints, user preferences).

\paragraph{\textbf{Databases and tools.}}
To connect the POMDP with multiple environments, we explicitly parameterize each environment by a pair
$e = (\mathcal{D}, \mathcal{F})$, where $\mathcal{D}$ denotes an environment database and $\mathcal{F}$ denotes a toolset.
Concretely, the database $\mathcal{D}$ is a primary carrier (storage) of the environment state $s^E$---it contains the structured records and/or files
that constitute the mutable external world.
The toolset $\mathcal{F}=\{f_k\}$ provides executable interfaces to interact with $s^E$:
each tool $f\in\mathcal{F}$ can be seen as a callable operator that reads and optionally writes the database,
thereby inducing environment state transitions.

\paragraph{\textbf{Action space ($A$).}}
The assistant chooses between tool-use actions and language-response actions: $A = A_{\text{tool}} \cup A_{\text{resp}}$. For $a_t\in A_{\text{tool}}$, the assistant invokes a tool with structured arguments (e.g., a function name with JSON parameters) to query/modify the environment;
for $a_t\in A_{\text{resp}}$, it emits a natural-language message (including intermediate responses or the final answer).

\paragraph{\textbf{Observation space ($O$).}}
At each turn $t$, the assistant observes $o_t\in O$ and then takes an action $a_t$.
We define $O = O_E \cup O_H$, where $O_E$ contains structured tool observations returned by tool execution (e.g., query results, logs, error codes),
and $O_H$ contains dialogue-side observations (e.g., user utterances, system prompts, or an explicit termination signal in offline training).
Importantly, the environment state $s^E$ is not directly observed and must be inferred indirectly from tool observations in $O_E$.

\paragraph{\textbf{State dynamics ($P$).}}
The transition model $P: S\times A \rightarrow \Pi(S\times O)$ specifies how the system evolves after an action.
Given $(s_t,a_t)$, the process transitions to $s_{t+1}$ and emits the next observation $o_{t+1}$:

\begin{itemize}[leftmargin=1em]
\item If $a_t\in A_{\text{tool}}$, a tool $f\in\mathcal{F}$ is executed against the database $\mathcal{D}$. This execution may update the environment state $s_{t+1}^E$ via reads/writes on $\mathcal{D}$ and produce a structured observation
$o_{t+1}^E\in O_E$.
The dialogue state $s_{t+1}^H$ is updated by appending the new tool interaction.
\item If $a_t\in A_{\text{resp}}$, the assistant updates the dialogue state (i.e., $s_{t+1}^H$) by emitting a response.
In interactive settings this may lead to a new user observation $o_{t+1}^H\in O_H$; in offline training it typically yields a termination signal.
The environment state remains unchanged for that turn: $s_{t+1}^E = s_t^E$.
\end{itemize}


\section{Methodology}
\label{sec:method}

We propose \textbf{Agent-World}, a general-purpose agent training arena that unifies scalable environment-task discovery with continuous self-evolving agent training. The method contains two tightly coupled components:

(1) \textbf{Agentic Environment-Task Discovery.} We collect thousands of real-world environment themes and build a deep-research pipeline that autonomously mines topic-aligned databases and executable tool interfaces from the web, forming a scalable and realistic environment ecosystem. On top of these environments, we synthesize diverse verifiable tasks through both graph-based and programmatic generation, and further expand task difficulty with executable validation.

(2) \textbf{Continuous Self-Evolving Agent Training.} We train agents via multi-environment reinforcement learning over ``agent--tool--database'' interaction rollouts, using executable rewards for state-aware supervision. The same environment ecosystem also serves as a dynamic diagnostic arena that refreshes evaluation tasks, identifies capability gaps, and drives targeted environment-task expansion, thereby enabling the co-evolution of agent policies and environments.

These two components form a closed loop: scalable environments support agent training, while training-time diagnosis feeds back into the next round of environment-task construction. Below, we describe each component in detail.

\subsection{Agentic Environment-Task Discovery}\label{section:method-env-discovery}

\begin{figure*}[!t]
\centering
\includegraphics[width=\linewidth]{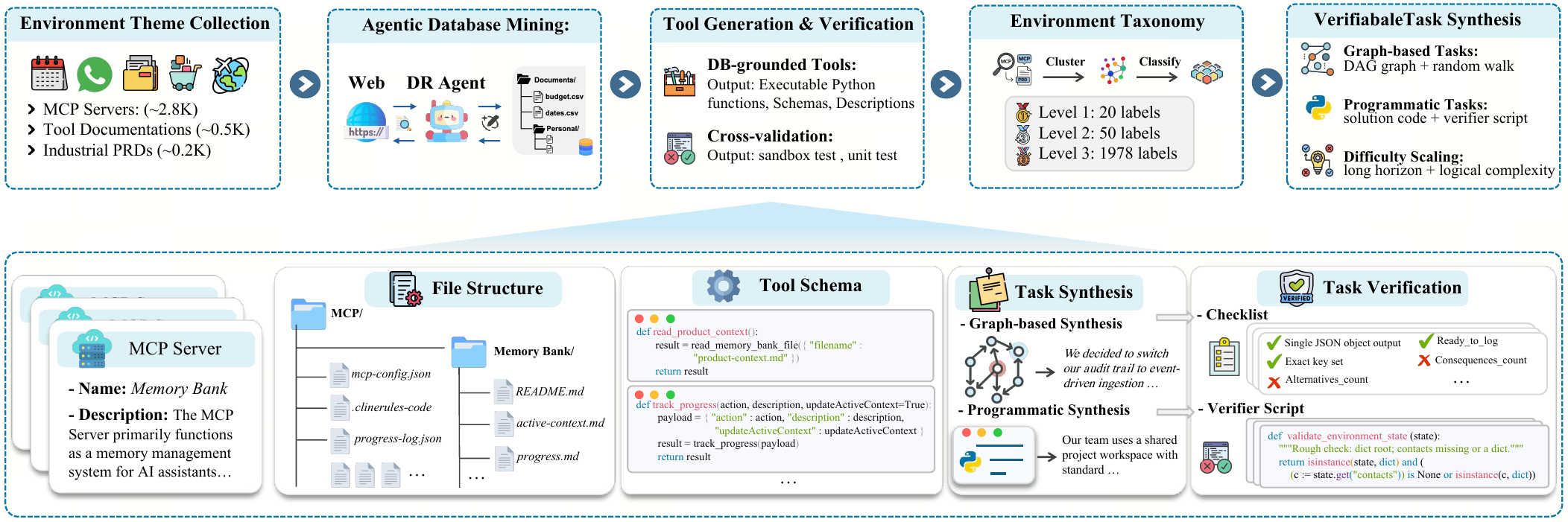}
\caption{
\textbf{The Pipeline of Agentic Environment-Task Discovery.} We start from real-world environment themes, mine topic-aligned databases from the web, generate and verify executable tool interfaces, and synthesize verifiable tasks with controllable difficulty.
}
\label{fig:env-task-discovery}
\end{figure*}

\textbf{Environment Theme Collection:} Scalable environment synthesis begins with diverse and high-quality environment themes as anchors. We therefore systematically gather environment themes from three real-world sources:

\textbf{(1) MCP Servers:} We obtain real-world MCP server specifications from Smithery~\footnote{\url{https://smithery.ai/servers}}. Each specification is accompanied by a structured JSON document that includes source-data descriptions and standardized tool definitions. We denote the corresponding topics as $m\in\mathcal{M}_1$.

\textbf{(2) Tool Documentations:} We broadly collect and filter open-source datasets covering real tool-use scenarios, extract tool-definition documents, and use an LLM to inversely map them to environment topics, denoted as $m\in\mathcal{M}_2$.

\textbf{(3) Industrial PRDs:} As product requirement documents for specific industries, PRDs naturally include background, domain workflows and system interfaces. We use them as theme anchors, denoted as $m\in\mathcal{M}_3$.

We finally merge these sources to form the seed topic set:
$
\mathcal{M}=\mathcal{M}_1\cup \mathcal{M}_2\cup \mathcal{M}_3.
$

\textbf{Agentic Database Mining:} Given the topic set $\mathcal{M}$, our goal is to mine topic-aligned real-world environment databases. Unlike prior work that emphasizes LLM-synthesized databases~\citep{song2026envscaler,guo2025genenv,tu2026scaleenvscalingenvironmentsynthesis}, we argue that the World Wide Web already contains abundant, high-value structured data that can be updated in real time.

Motivated by this, we design an agentic workflow to autonomously mine and process web data into environment databases. Concretely, we build a deep-research agent $\mathcal{G}$ centered on a policy model $\pi_\theta$ and an external toolset $\mathcal{T}$ including \textit{search, browser, code compiler, and operating-system (OS) tools}. For each topic $m\in\mathcal{M}$, the agent conducts iterative loops for in-depth information retrieval and data mining. After that process, the agent leverages OS tools for structuring and persistent storage, yielding the environment database as:
\[
\mathcal{D}(m)=\mathcal{G}(m;\pi_\theta,\mathcal{T}),\quad m\in\mathcal{M}
\]
where $\mathcal{G}(\cdot)$ denotes the topic-conditioned automated research pipeline. Empirically, a single autonomous mining flow often yields databases with limited scale and simple structure. To address this, we introduce a database complexification process $\phi$, which iteratively prompts a deep-research agent to expand and enrich topic-specific databases:
\[
\mathcal{D}^{(n+1)}(m)=\phi\big(\mathcal{D}^{(n)}(m),\,m,\,\mathcal{T}\big),\quad n=0,\dots,N-1,
\]
where the final database denotes $\mathcal{D}^{(N)}(m)$. In practice, repeating this procedure for $N$ rounds produces high-quality databases that better match realistic environment demands.

\begin{figure}[t]
    \centering
    \includegraphics[width=\linewidth]{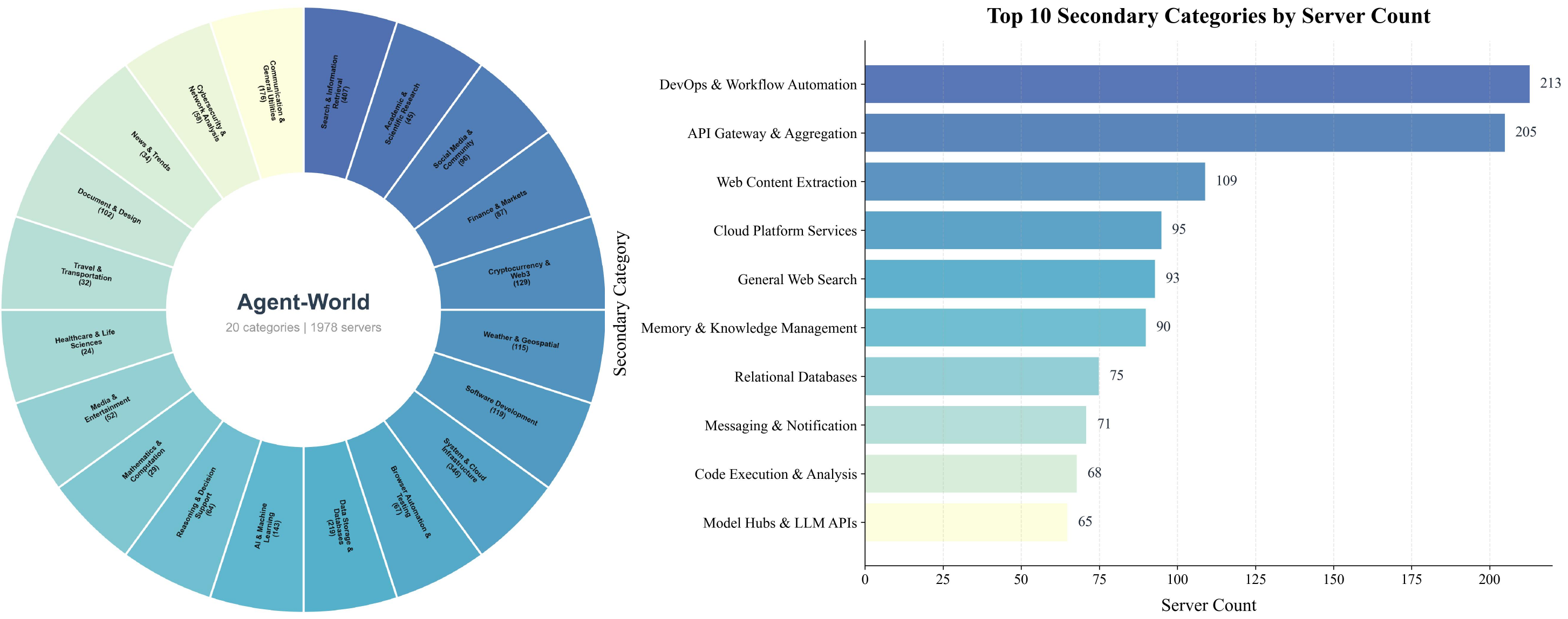}
  \caption{\textbf{Hierarchical environment taxonomy of Agent-World.}
\textbf{Left:} distribution of the 20 first-tier categories with their server counts.
\textbf{Right:} top-10 second-tier categories ranked by server count.}
    \vspace{-1em}
    \label{fig:mcp_taxonomy_method}
\end{figure}

\textbf{Tool Interface Generation and Verification.}
To construct a database-grounded executable toolset, we introduce a coding agent $\psi$ equipped with a \textit{code compiler} and \textit{OS tools}, denoted by $\hat{\mathcal T}$.
Given $(m,\mathcal D^{(N)}(m))$, the agent generates candidate tools together with their unit-test sets:
\[
\big\{(\hat f,\hat{\mathcal C}_{\hat f})\big\}
=\psi(m,\mathcal D^{(N)}(m);\pi_\theta,\hat{\mathcal T}),\quad m\in\mathcal M,
\]
where each tool $\hat f$ is associated with a set of test cases $\hat{\mathcal C}_{\hat f}$ (i.e., a one-to-many mapping).

Motivated by a series of automated execution-based verification procedures~\citep{autoif,zeng2024automatic}, we then perform cross-validation for quality control.
For each candidate tool $\hat f$, its test accuracy is defined as
\[
\mathrm{Acc}(\hat f;\hat{\mathcal C}_{\hat f})
=\frac{1}{\lvert\hat{\mathcal C}_{\hat f}\rvert}
\sum_{\hat c\in\hat{\mathcal C}_{\hat f}}
\mathbf 1[\hat f(\hat c)\ \text{passes}].
\]

A tool is retained only if it satisfies all of the following:
\begin{itemize}[leftmargin=1em]
\item the function can be successfully compiled by the Python compiler;
\item $\mathrm{Acc}(\hat f;\hat{\mathcal C}_{\hat f})>0.5$ on its associated test set;
\item the corresponding environment contains at least one valid tool and one valid test case.
\end{itemize}

After filtering, we obtain the quality-controlled tool set $\mathcal F(m)$.
Finally, we define the scalable environment ecosystem as
\[
\mathcal E=\{(\mathcal D^{(N)}(m),\mathcal F(m))\mid m\in\mathcal M\}.
\]

\textbf{Environment Taxonomy Construction:} 
To systematically organize the synthesized environments, we build a hierarchical taxonomy. Based on thousands of environment themes, we apply hierarchical clustering~\citep{ward1963hierarchical} to obtain 50 cluster centers; we then trace back the sample set covered by each cluster and randomly select representative samples. Building on TOUCAN's taxonomy~\citep{toucan}, we use GPT-OSS-120B~\citep{openai2025gptoss120bgptoss20bmodel} as a supervised summarization model to identify the central environment theme of each cluster, yielding 50 second-tier labels.

Since relying solely on LLM summarization may introduce templated text and bias, we invite three annotators to merge the second-tier labels and abstract them into 20 first-tier types; cross-validation and discussion yield the final hierarchical taxonomy of the environment ecosystem. As shown in Figure~\ref{fig:mcp_taxonomy_method}, the taxonomy contains 20 first-tier labels, 50 second-tier labels, and over 2K third-tier labels, providing a foundation for cross-environment task synthesis and stratified arena construction. We denote the set of first-tier categories by $\mathcal{C}$.







\subsubsection{Verifiable Task Synthesis}\label{section:method-task-synthesis}

After constructing the scalable environment ecosystem $\mathcal{E}$, we synthesize high-quality agentic tasks that simulate diverse real-world tool-use scenarios. To generate complex, long-horizon tasks grounded in reliable execution, we use two complementary synthesis strategies: \textbf{graph-based task synthesis} for modeling sequential tool dependencies, and \textbf{programmatic task synthesis} for modeling complex, non-linear reasoning and control flow. Both approaches rely on sandbox execution to collect execution traces, derive ground-truth answers, and preserve task verifiability~\citep{tu2026scaleenvscalingenvironmentsynthesis, cai2025autoforgeautomatedenvironmentsynthesis}.

\paragraph{\textbf{(1) Graph-Based Task Synthesis.}}

In real-world scenarios, agents often need to invoke a sequence of tools in a specific logical order to accomplish a goal. Thus, a valid tool execution sequence and its returned results naturally define the underlying information requirements and data flows needed to answer a specific user query. Building on this insight, we adopt a reverse-engineering paradigm: we first synthesize a valid tool-call sequence and then generate the corresponding task description~\citep{toucan}. To ensure task rationality and diversity, we build connected tool graphs and walk on the graph to obtain the tool sequences. We detail the construction process as follows: 

\textbf{Tool Graph Construction.} For each environment $(\mathcal{D}^{(N)}(m), \mathcal{F}(m)) \in \mathcal{E}$, we first construct a fully connected, weighted directed graph $G = (V, E)$, where each node $v\in V$ corresponds to a tool $f\in\mathcal{F}(m)$ and each edge encodes call dependencies between tools. We define three types of edges, evaluated and assigned by an LLM:
\begin{itemize}[leftmargin=1em]
    \item \textbf{Strong dependency ($f_i \to f_j$, $w_{ij}=3$):} The input of tool $f_j$ strictly relies on the output of tool $f_i$ (e.g., calling \texttt{create\_order} to obtain an \texttt{order\_id} before calling \texttt{get\_order\_details}). This forms a strictly directed edge, ensuring the most logical data flow.
    \item \textbf{Weak dependency ($f_i \leftrightarrow f_j$, $w_{ij}=2$):} The input of $f_j$ \textit{can} be derived from $f_i$'s output, but can also be obtained via other means (e.g., querying a database directly or using a constant). This is modeled as a bidirectional edge, offering flexibility during the walk.
    \item \textbf{Independent edge ($f_i \leftrightarrow f_j$, $w_{ij}=1$):} Tools with no parameter-level dependencies. These edges act as a fallback to guarantee that $G$ is fully connected, preventing dead ends during random walks.
\end{itemize}

\textbf{Random Walk on Tool Graph.} We generate a raw tool-call sequence $\tau = [f_1, f_2, \dots, f_k]$ by performing a random walk on $G$. We prioritize starting nodes $f_1$ that return tool output but have no strong dependency precursors. At step $t$, the next tool $f_{t+1}$ is sampled from the successors of $f_t$ with a probability distribution biased by the edge weights $w$, encouraging sequences with realistic reasoning. 

Once the tool sequence $\tau$ is sampled, we instantiate its input parameters: (1) For strong/weak dependencies, we pass the output of the preceding tool; (2) For independent edges, we randomly sample valid values from the database $\mathcal{D}^{(N)}(m)$. Finally, an LLM reviews the populated chain to prune redundancies, verify logical consistency, and output a refined, executable tool sequence $\tau^*$.

\textbf{Task and Rubric Generation.} Given $\tau^*$, an LLM drafts an initial task description $q_{init}$. To prevent data leakage, $q_{init}$ is strictly prohibited from containing technical details such as tool names or database schema. Next, we execute $\tau^*$ step-by-step within a Python sandbox, recording the intermediate execution trace and the final return results. Observing the actual data fields and formats allows the LLM to refine $q_{init}$ into a highly realistic and well-grounded final query $q_{final}$. Simultaneously, the LLM generates a strictly formatted JSON ground-truth answer $a^*$ and structured evaluation rubrics $R$ \citep{funreason-mt,selfrubric,rubric_achors,drtulu}. The rubrics $R$ enable automated evaluation across multiple dimensions, including field completeness, schema matching, and numerical tolerances.

\textbf{Quality Consistency and Verification.} To ensure task stability, we evaluate the generated task $(q_{final}, a^*)$ by deploying a ReAct agent to solve it 5 separate times within the sandbox. We retain the task only if the agent successfully reaches a consistent answer in at least two independent runs.

\textbf{Difficulty Scaling.} To increase task difficulty while maintaining solvability, we complicate the reasoning path in each task. Specifically, we scale difficulty by increasing the maximum step count of the random walk to expand the tool chain, and by increasing the sampling probability of weak dependencies and independent edges to reduce reliance on obvious sequential outputs. We further rewrite the final task description to obscure explicit mentions of tool names and execution logic, forcing the agent to infer the required workflow purely from abstract task goals.
The final task set synthesized by graph-based generation is denoted as $\mathcal{X}_{\text{graph}}$.

\paragraph{\textbf{(2) Programmatic Task Synthesis.}}

While graph-based synthesis effectively models sequential dependencies, real-world tasks often demand reasoning patterns that cannot be expressed linearly, such as conditional tool usage, multi-step loops, and result aggregation. To capture these behaviors, we introduce programmatic task synthesis. Unlike the graph-based method that simulates step-by-step sequences, this approach directly generates executable Python solutions capable of performing code-based reasoning over provided tools.

\textbf{Task and Solution Code Generation.} We prompt an LLM with the environment's tool schemas and database descriptions to generate a highly complex task query $q_{prog}$. The query must focus entirely on task scenarios and objectives without revealing details of tools or databases. Subsequently, the LLM acts as a solver to generate a comprehensive, end-to-end executable Python script $\pi_{code}$. This script must load the tool implementations and utilize complex control flows (e.g., \texttt{for} loops, \texttt{if-else} branches, statistical aggregations) to solve $q_{prog}$. To ensure $\pi_{code}$ is executable, we wrap this step in a ReAct loop: if the sandbox throws syntax or runtime errors, the agent iteratively debugs and repairs the code. The successfully executed script yields the final ground-truth answer $a^*$.

\textbf{Verification Code Generation.} Traditional string-matching evaluation falls short for complex programmatic tasks. Therefore, we input $(q_{prog}, \pi_{code}, a^*)$ to an LLM to generate an executable verification script $V_{code}(a, a^*)$. 
The script includes multi-level assertions and custom logic to robustly determine whether the candidate answer $a$ and the underlying database state $s^E$ satisfy all task constraints. Similar to solution code generation, a ReAct agent debugs $V_{code}$ in the sandbox to guarantee its reliability.

\begin{figure*}[!t]
\centering
\includegraphics[width=\linewidth]{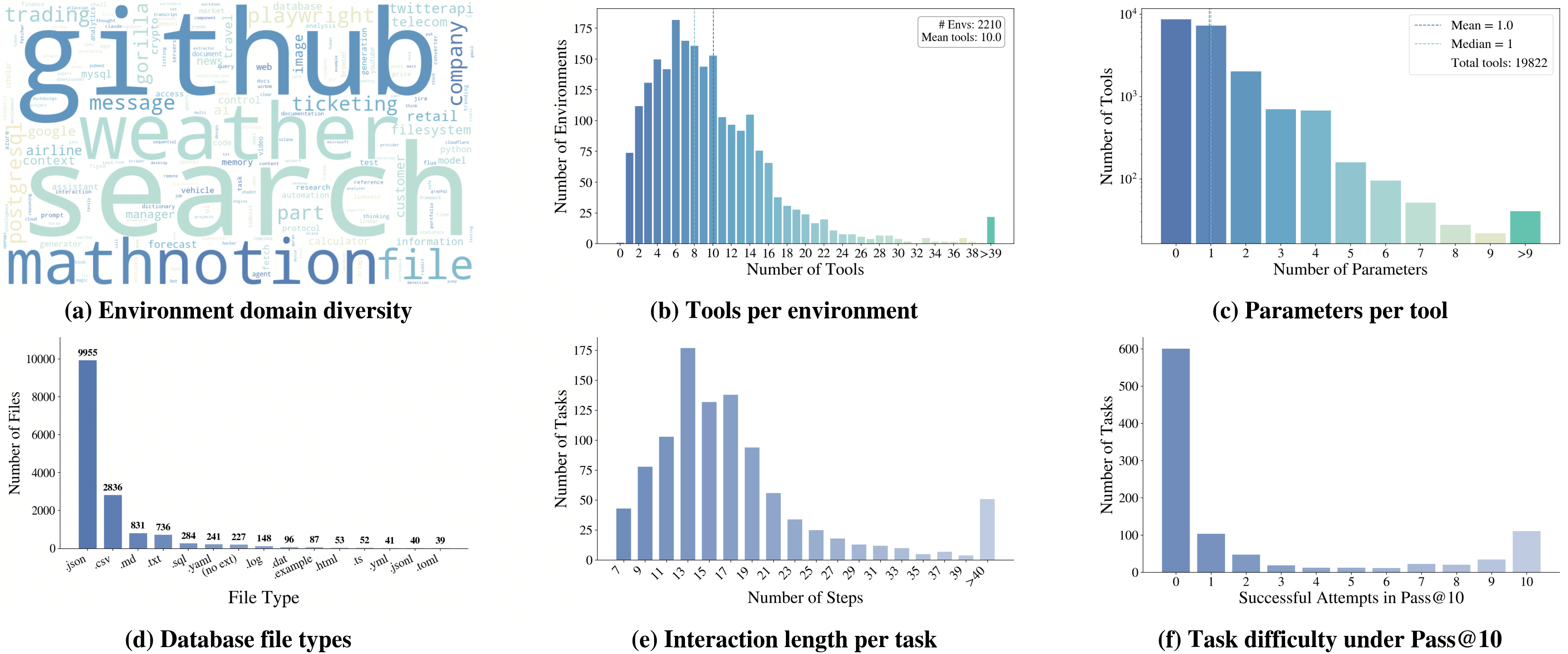}

\caption{
\textbf{Comprehensive statistics of Agent-World environments and synthesized tasks}, including environment diversity, tool coverage, file-type distribution, and task difficulty characteristics.}
\label{fig:env_stastic}
\end{figure*}

\textbf{Quality Consistency and Verification.} Following the same rigorous filtering protocol as the graph-based method, we execute a ReAct agent 5 times against $q_{prog}$. The generated verification script $V_{code}$ evaluates the agent's output. Tasks are preserved only if the agent achieves a stable pass rate (at least 2 successful runs), ensuring the synthesized tasks are challenging yet solvable.

\textbf{Difficulty Scaling.} Similar to graph-based synthesis, we also scale the difficulty of programmatic tasks. Specifically, we increase the number of unique tools and invocations through modifying LLM instructions. We also inject instructions of implementing intricate inter-tool logic such as conditional branches and mandate advanced data operations like cross-database aggregations, sorting, and filtering. Finally, similar to the graph-based approach, we rewrite the task description of any direct references to APIs or execution traces, ensuring the agent must plan complex programmatic logic entirely from high-level user intents.
The final task set synthesized by programmatic generation is denoted as $\mathcal{X}_{\text{prog}}$.

\textbf{(3) Static Statistics of Environment--Task Data:}
To more comprehensively demonstrate the quality of our agentic environment scaling stage, Figure~\ref{fig:env_stastic} provides a detailed analysis of Agent-World environments and tasks through six subfigures.

\textbf{Environment Diversity:}
We observe that \textbf{(a)} Agent-World covers a broad range of environment types, with over 2,000 environments in total (1,978 retained after filtering).
\textbf{(b)} Each environment is equipped with a diverse toolset, averaging more than 10 tools, with some environments containing over 40 tools.
\textbf{(c)} The overall ecosystem includes 19,822 distinct tools, each with rich parameters, ensuring both atomic functionality and tool diversity.
Interestingly, \textbf{(d)} the underlying database file types are also highly diverse, including \texttt{json}, \texttt{csv}, \texttt{sql}, and \texttt{html}, as well as environment-specific formats such as \texttt{tex} and \texttt{yaml}. This further reflects the diversity of our databases and their alignment with real-world workspace file formats.

\textbf{Task Difficulty:}
As shown in \textbf{(e)}, all synthesized tasks contain at least 7 interaction turns, with an average of over 20 turns and a non-trivial portion exceeding 40 turns, already indicating substantial difficulty.
To quantify difficulty more directly, in \textbf{(f)} we evaluate task execution under Pass@10 using the strong proprietary model Doubao-Seed-2.0-pro~\citep{seed2}. Only a small fraction of tasks are solved in all 10 attempts; most are solved only once out of 10, and some are not solved at all. This shows that our difficulty scaling strategy is effective at increasing task complexity.

Beyond aggregate statistics, we provide reader-facing \emph{environment cards} in Appendix~\ref{sec:appendix-env-visualizations}, where each card summarizes a seed domain's on-disk layout and representative callable tool interfaces.
In addition, Appendix~\ref{sec:case-study} presents \emph{verifiable tasks}, including the environment, tools, rubrics, and interaction trajectories.

\subsection{Continuous Self-Evolving Agent Training}\label{sec:self-evolving-training}

In this section, we introduce continuous self-evolving agent training. Given a scalable environment ecosystem $\mathcal{E}=\{(\mathcal{D}^{(N)}(m),\mathcal{F}(m))\mid m\in\mathcal{M}\}$, where each environment pairs a database with an executable toolset, we train general-purpose agents with multi-environment ``agent--tool--database'' interaction rollouts and executable rewards. Crucially, $\mathcal{E}$ also serves as a dynamic diagnostic arena: the current policy is evaluated on fresh tasks in held-out environments, its capability gaps are identified from executable evidence, and the resulting diagnosis guides targeted environment-task expansion. This creates a self-evolving loop in which agent policies and environments co-evolve over training rounds.

\subsubsection{Multi-Environment Agent Reinforcement Learning} \label{sec:me-rl}
After constructing a scalable environment ecosystem and synthesizing verifiable tasks, we perform multi-environment agent RL to improve state-aware reasoning, long-horizon tool use, and environment interaction robustness.

\textbf{Multi-environment Rollout.} Unlike static tool-calling scenarios, we implement a closed-loop interaction among three components:
\begin{itemize}[leftmargin=1em]
\item \textbf{An LLM policy} $\pi_\theta$, which generates the next action conditioned on the dialogue history and tool feedback;
\item \textbf{A tool interface/runtime}, which executes the environment-specific tool set $\mathcal{F}(m)$ and maintains environment-side states (database connections, caches, etc.);
\item \textbf{A database state} $\mathcal{D}^{(N)}(m)$, which serves as the read/write substrate for tool execution and provides a verifiable, updatable structured data backbone.
\end{itemize}
At each step, the model produces both natural-language reasoning and tool/action decisions. When a tool call is triggered, the interface executes the selected tool in a sandboxed environment to read or update the environment database state, and returns structured observations to the policy for subsequent decision making. 

Formally, given a task $x$ and its training environment $(\mathcal{D}^{(N)}(m),\mathcal{F}(m))\in\mathcal{E}$, following Section~\ref{sec:preliminary}, the policy $\pi_\theta$ samples an action $a_t$ based on the instruction and history $h_t=(o_0,a_0,\dots,o_t)$. If $a_t\in A_{\text{tool}}$, it executes $f\in\mathcal{F}(m)$ on $\mathcal{D}^{(N)}(m)$ and returns a structured observation $o_{t+1}^{E}\in O_E$; if $a_t\in A_{\text{resp}}$, it outputs a natural-language response (typically the final answer or completion marker) and terminates the trace. This yields a model output $y=(\tau,a_{\text{final}})$, where $\tau=(o_0,a_0,\ldots,o_T,a_T)$ is the interaction trajectory and $a_{\text{final}}$ is the final answer. Following Group Relative Policy Optimization (GRPO)~\citep{grpo}, we sample $N$ outputs per task $x$, and tasks within each global batch are paired with independent and dynamic environments to realize multi-environment rollouts.

\textbf{Structured Verifiable Reward.} Reward signals define the optimization objective and directly guide policy behavior. Distinct from prior static tool-RL settings~\citep{retool,dong2025toolstar}, automatic reward assignment for environment agents must account for multiple factors beyond answer correctness, including environment state, efficiency constraints, and format compliance. Accordingly, we instantiate two reward types.

(i) \textbf{Graph-based tasks} ($\mathcal{X}_{\text{graph}}$) provide a structured rubric $R=\{r_j\}_{j=1}^{n}$ (schema matching, fact checking, etc.). We use a rubric-conditioned LLM-as-judge to evaluate each criterion $r_j$ from model output $y$ under task $x$, and compute an overall pass rate by averaging criterion-level pass indicators.

(ii) \textbf{Programmatic tasks} ($\mathcal{X}_{\text{prog}}$) provide an executable validation script $V_{\text{code}}$ per task, which we run in the sandbox to verify either the predicted answer or the resulting database state. Therefore, the output-level reward is computed as
\[
r(x,y)=
\begin{cases}
\mathbb{I}\Big[\frac{1}{n}\sum_{j=1}^{n}\mathbb{I}\big[\mathrm{Judge}(x,y,r_j)\big] == 1\Big], & \text{if } x\in \mathcal{X}_{\text{graph}}, \ r_j \in R\\
\mathbb{I}\big[\mathrm{Execute}(V_{\text{code}}(y,y^{*}))\big], & \text{if } x\in \mathcal{X}_{\text{prog}}.
\end{cases}
\]

where \(\mathbb{I}[\cdot]\) is the indicator function. \(\mathrm{Judge}(x,y,r_j)\) denotes a rubric-conditioned LLM judge that assesses whether model output \(y\) satisfies criterion \(r_j\) under task \(x\). \(\mathrm{Execute}(V_{\text{code}}(y,y^{*}))\) denotes running the task-specific validation script \(V_{\text{code}}\) in a sandbox over model output \(y\) to verify that answer/state are satisfied with the ground truth $y^{*}$.

\begin{figure*}[!t]
\centering
\includegraphics[width=\linewidth]{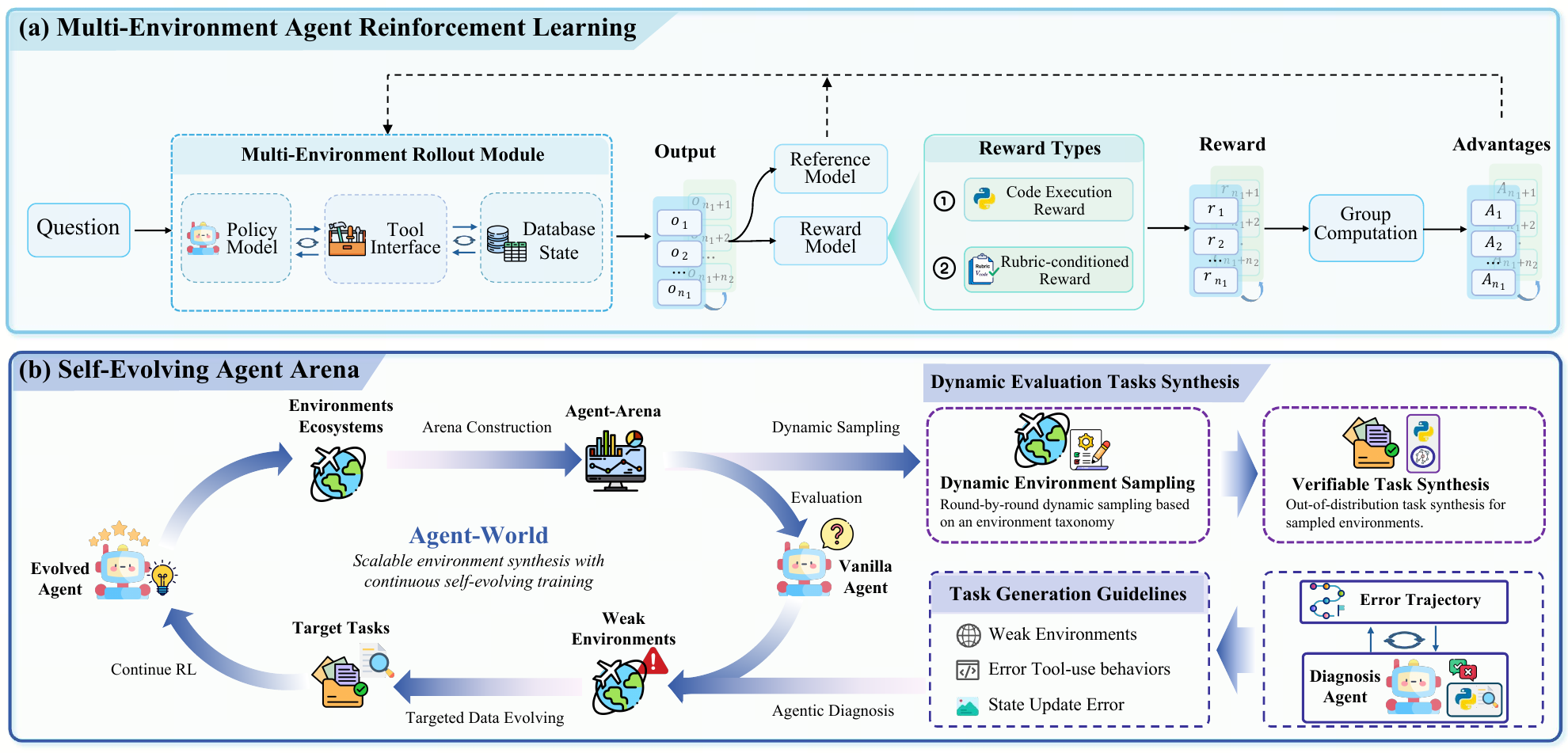}

\caption{
\textbf{The Overall Framework of Continuous Self-Evolving Agent Training.} The agent is trained with multi-environment RL under executable rewards (top), evaluated in a dynamic arena, diagnosed for capability gaps, and improved through targeted environment-task expansion (bottom).
}
\label{fig:training}
\end{figure*}

\textbf{Policy Update.} To enable stable training with environment interaction, we adopt Group Relative Policy Optimization (GRPO)~\citep{grpo} to directly maximize the verifiable returns defined above. Concretely, for each input task $x$ sampled from dataset $D$, we draw a group of $G$ trajectories/outputs $\{y_i\}_{i=1}^{G}$ from the behavior policy $\pi_{\theta_{\text{old}}}(\cdot\mid x)$, compute token-level advantages $\hat{A}_{i,t}$, and update $\pi_\theta$ by maximizing the GRPO objective with a clipped importance ratio and a KL penalty to a reference policy $\pi_{\text{ref}}$:
\begin{equation}
\begin{split}
J_{\text{GRPO}}(\theta) = \mathbb{E}_{x \sim D, \{y_i\}_{i=1}^G \sim \pi_{\theta_{\text{old}}}(\cdot \mid x)} \Bigg[ &\frac{1}{G} \sum_{i=1}^G \frac{1}{|y_i|} \sum_{t=1}^{|y_i|} \min \Big( r_{i,t}(\theta) \hat{A}_{i,t}, \\
& \text{clip} \left( r_{i,t}(\theta), 1 - \epsilon, 1 + \epsilon \right) \hat{A}_{i,t} \Big) - \beta D_{\text{KL}}(\pi_{\theta} \parallel \pi_{\text{ref}}) \Bigg].
\end{split}
\label{eq:grpo}
\end{equation}
where \(\epsilon\) and \(\beta\) are hyperparameters, \(y_i\) denotes the model output (including interaction trajectory and final answer), and $\hat{A}_{i,t}$ is the normalized advantage of the \(i\)-th rollout within the group.

\subsubsection{Self-Evolving Agent Arena}\label{sec:self-evolving-arena}

\textbf{Motivation.}
Our scalable environment ecosystem $\mathcal{E}=\{(\mathcal{D}^{(N)}(m),\mathcal{F}(m))\mid m\in\mathcal{M}\}$ serves not only as a training source but also as an agentic diagnostic arena. Beyond synthesizing training data, we aim to continuously identify weaknesses of the current agent policy and then expand environments and tasks in a targeted manner to close those gaps. This yields a self-reinforcing loop in which evaluation, diagnosis, and data generation evolve together with the agent.

\textbf{Arena Construction.}
Based on the hierarchical environment taxonomy (Sec.~\ref{section:method-env-discovery}), we construct an evaluation arena by stratified sampling.
Specifically, for each first-tier category $c\in\mathcal{C}$, we randomly select $K$ environments ($K=5$) and merge them into the arena set
$\mathcal{E}_{\text{arena}}=\{(\mathcal{D}^{(N)}(m_i),\mathcal{F}(m_i))\}_{i=1}^{|\mathcal{E}_{\text{arena}}|}$.
This design ensures broad coverage over different environment types while keeping evaluation cost controllable.

\textbf{Dynamic Evaluation Task Synthesis.}
For each arena environment $(\mathcal{D}^{(N)}(m_i),\mathcal{F}(m_i))\in\mathcal{E}_{\text{arena}}$, we follow Section~\ref{section:method-task-synthesis} and synthesize a fresh batch of verifiable tasks and validators at each iteration.
Concretely, at iteration $r$ we instantiate a task set
$\mathcal{X}^{(r)}_{\text{arena}}(m_i)$ consisting of both graph-based tasks and programmatic tasks, each paired with an executable rubric $R$ or verification code $V_{\text{code}}$.
The full evaluation set is defined as $\mathcal{X}^{(r)}_{\text{arena}}=\bigcup_i \mathcal{X}^{(r)}_{\text{arena}}(m_i)$.
Importantly, both the sampled environments and the synthesized tasks are dynamic across rounds, preventing overfitting to a static evaluation and enabling continual diagnosis.

\textbf{Agentic Diagnosis.}
Given a trained agent policy $\pi_{\theta^{(r)}}$, we evaluate it on synthesized tasks $\mathcal{X}^{(r)}_{\text{arena}}$ under the \texttt{agent-tool-database} execution protocol, with task-level assessment performed by the corresponding executable rubric $R$ or verification code $V_{\text{code}}$.

We then employ an auto-diagnosis agent $\delta$, equipped with a Python interpreter and search tools, to analyze failure patterns. The diagnosis agent takes as input:
\textbf{(i)} per-task failure traces (tool logs, intermediate observations, and validator feedback),
\textbf{(ii)} error distribution statistics by environment and taxonomy category, and
\textbf{(iii)} environment metadata (tool schemas and database descriptions).

The diagnosis agent outputs \textbf{(a)} a ranked set of weak environments
$\mathcal{W}^{(r)}\subseteq \mathcal{E}_{\text{arena}}$ and \textbf{(b)} environment-specific task-generation guidelines
$\mathcal{G}^{(r)}_{\text{guide}}(m)$ that characterize missing capabilities (e.g., erroneous tool use or state-update mistakes).
These outputs serve as anchors for subsequent environment and task expansion.
Detailed prompts for agentic diagnosis are provided in Appendix~\ref{sec:diagnosis-prompt}.

\textbf{Agent-Environment Co-Evolution.}
Conditioned on $\mathcal{W}^{(r)}$ and $\mathcal{G}^{(r)}_{\text{guide}}=\{\mathcal{G}^{(r)}_{\text{guide}}(m_i) | {(\mathcal{D}^{(N)}(m_i),\mathcal{F}(m_i))\in\mathcal{W}^{(r)}}\}$, we re-run the verifiable task synthesis pipeline (Sec.~\ref{section:method-task-synthesis}) to generate a targeted training set
$\mathcal{X}^{(r)}_{\text{target}}$,
optionally accompanied by environment expansion via database complexification when the weakness is due to insufficient state diversity.

Starting from $\pi_{\theta^{(r)}}$, we then perform multi-environment agent RL (Sec.~\ref{sec:me-rl}) on the augmented data to obtain an improved policy $\pi_{\theta^{(r+1)}}$.
Iterating the above steps yields a self-evolving agent-arena loop:
\[
\pi_{\theta^{(r)}} \xrightarrow{\text{evaluate}}
\mathcal{W}^{(r)} \xrightarrow{\text{diagnose+target}}
\mathcal{X}^{(r)}_{\text{target}} \xrightarrow{\text{continue RL}}
\pi_{\theta^{(r+1)}}.
\]

This arena-driven loop turns scalable environments into an automated curriculum engine, continuously driving targeted learning and enabling the co-evolution of agent policies and environments.

\begin{algorithm}[t]
\caption{\textbf{Self-Evolving Agent Arena Loop}}
\label{alg:self-evolving-arena}
\KwIn{agent environment arena $\mathcal{E}_{\text{arena}}\subset\mathcal{E}$; initial policy $\pi_{\theta^{(0)}}$; number of evolving rounds $R$}
\KwOut{Evolved policy $\pi_{\theta^{(R)}}$}
\For{$r=0,\dots,R-1$}{
    \tcp{\textbf{Phase 1: Dynamic Evaluation Task Synthesis}}
\ForEach{$(\mathcal{D}^{(N)}(m_i),\mathcal{F}(m_i))\in\mathcal{E}_{\text{arena}}$}{
        Synthesize fresh verifiable tasks $\mathcal{X}^{(r)}_{\text{arena}}(m_i)$ with executable rubric $R$ or verification code $V_{\text{code}}$;
    }
    Define the full evaluation set $\mathcal{X}^{(r)}_{\text{arena}}=\bigcup_{i}\mathcal{X}^{(r)}_{\text{arena}}(m_i)$\;
    Evaluate $\pi_{\theta^{(r)}}$ on $\mathcal{X}^{(r)}_{\text{arena}}$ under \texttt{agent-tool-database} execution with assessment by $R$ or $V_{\text{code}}$\;

    \tcp{\textbf{Phase 2: Agentic diagnosis}}
    Input per-task failure traces, environment error statistics and metadata to diagnosis agent $\delta$\;
    outputs weak environments $\mathcal{W}^{(r)}\subseteq\mathcal{E}_{\text{arena}}$ and task-generation guidelines $\mathcal{G}^{(r)}_{\text{guide}}(m)$\;

    \tcp{\textbf{Phase 3: Agent-Environment Co-Evolution.}}
    \ForEach{$(\mathcal{D}^{(N)}(m),\mathcal{F}(m))\in\mathcal{W}^{(r)}$}{
        Complexify database: $\mathcal{D}^{(N)}(m)\leftarrow\phi(\mathcal{D}^{(N)}(m),\cdot)$\;
        Generate targeted tasks $\mathcal{X}^{(r)}_{\text{target}}(m)$ conditioned on $\mathcal{G}^{(r)}_{\text{guide}}(m)$\;
    }
    Define $\mathcal{X}^{(r)}_{\text{target}}=\bigcup_{(\mathcal{D}^{(N)}(m),\mathcal{F}(m))\in\mathcal{W}^{(r)}}\mathcal{X}^{(r)}_{\text{target}}(m)$\;
    Continue RL on $\mathcal{X}^{(r)}_{\text{target}}$ obtain: $\pi_{\theta^{(r+1)}}\leftarrow\pi_{\theta^{(r)}}$\;
}
\Return{$\pi_{\theta^{(R)}}$}\;
\end{algorithm}

\fi
\section{Experiment}

In this section, we conduct experiments to evaluate the effectiveness of Agent-World and further analyze its key properties.
First, we introduce the details of the experimental settings (Sec.~\ref{sec:experiment_settings}).
Next, we present the main results of Agent-Wolrd (Sec.~\ref{sec:main_results}).
Finally, we present quantitative and qualitative analyses of our approach (Sec.~\ref{sec:detailed_analysis}).

\subsection{Experimental Settings}
\label{sec:experiment_settings}

In this part, we introduce the datasets used for training and evaluation, the baseline approaches, and the implementation details.

\paratitle{Baselines.}
We compare Agent-World against three baseline groups, consistent with Table~\ref{tab:main_results}:
\begin{itemize}
\item \textbf{Frontier Proprietary Models:} GPT-5.2 High~\citep{gpt52}, Claude Sonnet-4.5~\citep{claude45}, Gemini-3 Pro~\citep{gemini3pro}, Seed2.0~\citep{seedseed2}.

\item \textbf{Open-Source Foundation Models (8B–685B):}
DeepSeek-V3.2-685B~\citep{liu2025deepseek}, GPT-OSS-120B~\citep{openai2025gptoss120bgptoss20bmodel}, Qwen3-235B-A22B~\citep{qwen3}, and Qwen3-8B, 14B, 32B~\citep{qwen3}.
\item \textbf{Open-Source Environment Scaling Methods (7B-14B):} Simulator-8B~\citep{li2025simulating}, TOUCAN-7B~\citep{toucan}, 
EnvScaler-8B~\citep{song2026envscalerscalingtoolinteractiveenvironments}, AWM-8B, 14B~\citep{wang2026agentworldmodelinfinity}, and ScaleEnv-8B~\citep{tu2026scaleenvscalingenvironmentsynthesis}.
\end{itemize}

\paratitle{Evaluation Benchmarks.}
We evaluate Agent-World on \textbf{23 benchmarks} spanning complementary capabilities:
\begin{itemize}
\item \textbf{Core agentic tool-use suites:} MCP-Mark~\citep{wu2025mcpmarkbenchmarkstresstestingrealistic}, BFCL V4~\citep{patil2025the}, and $\tau^2$-Bench~\citep{barres2025tau2benchevaluatingconversationalagents}.

\item \textbf{Advanced AI assistant benchmarks:} SkillsBench~\citep{li2026skillsbenchbenchmarkingagentskills}, ARC-AGI-2~\citep{chollet2026arcagi2newchallengefrontier}, and Claw-Eval~\citep{claweval}.

\item \textbf{General reasoning benchmarks:} MATH500~\citep{lightman2023letsverifystepstep}, GSM8K~\citep{cobbe2021trainingverifierssolvemath}, MATH~\citep{hendrycks2021measuringmathematicalproblemsolving}, AIME24~\citep{aime24}, AIME25~\citep{aime25}, KOR-Bench (Cipher)~\citep{ma2025korbenchbenchmarkinglanguagemodels}, and OlympiadBench ($OE\_TO\_maths\_en\_COMP$)~\citep{he2024olympiadbenchchallengingbenchmarkpromoting}.

\item \textbf{Agentic search \& coding benchmarks:} WebWalkerQA~\citep{2501_WebWalker}, SWE-Bench Verified (SWE)~\citep{jimenez2024swebenchlanguagemodelsresolve}, SWE-bench Multilingual~\citep{zan2025multiswebenchmultilingualbenchmarkissue}, Terminal-Bench 1.0, Terminal-Bench 2.0~\citep{merrill2026terminalbenchbenchmarkingagentshard}, General AI Assistants (GAIA)~\citep{GAIA}, and Humanity's Last Exam (HLE)~\citep{HLE}.

\item \textbf{Knowledge and MCP benchmarks:} MMLU~\citep{hendrycks2021measuringmassivemultitasklanguage}, SuperGPQA~\citep{2502_SuperGPQA}, MCP-Universe 5 sub-domains (Financial Analysis, Browser Automation, Web Searching, Location Navigation, and Repository Management)~\citep{luo2025mcp}.

\end{itemize}

All baselines and benchmarks are evaluated using in-house evaluation framework, with results aligned to official scores. Following prior work~\citep{deepagent,dong2025arpo,dong2025toolstar,hira,li2026omnigaia}, we use sampled subsets for some benchmarks (e.g., GAIA and HLE) to accelerate evaluation.


\paratitle{Implementation Details.}
In Agentic Environment-Task Discovery, we use GPT-OSS-120B~\citep{openai2025gptoss120bgptoss20bmodel} as the policy model for environment mining. The same policy model is also used for task synthesis and for generating code and rubric artifacts across different toolsets. In Agentic Diagnosis, GPT-OSS-120B is likewise used to execute diagnosis trajectories and identify failure modes. For training initialization, we perform a cold-start supervised fine-tuning stage using the same data-synthesis strategy as Agentic Environment-Task Discovery, where 40K trajectories are generated by an in-house Doubao-Seed-1.8 policy version model~\citep{seed2026seed18modelcardgeneralized}.

After cold-start SFT, we initialize the Qwen3-8B/14B backbones~\citep{qwen3}, synthesize 5K RL samples, and apply GRPO~\citep{grpo} as the RLVR algorithm for subsequent training.
To enhance training stability, we follow prior work~\citep{dapo} and set the clip ratio $\varepsilon_\text{low}=0.2$ and $\varepsilon_\text{high}=0.28$.
Moreover, the maximum trajectory 
length is set to 80K tokens, and the maximum
generation length per step is capped at 32k tokens.
In each training step, we sample 32 tasks and perform 8 rollouts to collect RLVR experience, with $\texttt{temperature}=1.0$ and $\texttt{top\_p}=1.0$.
For evaluation, we also use $\texttt{temperature}=1.0$ and $\texttt{top\_p}=1.0$ for decoding.
To reduce random variance, we repeat each experiment eight times and report average accuracy (\%).

\subsection{Main Results}
\label{sec:main_results}

The experimental results are shown in Table~\ref{tab:main_results}. Overall, Agent-World consistently outperforms existing environment-scaling baselines across diverse agentic tool-use benchmarks, demonstrating stronger robustness and more comprehensive generalization in long-horizon settings. We summarize the main findings as follows.

(1) \textbf{Foundation models remain limited in complex agentic tool-use scenarios.}
Even advanced proprietary models show clear limitations on challenging benchmarks. For instance, GPT-5.2 High achieves only 53.1\% on MCP-Mark, while Gemini-3 Pro reaches 50.8\%. Moreover, open-source foundation models are even more constrained, with GPT-OSS-120B and Qwen3-235B-A22B scoring only 4.7\% and 5.8\% on MCP-Mark. Since these benchmarks cover diverse stateful environments, the results suggest that current foundation models still struggle with long-horizon tool use requiring multi-step planning, tool orchestration, and state tracking.

(2) \textbf{Existing environment-scaling methods still suffer from uneven capability gains.}
Compared with the Qwen3 backbones, existing environment-scaling methods improve some benchmarks, but their gains remain uneven across environments. Simulator-based methods such as Simulator-8B achieve good results on $\tau^2$-Bench, yet still perform poorly on MCP-Mark and BFCL V4, suggesting that simulated environments are insufficient to capture complex real-world state transitions. programmatic environment-scaling methods such as EnvScaler-8B and AWM-8B/14B provide broader gains, but still show clear weaknesses on specific environments, including GitHub and Notion. This highlights that robust generalization depends not only on realistic feedback, but also on the diversity and quality of synthesized environments.

(3) \textbf{Agent-World achieves more consistent cross-environment generalization.}
Under the same training setting, Agent-World consistently outperforms prior environment-scaling baselines across all three benchmark suites. In detail, Agent-World-8B achieves 61.8\% on $\tau^2$-Bench, 51.4\% on BFCL V4, and 8.9\% on MCP-Mark. These results clearly outperform EnvScaler-8B, ScaleEnv-8B and even Qwen3-235B-A22B. 

Moreover, Agent-World-14B achieves an additional improvement of about 5\% over Agent-World-8B. It not only surpasses all prior environment-scaling baselines, but also delivers competitive performance against large open-source LLMs, particularly DeepSeek-V3.2-685B on BFCL-V4 (55.8\% vs.\ 54.1\%). These results indicate that Agent-World produces more consistent gains across diverse benchmarks and environments. We attribute this advantage to its unified framework, which tightly integrates scalable environment-task discovery with continuous self-evolving agent training.

\subsection{Quantitative and Qualitative Analyses}
\label{sec:detailed_analysis}

\begin{table*}[!t]
  \centering
  \caption{
      \textbf{Main results on agentic tool-use benchmarks.}
      We report accuracy (\%) across three benchmark suites: MCP-Mark, BFCL V4, and $\tau^2$-Bench.
      In the \textit{Open-Source Environment Scaling Methods} block, the best result in each column is marked in \textbf{bold} and the second best is \underline{underlined}. 
  }
  \label{tab:main_results}
  \setlength\tabcolsep{3pt}
  \fontsize{10pt}{15.5pt}\selectfont
  \renewcommand{\arraystretch}{1.2}
  \resizebox{\textwidth}{!}{%
  \begin{tabular}{>{\raggedright\arraybackslash}p{3.9cm} cccccc cccccccc cccc}
  \toprule
  \multirow{2}[2]{*}{\textbf{Method}} &
  \multicolumn{6}{c}{\textbf{MCP-Mark}} &
  \multicolumn{8}{c}{\textbf{BFCL V4}} &
  \multicolumn{4}{c}{\textbf{$\tau^2$-Bench}} \\
  \cmidrule(lr){2-7} \cmidrule(lr){8-15} \cmidrule(lr){16-19}
   & \mcpicon{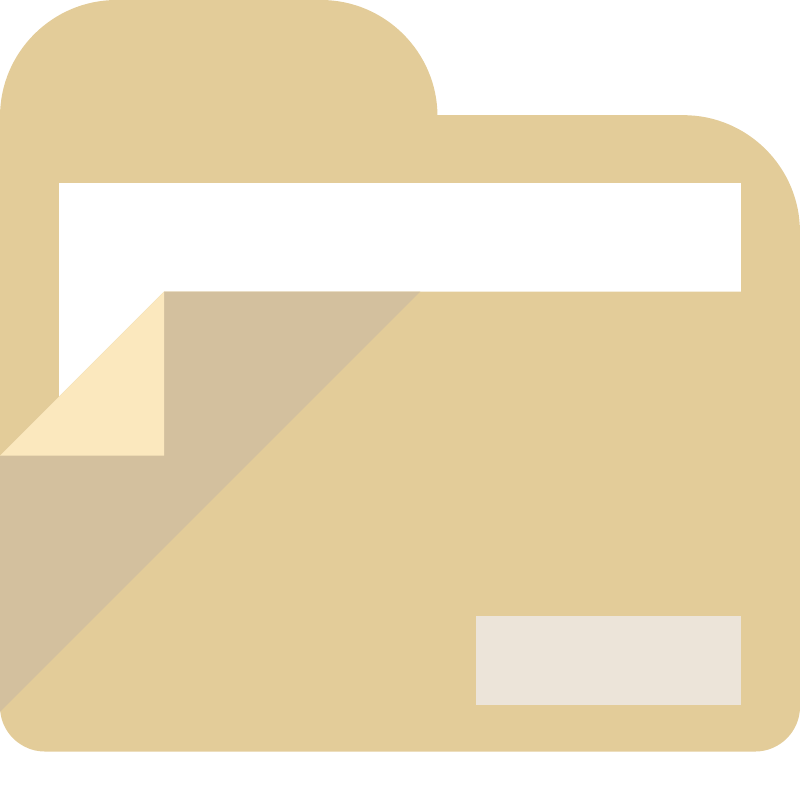}File. & \mcpicon{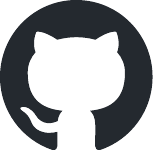}Github & \mcpicon{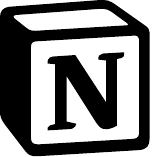}Notion & \mcpicon{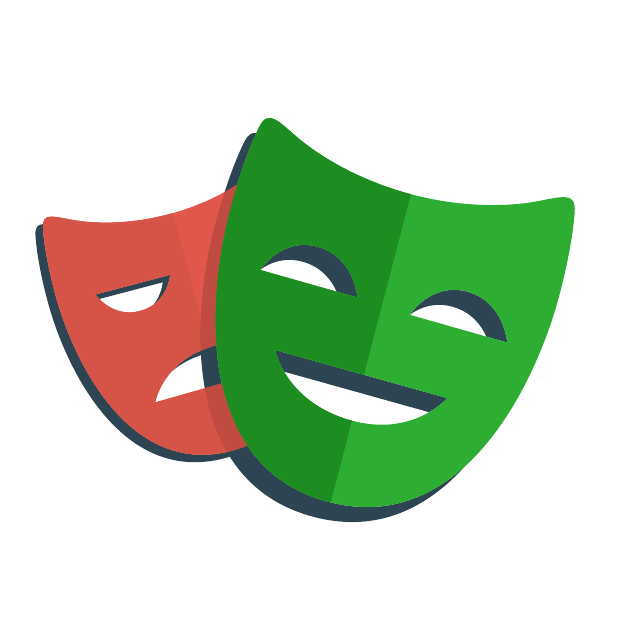}Play. & \mcpicon{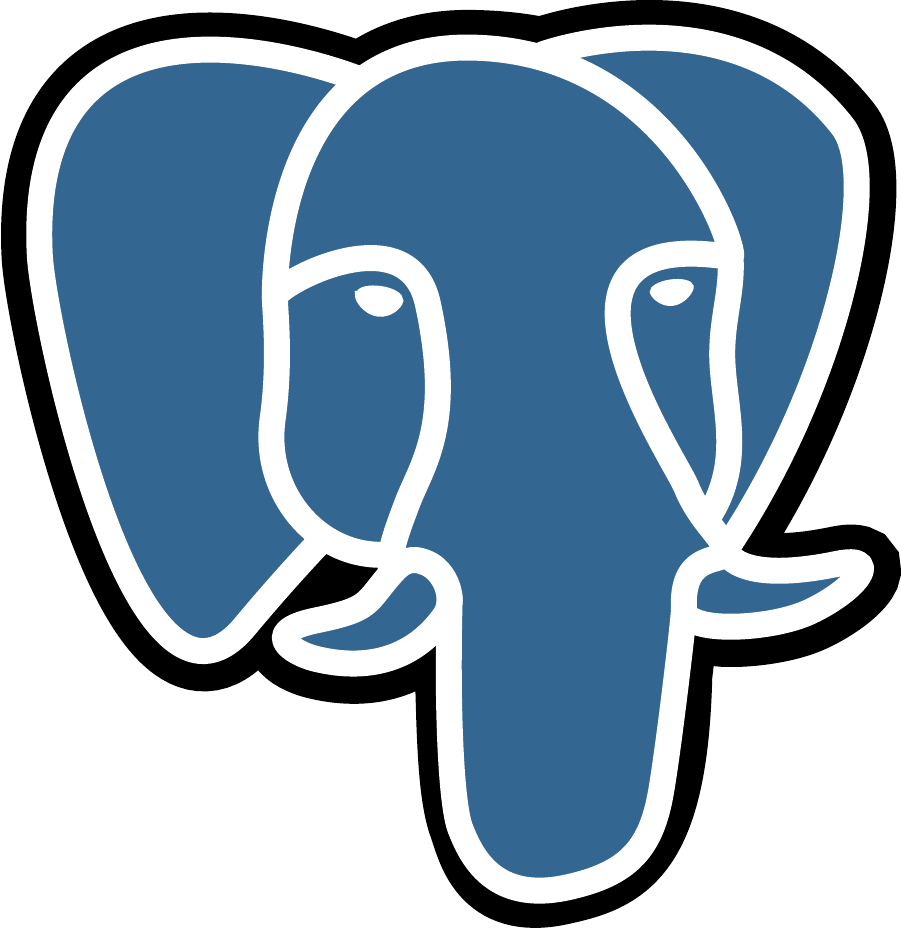}Post. & \textbf{Avg.}
    & WebSearch& Memory & Multi-T. & No live & Live & Relev. & Irrelev. & \textbf{Avg.}
    & Retail & Telecom & Airline & \textbf{Avg.}
    \\
  \midrule
  \rowcolor[HTML]{f0f0f0}
  \multicolumn{19}{c}{\textbf{\textit{Frontier Proprietary Models}}} \\
  \modelicon{1.1em}{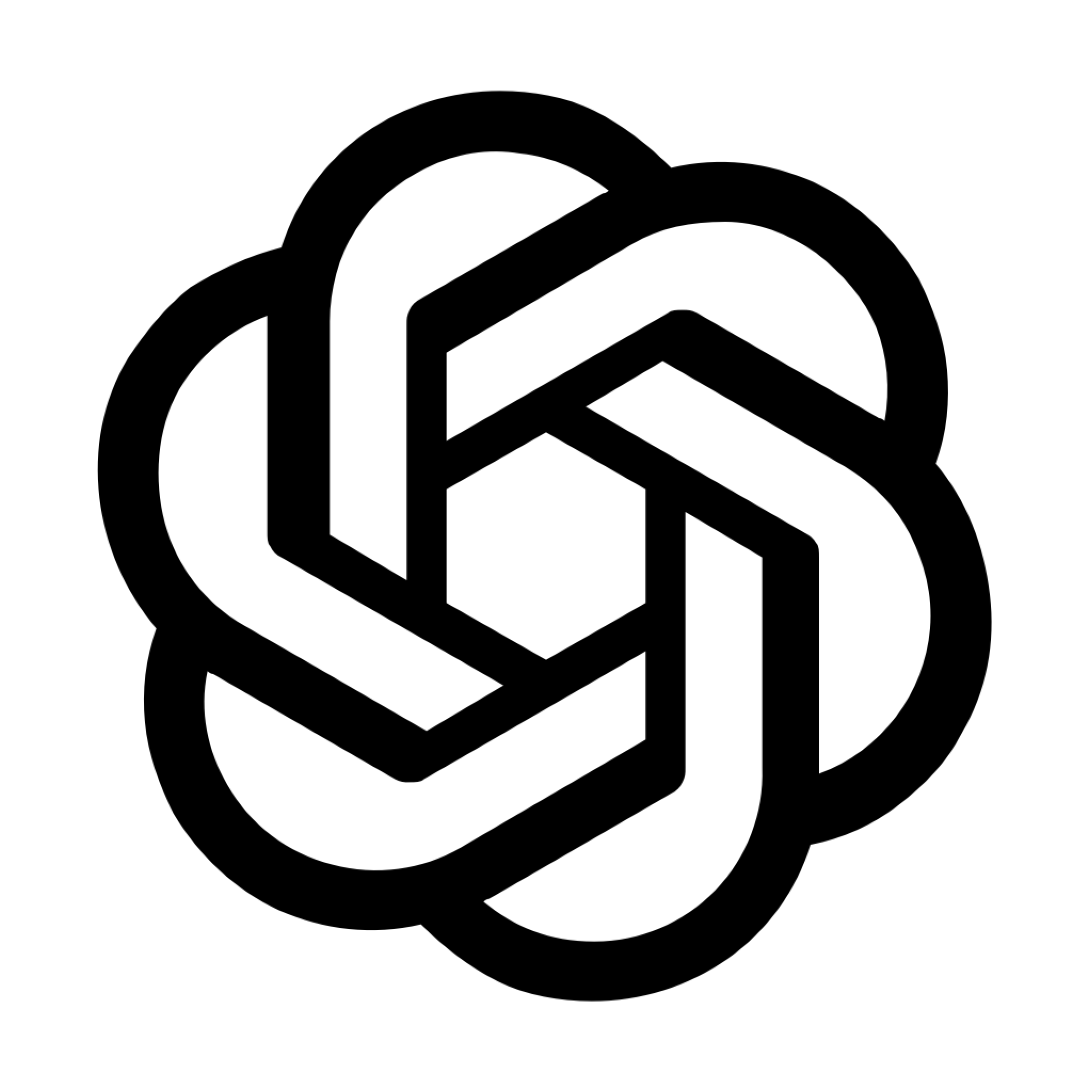}GPT-5.2 High
    & 60.0 & 47.8 & 42.9 & 40.0 & 66.7 & 53.1 & 75.5 & 45.8 & 48.5 & 81.9 & 70.4 & 75.0 & 88.7 & 62.9 & 81.6 & 95.8 & 62.5 & 80.2 \\
  {\scriptsize\textcolor{clrClaude}{\faCircle}}\,Claude Sonnet-4.5
    & 32.5 & 29.4 & 25.0 & 27.0 & 50.0 & 33.3 & 81.0 & 65.0 & 61.4 & 88.7 & 81.1 & 68.8 & 86.6 & 73.2 & 86.2 & 98.0 & 70.1 & 84.7 \\
  \modelicon{1.1em}{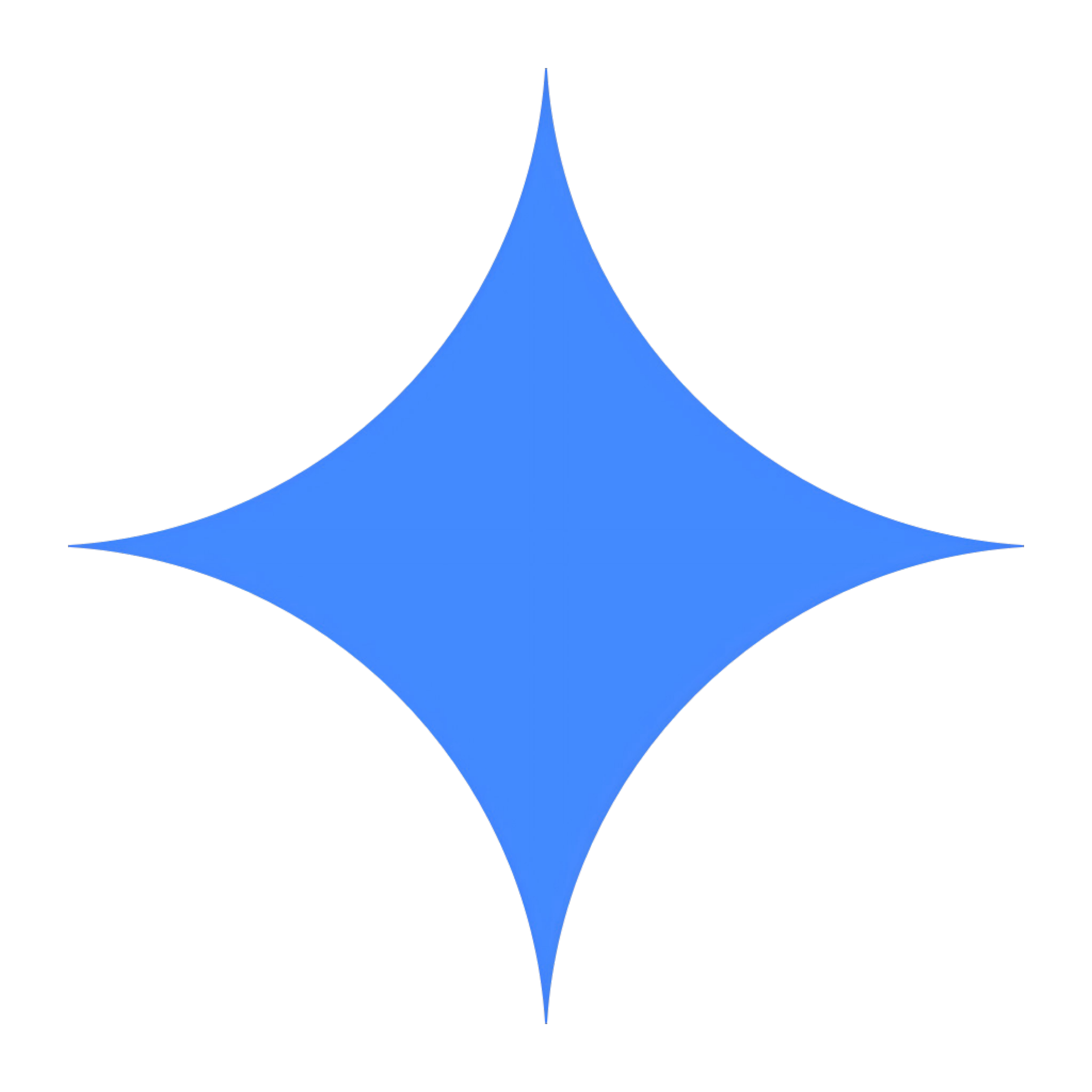}Gemini-3 Pro
    & 56.7 & 45.7 & 43.8 & 40.0 & 70.2 & 50.8 & 80.0 & 61.7 & 60.8 & 90.7 & 83.1 & 68.8 & 85.6 & 72.5 & 85.3 & 98.0 & 72.7 & 85.4 \\
  \modelicon{1.1em}{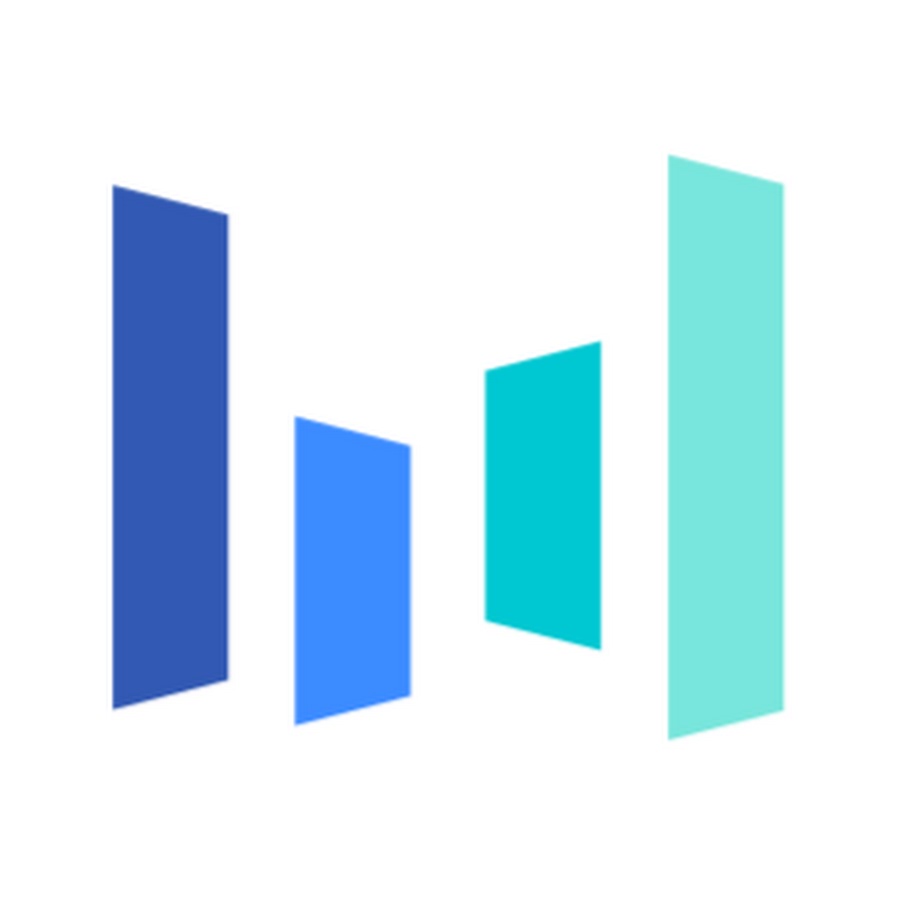}Seed 2.0
    & 60.0 & 39.1 & 53.6 & 40.0 & 81.0 & 54.7 & 92.0 & 57.8 & 62.3 & 89.0 & 82.2 & 76.6 & 75.0 & 73.4 & 90.4 & 94.2 & 64.4 & 83.0 \\
  \rowcolor[HTML]{f0f0f0}
  \midrule
  \multicolumn{19}{c}{\textbf{\textit{Open-Source  Foundation Models (8B–685B)}}} \\

  \modelicon{1.1em}{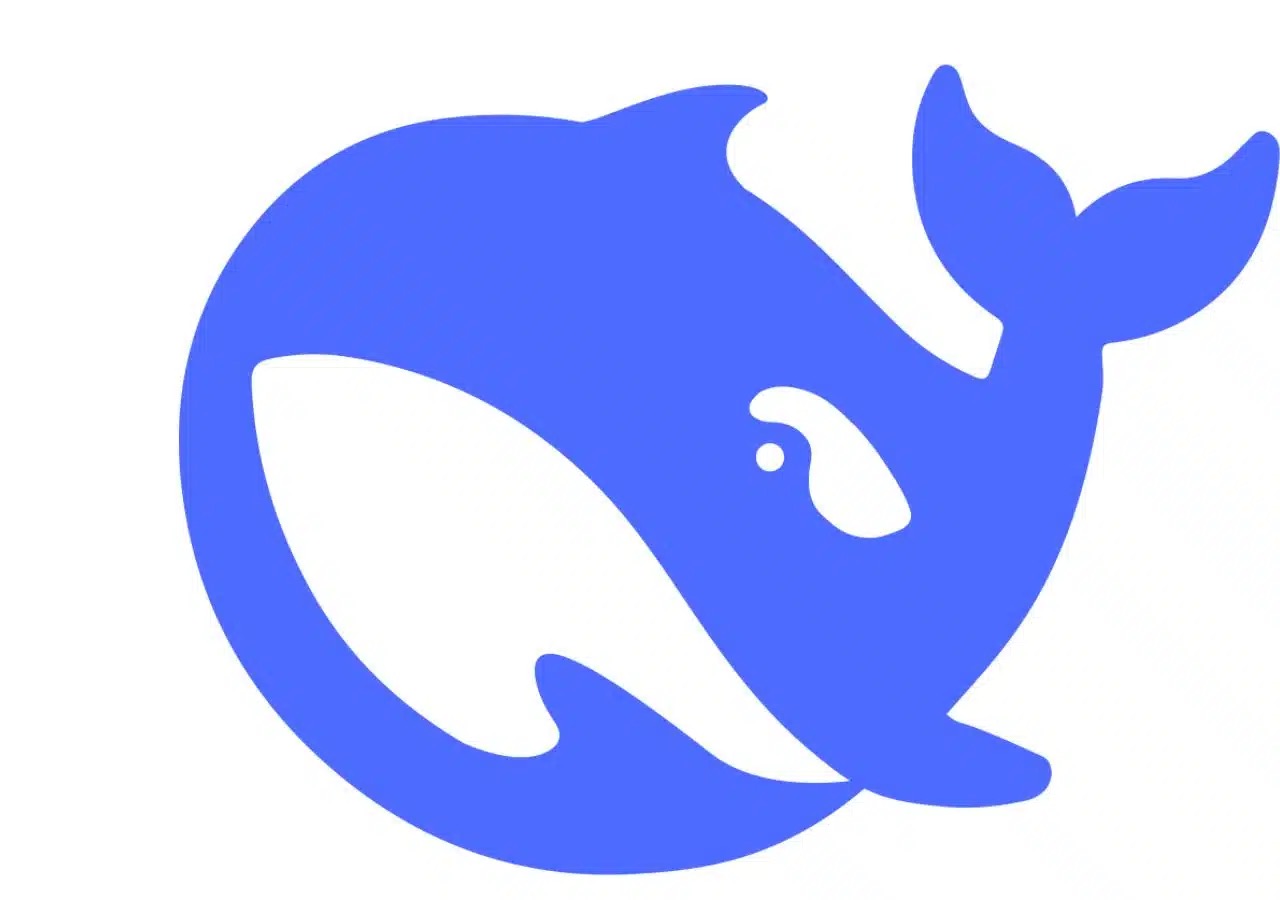}DeepSeek-V3.2-685B
    & 36.7 & 20.7 & 45.5 & 17.0 & 66.6 & 36.7 & 69.5 & 54.2 & 37.4 & 34.9 & 53.7 & 37.5 & 93.2 & 54.1 & -- & -- & -- & 80.3 \\
  \modelicon{1.1em}{openai.pdf}GPT-OSS-120B
    & 5.8 & 4.4 & 3.6 & 3.0 & 7.1 & 4.7 & -- & -- & -- & -- & -- & -- & -- & -- & 67.8 & 49.2 & 48.0 & 55.0 \\
  
    \modelicon{1.0em}{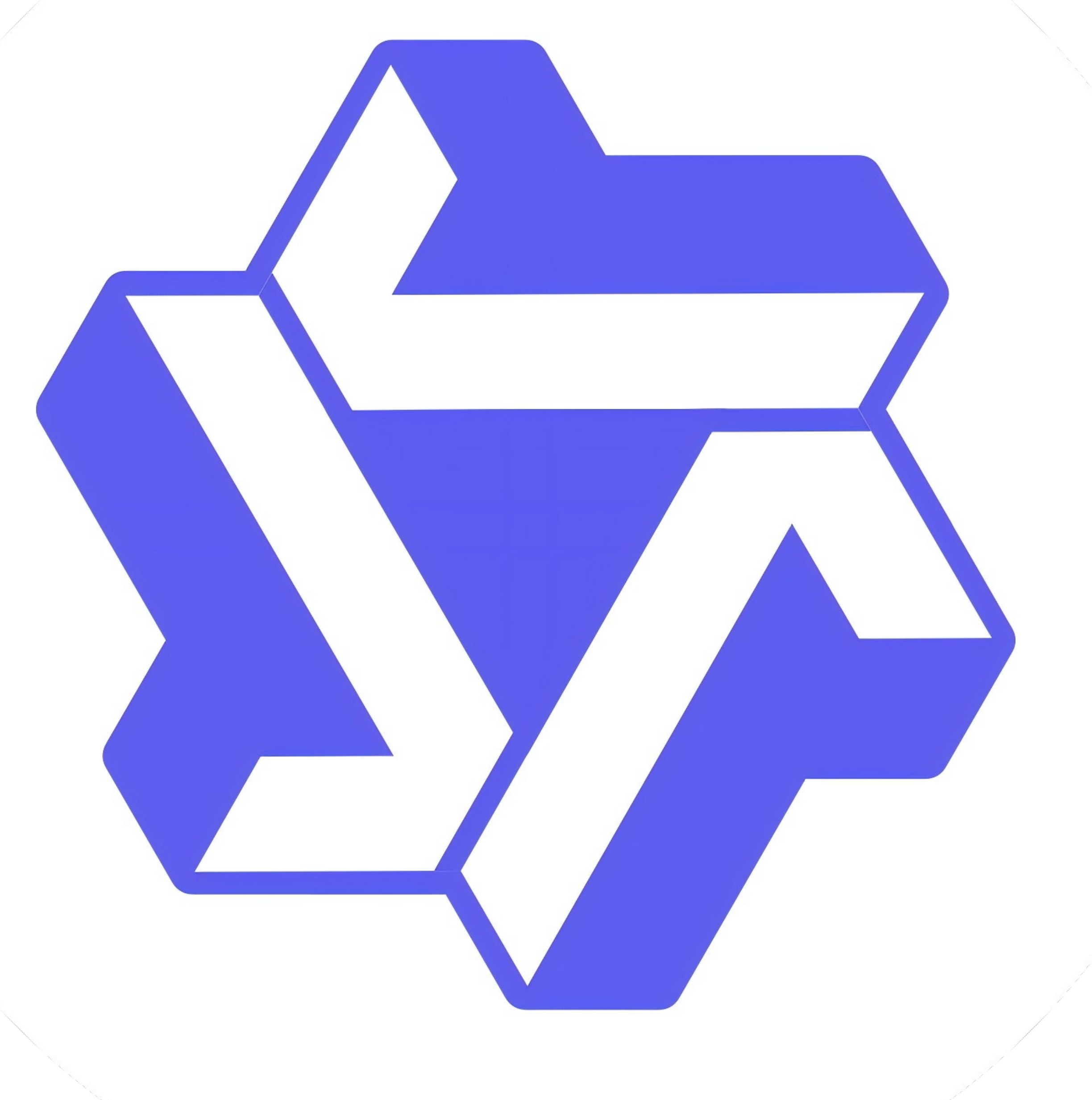}Qwen3-8B
    & 3.3 & 0.0 & 0.0 & \textbf{4.0} & 4.8 & 2.4 & 7.0 & 17.6 & 35.4 & \textbf{90.2} & 80.9 & \underline{81.3} & 77.2 & 40.4 & 34.0 & 18.0 & 26.5 & 26.2 \\
  \modelicon{1.0em}{qwen.pdf}Qwen3-14B
    & 3.3 & \textbf{4.4} & 0.0 & 0.0 & 9.5 & 3.4 & 4.0 & 19.8 & 36.9 & \underline{90.0} & \textbf{82.4} & \underline{81.3} & 79.4 & 41.0 & 55.3 & 14.9 & 27.0 & 32.4 \\
  \modelicon{1.1em}{qwen.pdf}Qwen3-32B
    & 10.0 & 0 & 3.6 & 0 & 23.8 & 7.5 & 26.0 & 15.7 & 43.3 & 90.3 & 82.0 & 81.3 & 82.4 & 46.7 & 59.5 & 27.2 & 48.0 & 44.9 \\
  \modelicon{1.1em}{qwen.pdf}Qwen3-235B-A22B
    & 13.3 & 0 & 10.7 & 0 & 4.8 & 5.8 & 54.0 & 23.9 & 45.4 & 37.4 & 68.9 & 87.5 & 81.7 & 47.9 & 71.9 & 58.0 & 45.6 & 58.5 \\
    
  \midrule
  \rowcolor[HTML]{f0f0f0}
  \multicolumn{19}{c}{\textbf{\textit{Open-Source Environment Scaling Methods (7B-14B)}}} \\
  
  \modelicon{1.0em}{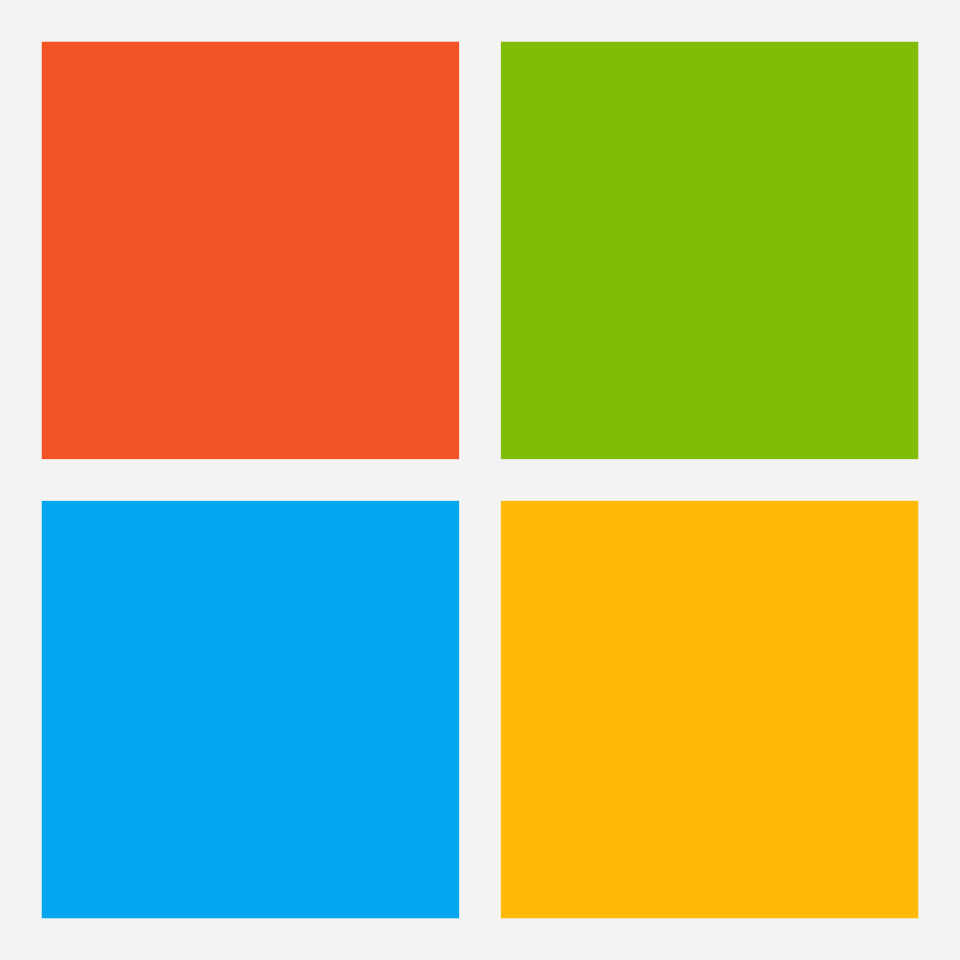}Simulator-8B
    & 3.3 & 0.0 & 0.0 & \textbf{4.0} & 4.8 & 2.4 & 17.5 & 6.0 & 4.1 & 47.6 & 44.6 & 31.3 & \textbf{87.3} & 23.9 & 32.2 & 29.2 & 34.0 & 31.8 \\
  \modelicon{1.0em}{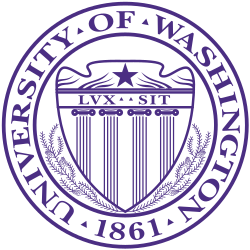}TOUCAN-7B
    & 0.0 & 0.0 & 0.0 & 0.0 & 4.8 & 1.0 & 21.0 & 18.5 & 17.8 & 81.0 & 73.9 & \underline{81.3} & 78.6 & 36.6 & 22.8 & 10.5 & 20.0 & 17.7 \\
  
  \modelicon{1.0em}{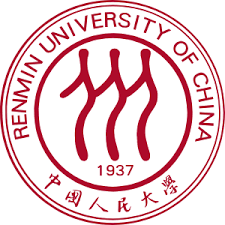}EnvScaler-8B
    & 10.0 & \underline{4.4} & 0.0 & \textbf{4.0} & 9.5 & 5.6 & 23.0 & \underline{21.9} & \underline{47.1} & \underline{88.5} & \textbf{82.2} & \textbf{93.8} & 74.6 & 47.6 & 49.6 & 32.7 & 31.5 & 37.9 \\
  \modelicon{1.0em}{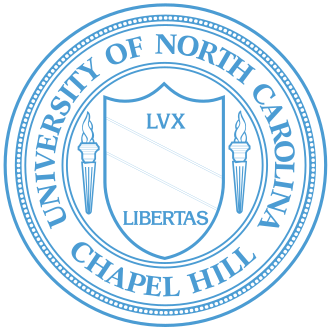}AWM-8B
    & 3.3 & 0.0 & 0.0 & \textbf{4.0} & 4.8 & 2.4 & 9.5 & 15.7 & 34.9 & \textbf{90.2} & 80.5 & \textbf{93.8} & 73.9 & 40.0 & 41.2 & 38.5 & 23.5 & 34.4 \\
  \modelicon{1.0em}{AWM.png}AWM-14B
    & 3.3 & \textbf{8.7} & 0.0 & \textbf{4.0} & 9.5 & 5.1 & 10.0  &19.8  &37.6  &\textbf{90.2} &81.5  & 75.0 & 79.4 &42.4  & 63.6 &17.8 & 31.5 & 39.0 \\

  \modelicon{1.0em}{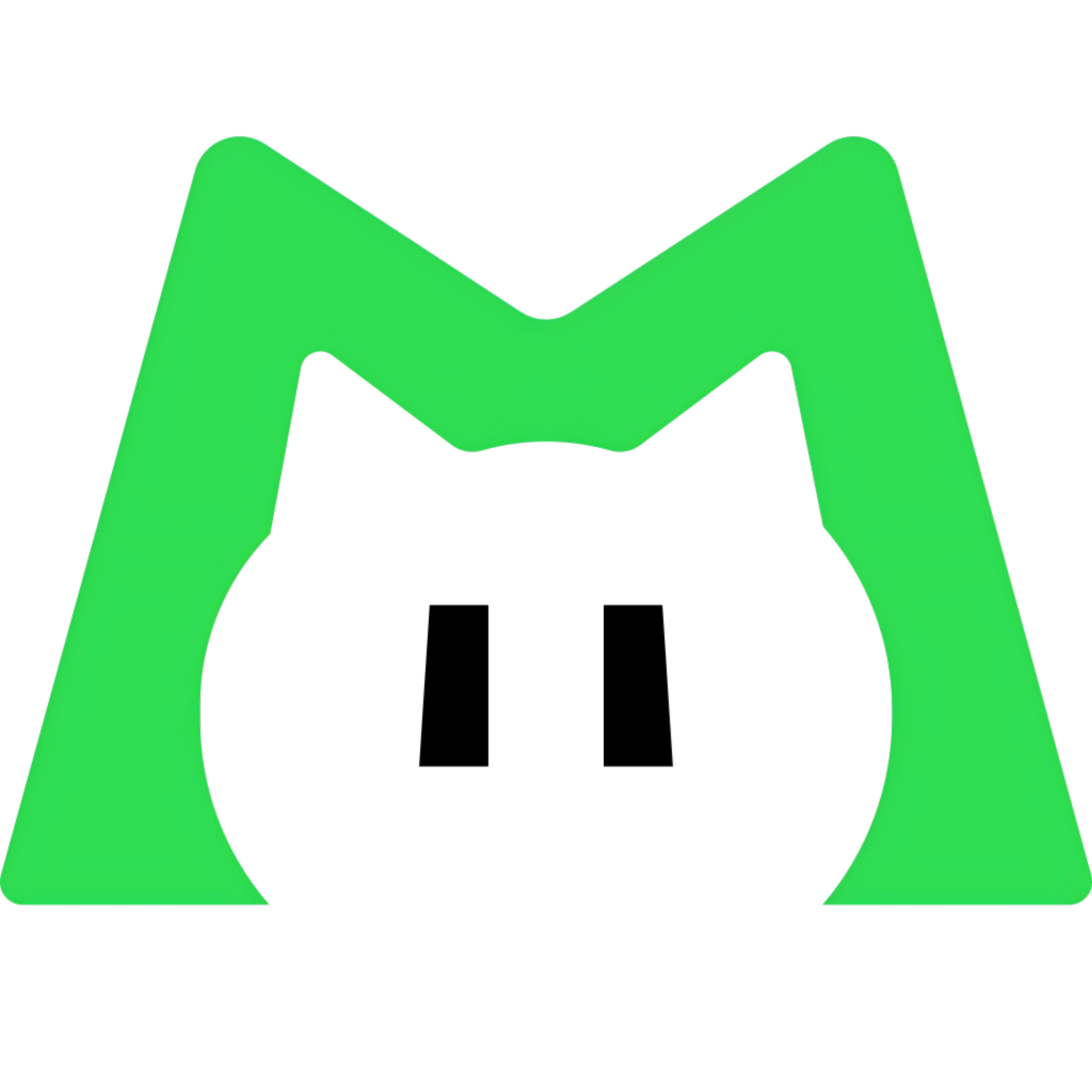}ScaleEnv-8B
    & -- & -- & -- & -- & -- & -- & -- & -- & -- & -- & -- & -- & -- & -- & 50.9 & 27.2 & \underline{37.5} & 38.5 \\

  \hdashline
  \rowcolor[HTML]{e8f0fe}
  \modelicon{1.0em}{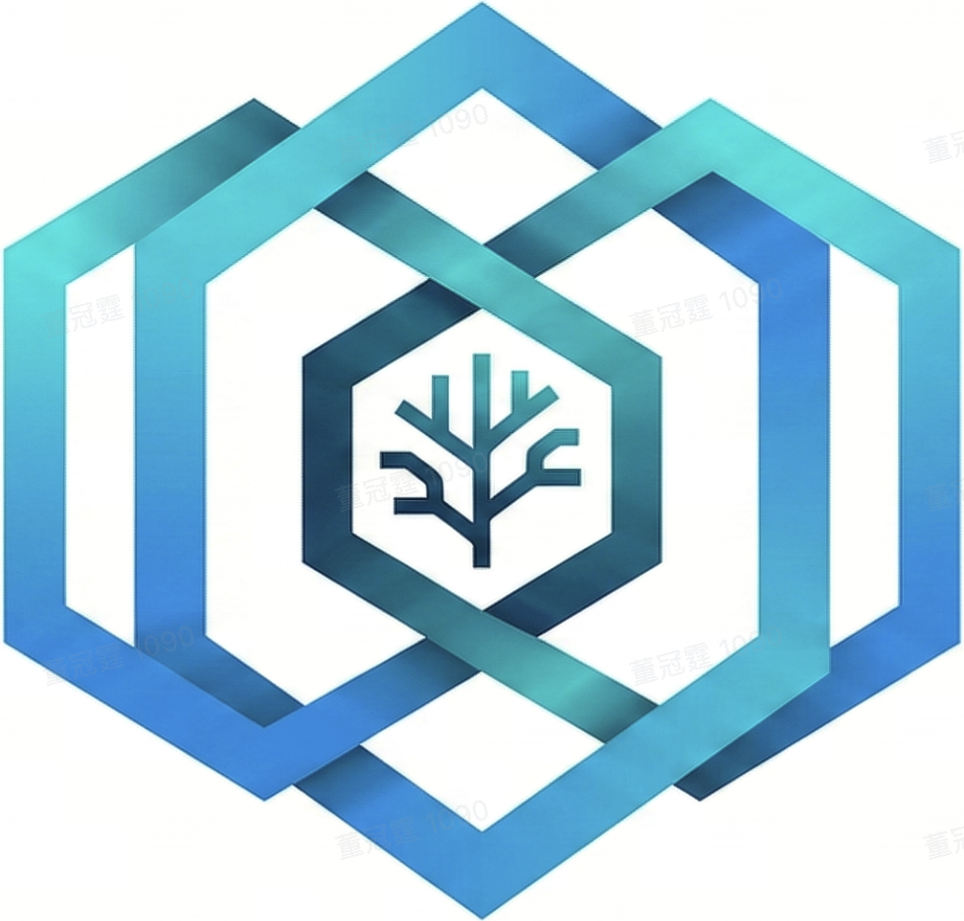}\textbf{Agent-World-8B}
    & \underline{13.3} & \underline{4.4} & \textbf{3.6} & \textbf{4.0} & \underline{19.1} & \underline{8.9} & \underline{47.0} & 21.7 & 44.5 & 83.3 & 79.6 & \textbf{93.8} & 80.2 & \underline{51.4} & \underline{72.8} & \underline{50.9} & \underline{40.0} & \underline{61.8} \\
  
  \rowcolor[HTML]{e8f0fe}
  \modelicon{1.0em}{aw_logo.png}\textbf{Agent-World-14B}
    & \textbf{16.6} & \underline{4.4} & \textbf{3.6} & \textbf{4.0} & \textbf{38.1} & \textbf{13.3} & \textbf{53.0} & \textbf{23.9} & \textbf{53.9} & 82.3 & 79.3 & \textbf{93.8} & \underline{81.0} & \textbf{55.8} & \textbf{74.5} & \textbf{56.1} & \textbf{52.0} & \textbf{65.4} \\
  \bottomrule
  \end{tabular}
  }
  \end{table*}

\subsubsection{Generalization on Long-horizon Agentic Reasoning Scenarios}

\begin{figure}[t]
    \centering
    \includegraphics[width=1.0\linewidth]{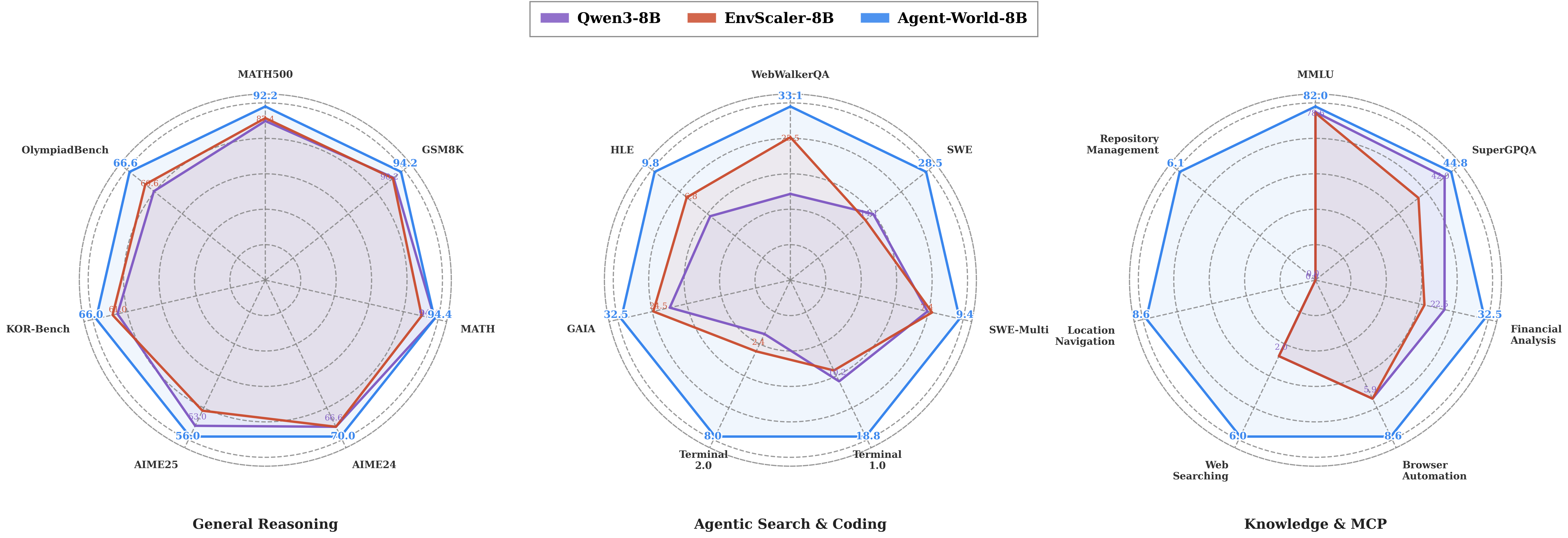}
\caption{\textbf{Generalization across long-horizon agentic reasoning scenarios.} Comparison of Qwen3-8B, EnvScaler-8B, and Agent-World-8B from three capability groups: General Reasoning, Agentic Search \& Coding, and Knowledge \& MCP.}
    \label{fig:longhorizon_agentic_reasoning}
\end{figure}

To further assess long-horizon generalization in agentic tool-use scenarios, we compare Agent-World-8B against strong baselines on 17 benchmarks, organized into three complementary perspectives in Figure~\ref{fig:longhorizon_agentic_reasoning}: \textbf{General Reasoning}, \textbf{Agentic Search \& Coding}, and \textbf{Knowledge \& MCP}. Overall, Agent-World demonstrates strong cross-domain generalization without benchmark-specific tuning, further validating the transferability of our environment-scaling paradigm. The key findings are summarized below.

(1) \textbf{Agent-World strengthens agentic behavior while preserving strong general reasoning.}
On the \textbf{General Reasoning} axis, Agent-World-8B achieves the best overall profile across seven widely-used reasoning benchmarks (MATH500, GSM8K, MATH, AIME24, AIME25, KOR-Bench, and OlympiadBench), with clear gains on most dimensions and no degradation on core math reasoning. This indicates that our Agent-World training pipeline improves difficult multi-step reasoning without sacrificing foundational reasoning capability.

(2) \textbf{The largest gains are observed in long-horizon search and coding tasks.}
On \textbf{Agentic Search \& Coding}, Agent-World-8B consistently outperforms both baselines on WebWalkerQA, SWE-bench Verified, SWE-bench
Multilingual, Terminal~1.0, Terminal~2.0, GAIA, and HLE. These benchmarks stress iterative planning, long-horizon software engineering, deep information retrieval, and multi-tool coordination. The consistent improvements indicate that Agent-World acquires transferable agentic strategies rather than benchmark-specific heuristics. Notably, EnvScaler-8B underperforms its Qwen3-8B backbone on SWE and Terminal~1.0, possibly because its environment expansion is less effective at eliciting complex software-engineering reasoning patterns.

(3) \textbf{Agent-World shows stronger robustness in heterogeneous knowledge and MCP environments.}
On \textbf{Knowledge \& MCP}, Agent-World-8B also substantially outperforms baselines on five relatively orthogonal MCP-Universe capabilities: Browser Automation, Web Searching, Location Navigation, Repository Management, and Financial Analysis. In addition, Agent-World-8B maintains consistent improvements on knowledge-centric dimensions (e.g., MMLU and SuperGPQA), highlighting stronger compositional generalization and adaptation to structurally diverse external tools.

\begin{figure*}[!t]
\centering
\includegraphics[width=\linewidth]{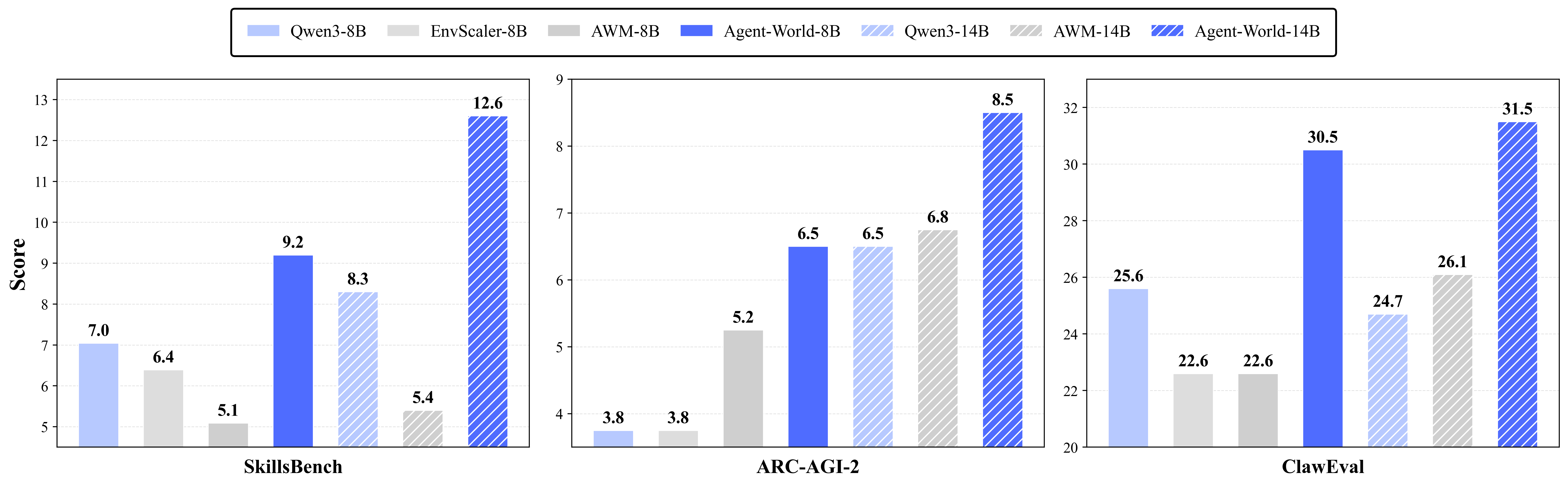}
\caption{\textbf{Generalization on advanced agentic assistant benchmarks.} Comparison of Qwen3, EnvScaler, AWM, and Agent-World series on SkillsBench, ARC-AGI-2, and Claw-Eval.}
\label{fig:agentic_ai_assistant_generalization}
\end{figure*}

\subsubsection{Generalization on Agentic AI Assistant Scenarios}

To further stress-test transfer in advanced assistant settings, we evaluate on three recent highly challenging AI Assistant benchmarks: \textbf{SkillsBench}, \textbf{ARC-AGI-2}, and \textbf{ClawEval}. These benchmarks emphasize long-horizon planning and execution in real-world assistant scenarios. We have the following observations:

(1) \textbf{Existing open-source baselines struggle in real-world AI assistant settings.}
Most baseline models obtain average scores below 20\% across the three benchmarks and do not show consistent gains from 8B to 14B. For example, Qwen3 drops on ClawEval (25.6\% $\rightarrow$ 24.7\%), and AWM shows uneven improvements across tasks. This suggests that naive parameter scaling alone is insufficient for stable long-horizon agentic generalization.

(2) \textbf{Agent-World generalizes strongly to unseen advanced assistant domains.}
Without benchmark-specific training, Agent-World still outperforms strong open-source baselines on these challenging settings. At 8B, Agent-World achieves 9.2\%/6.5\%/30.5\% on SkillsBench/ARC-AGI-2/Claw-Eval, surpassing Qwen3-8B, EnvScaler-8B, and AWM-8B across all three tasks.

(3) \textbf{Agent-World exhibits stable cross-scale gains.}
Unlike the unstable scaling trends of several baselines, Agent-World improves consistently from 8B to 14B (SkillsBench: 9.2\% $\rightarrow$ 12.6\%, ARC-AGI-2: 6.5\% $\rightarrow$ 8.5\%, Claw-Eval: 30.5\% $\rightarrow$ 31.5\%). This supports that our method remains effective across parameter scales and transfers robustly to complex, integrated assistant scenarios.

\begin{figure}[t]
    \centering
    \includegraphics[width=1.0\linewidth]{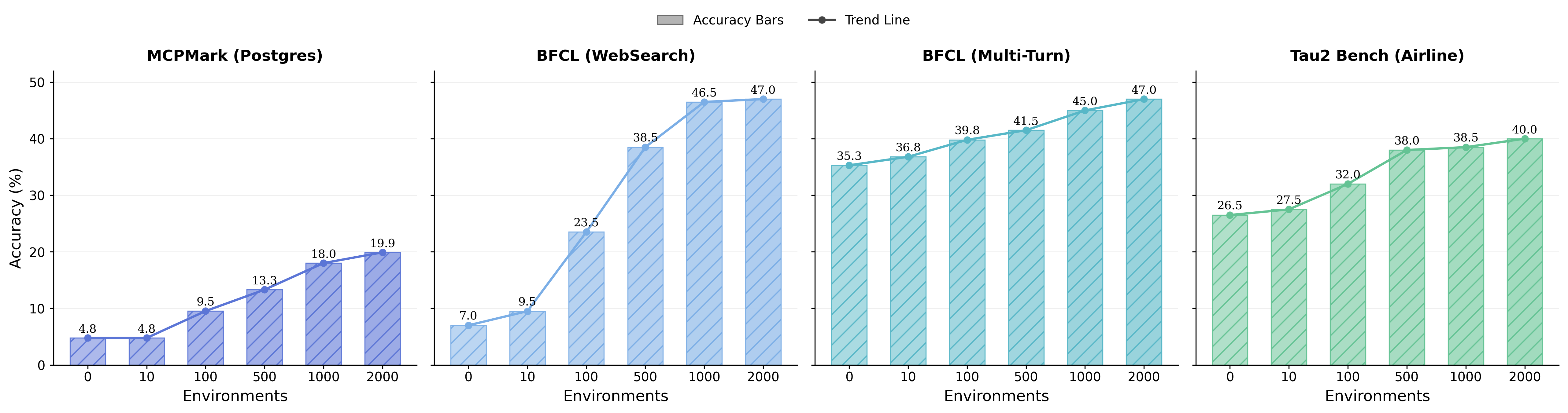}
\caption{\textbf{Scaling relationship of training environments:} Downstream agent performance scales positively with the number of synthesized training environments.}
    \label{fig:scaling_env}
\end{figure}

\subsubsection{Scaling Analysis of Training Environments}
\label{sec:scaling_env}

To analyze how environment scaling affects downstream agentic tool-use performance, we progressively increase the number of training environments from 0 to 10, 100, 500, 1000, and ~2000 (1,978), and evaluate the resulting models on four representative domains in Figure~\ref{fig:intro}: MCPMark (Postgres), BFCL (WebSearch), BFCL (Multi-Turn), and $\tau^2$-Bench (Airline).

Overall, performance improves consistently across all four domains as the environment scale grows, indicating a clear positive scaling relationship. Averaged over the four domains, the score rises from 18.4\% to 38.5\% (+20.1 points), more than doubling the initial level.

A notable trend is the stage-wise gain pattern: performance jumps markedly from 10 to 100 environments and again from 100 to 500, suggesting that moderate-scale expansion rapidly improves coverage of critical interaction patterns. This effect is especially evident on BFCL-V4 and MCPMark: MCPMark (Postgres) improves from 4.8\% to 19.9\%, while BFCL (WebSearch) increases from 7.0\% to 47.0\%. BFCL (Multi-Turn) and $\tau^2$-Bench (Airline) also improve steadily, indicating broad transfer across task types.

From 500 to 2000 environments, the trend remains upward but the marginal improvement gradually decreases, indicating diminishing-yet-positive returns at larger scales. This suggests that early expansion mainly captures missing high-impact environment diversity, while later expansion contributes finer-grained robustness gains.

\subsubsection{Analysis of Continuous Self-Evolution}
\label{sec:self_evolving_analysis}

\begin{wraptable}{r}{0.60\linewidth}
  \vspace{-2em}
  \centering
  \caption{\textbf{Effect of continuous self-evolution.} We run iterative self-evolving loops for Agent-World-14B and the EnvScaler-8B base model.}
  \vspace{-0.5em}
  \label{tab:self_evolving_analysis}
  \setlength\tabcolsep{3.2pt}
  \small
  \begin{tabular}{lccc}
  \toprule
  \textbf{Model / Round} & \textbf{$\tau^2$-Bench} & \textbf{BFCL-V4} & \textbf{MCP-Mark (Post.)} \\
  \midrule
  Agent-World-14B (base) & 60.2 & 52.4 & 29.5 \\
  \quad +1 round & 63.5 {\scriptsize\textcolor{ForestGreen}{(+3.3)}} & 54.9 {\scriptsize\textcolor{ForestGreen}{(+2.5)}} & 36.3 {\scriptsize\textcolor{ForestGreen}{(+6.8)}} \\
  \quad +2 rounds & \textbf{65.4} {\scriptsize\textcolor{ForestGreen}{(+1.9)}} & \textbf{55.8} {\scriptsize\textcolor{ForestGreen}{(+0.9)}} & \textbf{38.1} {\scriptsize\textcolor{ForestGreen}{(+1.8)}} \\
  \midrule
  EnvScaler-8B (base) & 37.9 & 47.6 & 9.5 \\
  \quad +1 round & 40.2 {\scriptsize\textcolor{ForestGreen}{(+2.3)}} & 49.1 {\scriptsize\textcolor{ForestGreen}{(+1.5)}} & 13.9 {\scriptsize\textcolor{ForestGreen}{(+4.4)}} \\
  \quad +2 rounds & \textbf{41.6} {\scriptsize\textcolor{ForestGreen}{(+1.4)}} & \textbf{50.0} {\scriptsize\textcolor{ForestGreen}{(+0.9)}} & \textbf{15.1} {\scriptsize\textcolor{ForestGreen}{(+1.2)}} \\
  \bottomrule
  \end{tabular}
  \vspace{-0.8em}
  \end{wraptable}

To validate Continuous Self-Evolving Agent Training, we run the same two-round self-evolving arena loop (Sec.~\ref{sec:self-evolving-training}) from two different starting points: Agent-World-14B and the EnvScaler-8B base model. In each round, the current policy is first evaluated on newly synthesized verifiable tasks in held-out arena environments; a diagnosis agent then identifies weak environments and failure modes from executable traces; finally, targeted synthesis and continual RL produce the next-round policy.

Table~\ref{tab:self_evolving_analysis} shows monotonic gains on all three evaluation suites for both models. For Agent-World-14B, performance on $\tau^2$-Bench/BFCL-V4/MCP-Mark improves from 45.3\%/52.4\%/29.5\% to 50.5\%/55.8\%/38.1\% after two rounds. Importantly, EnvScaler-8B also improves from 37.9\%/47.6\%/9.5\% to 41.6\%/50.0\%/15.1\%, indicating that the loop not only benefits our base model but also yields sustained gains for other environment-scaling baselines without relying on Agent-World initialization.

Notably, the largest gains across two rounds appear on MCP-Mark: +8.6\% for Agent-World and +5.6\% for EnvScaler. This benchmark requires stronger state tracking and deeper interaction with realistic MCP server environments. This matches our self-evolving objective in Sec.~\ref{sec:self-evolving-training}: diagnosis continually localizes environment-specific weaknesses from closed-loop traces, while targeted synthesis generates harder instances around those failures, which is particularly beneficial for challenging agentic execution scenarios. BFCL-V4 and $\tau^2$-Bench also improve steadily, indicating concurrent gains in environment grounding and multi-turn tool coordination.

Furthermore, second-round gains are smaller than first-round gains but remain positive, reflecting diminishing yet still effective returns. From the environment-diagnosis perspective, we find that early rounds mainly fix pattern-level errors in unfamiliar environment interactions, while later rounds focus on residual failures, especially in long-horizon complex interaction cases.

Overall, Agent-World treats scalable environments as a persistent diagnostic arena and achieves continual policy improvement through agent-environment co-evolution, substantially outperforming one-pass static training.

\begin{figure}[!t]
    \centering
    \includegraphics[width=0.85\linewidth]{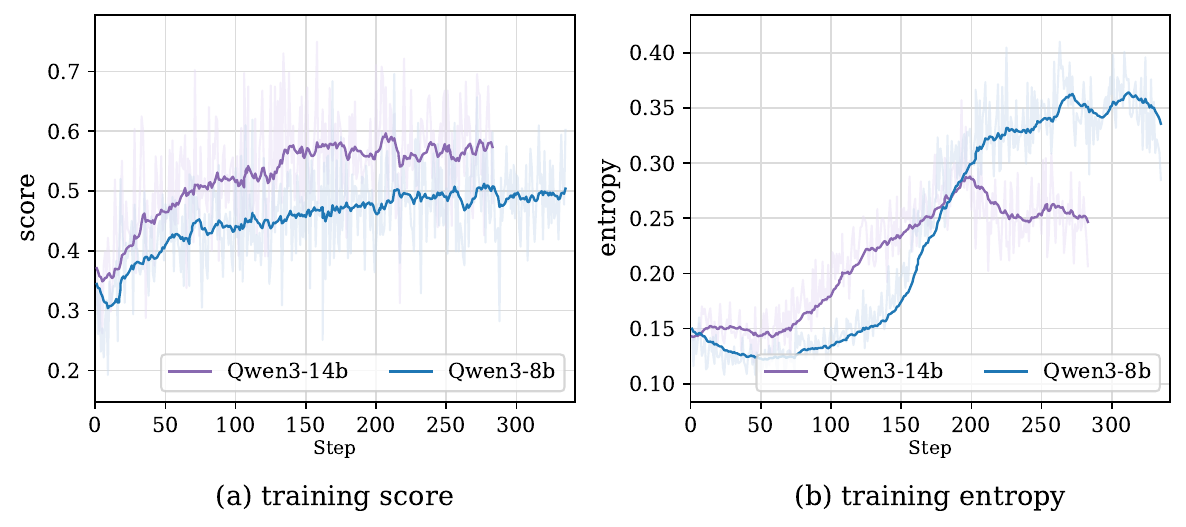}
    \caption{\textbf{Training Dynamics of Agent-World.} (a) Training reward score and (b) actor entropy over training steps for Qwen3-8B and Qwen3-14B backbones using GRPO on synthesized environments. Curves are exponentially smoothed for clarity.}
    \label{fig:training_dynamics}
\end{figure}

\subsection{Training Dynamics of Agent-World.} As shown in Figure~\ref{fig:training_dynamics}, we present the multi-environment reinforcement learning curves of Agent-World-8B and Agent-World-14B.  
We observe clear upward reward trends for both Qwen3-8B/14B backbones. These results indicate that policy performance improves steadily under GRPO supervision; this trend is also consistent across different environment complexities, further supporting the general effectiveness of multi-turn RL with executable rewards. Meanwhile, tool-use training shows relatively stable entropy growth over time (Figure~\ref{fig:training_dynamics}b). This suggests that as the model gradually adapts to unseen APIs and heterogeneous state transitions, it maintains or even expands its exploration space, learning new interaction patterns instead of collapsing prematurely into narrow exploitation. This behavior indicates that Agent-World sustains exploration of agent execution patterns in structurally diverse and highly interactive real-world MCP environments.

\section{Related Work}

\subsection{Scalable Environment Synthesis for Agent Training}
As agent training shifts from imitation learning to self-exploration and evolution within interactive environments, training environments have become essential infrastructure~\citep{huang2025scaling,andrews2025scaling,fang2025TowardsGeneralAgenticIntelligenceViaEnvironmentScaling}.
However, real-world services and systems often have restricted access, while manually constructed sandboxes suffer from high costs and poor scalability.
To automatically scale environments for training LLM agents, one line of research focuses on LLM-driven simulation~\citep{li2025simulatingenvironmentsreasoningmodels,ren2026simworldopenendedrealisticsimulator,li2025wordworldlargelanguage,feng2025webworldmodels}. By leveraging the intrinsic world modeling and reasoning capabilities of LLMs, these systems use LLMs to simulate environmental feedback and state transitions.
Another line of research focuses on programmatic environment synthesis, constructing deterministic sandboxes via programs, database backends, or finite-state machines~\citep{tian2026astraautomatedsynthesisagentic,wang2026agentworldmodelinfinity,song2026envscaler,sullivan-etal-2025-procedural,cai2025autoforgeautomatedenvironmentsynthesis,fang2025TowardsGeneralAgenticIntelligenceViaEnvironmentScaling,tu2026scaleenvscalingenvironmentsynthesis}. Frameworks including EnvScaler~\citep{song2026envscaler}, AWM~\citep{wang2026agentworldmodelinfinity}, and AutoForge~\citep{cai2025autoforgeautomatedenvironmentsynthesis} leverage LLMs to plan and generate sandboxes and tasks comprising executable programs, databases, or tool interfaces, and provide rule-based reward signals for reinforcement learning.
In addition, InfiniteWeb~\citep{zhang2026infinitewebscalablewebenvironment} expands synthesis to web-based and multimodal contexts. Meanwhile, ARE~\citep{andrews2025scaling} incorporates asynchronous temporal dynamics to better align simulated environments with reality.
Distinct from these works, Agent-World utilizes real MCP server metadata for intelligent environment discovery and modeling. By autonomously building theme-matched databases and executable tools from the web, it achieves deep anchoring within the real-world tool ecosystem. Moreover, by leveraging tool graphs and programmatic synthesis to generate verifiable tasks with progressive difficulty, it establishes a broad and challenging training foundation for agents.

\subsection{Agentic Reinforcement Learning}

Recent agentic reinforcement learning has rapidly expanded from single-tool optimization to long-horizon web agents~\citep{zhang2025landscape}. Early search-centric systems show that rule-based or verifiable RL can improve autonomous information-seeking behavior~\citep{r1searcher,chen2025research,search-r1}. Follow-up work further strengthens tool-use training through improved reward design and policy optimization, including Tool-Star, ToolRL, OTC, and ARPO~\citep{dong2025toolstar,qian2025toolrl,wang2025otc,dong2025arpo}. In parallel, scalability-oriented research explores asynchronous pipelines and large-scale post-training for long-horizon agents~\citep{asearcher,WebSailor}, while tree-structured rollouts improve exploration efficiency under high-entropy action spaces~\citep{GIGPO,li2025treepobridginggappolicy}. Beyond single-agent settings, recent studies increasingly adopt multi-agent and interactive training paradigms, such as agentic RL with multi-agent distillation and multi-turn, user-interacting RL~\citep{Chain-of-Agents,zhao2025muarlmultiturnuserinteractingagent,yuan2025marshal,chen2025improving}, and this line of work is now extending into multimodal settings~\citep{deepeyes,zhang2025thyme,wemath,We-Math2.0,qiao2025v}. Despite this progress, most existing methods still emphasize policy optimization on relatively fixed training distributions. Recent environment-scaling efforts expose agents to broader and more diverse environments~\citep{song2026envscaler,wang2026agentworldmodelinfinity,tu2026scaleenvscalingenvironmentsynthesis,mou2026toolsafe}, yet explicit coupling of diagnosis, targeted environment--task refresh, and continual RL remains limited. Agent-World is designed to fill this gap through a continuous agent RL loop.

\section{Conclusion}

In this paper, we presented \textbf{Agent-World}, a self-evolving training arena for general-purpose agents in realistic tool environments. Agent-World unifies two tightly coupled components: \textbf{Agentic Environment-Task Discovery}, which mines topic-aligned real-world databases and executable toolsets from large-scale themes and synthesizes verifiable tasks with controllable difficulty; and \textbf{Continuous Self-Evolving Agent Training}, which combines multi-environment reinforcement learning with an agentic diagnostic arena to identify capability gaps and drive targeted iterative data expansion. Experiments across 23 challenging benchmarks demonstrate that Agent-World consistently improves performance over strong baselines. Further analyses reveal clear scaling trends with respect to environment diversity, evolution rounds, and task difficulty, suggesting that scalable realistic environments are not only useful data sources, but also critical infrastructure for advancing general agent capabilities. 


\clearpage
\phantomsection
\section*{Contributions}
\label{sec:contribution}

\textbf{Authors} Guanting Dong$^{1,*}$, Junting Lu$^2$, Junjie Huang$^{2,*}$, Wanjun Zhong$^{2,\dagger}$, Longxiang Liu$^2$, Shijue Huang$^{2,*}$, Zhenyu Li$^2$, Yang Zhao$^{2,*}$, Xiaoshuai Song$^1$, Xiaoxi Li$^1$, Jiajie Jin$^1$, Yutao Zhu$^1$, Hanbin Wang$^{2,*}$, Fangyu Lei$^{2,*}$, Qinyu Luo$^2$, Mingyang Chen$^2$, Zehui Chen$^2$, Jiazhan Feng$^2$, Ji-Rong Wen$^1$, Zhicheng Dou$^{1,\dagger}$

\textbf{Affiliations} $^1$Gaoling School of Artificial Intelligence, Renmin University of China, $^2$ByteDance Seed

\textbf{Acknowledgment} We greatly thank Yujia Qin$^2$ and Guang Shi$^2$ for supporting this work and providing valuable suggestions. We also thank Yifei Chen$^1$ for valuable discussions.

$^{*}$ Work was done during their internship at ByteDance Seed\\
$^{\dagger}$ Corresponding Author

\clearpage

\bibliographystyle{plainnat}
\bibliography{main}

\clearpage

\beginappendix

%
%


\startcontents[sections]
\printcontents[sections]{l}{1}{\setcounter{tocdepth}{3}}


\section{Agentic Diagnosis Prompt}\label{sec:diagnosis-prompt}

As described in Section~\ref{sec:self-evolving-arena}, the diagnosis agent $\delta$ receives per-task failure traces, error statistics, and environment metadata, and outputs a structured weakness report with targeted task-generation guidelines. The full prompt template is presented below; Table~\ref{tab:diagnosis-output-schema} specifies the expected output schema.

\begin{tcolorbox}[
colback=white!10!white,
colframe=CaseBlue!85!black,
title=Prompt A: System Prompt for Diagnosis Agent $\delta$,
breakable]
You are an expert agent diagnostician. You will be given evaluation results from an AI agent operating in tool-calling environments. Your task is to:

\begin{enumerate}[leftmargin=1.4em, itemsep=2pt]
    \item \textbf{Identify failure patterns}: Categorize each failure into root-cause types.
    \item \textbf{Rank weak environments}: Determine which environments are most problematic.
    \item \textbf{Generate targeted guidelines}: For each weak environment, produce specific task-generation guidelines to address the identified weaknesses.
\end{enumerate}

Output a structured JSON diagnosis report following the schema provided.
\end{tcolorbox}

\begin{tcolorbox}[
colback=white!10!white,
colframe=CaseBlue!85!black,
title=Prompt B: User Prompt Template for Diagnosis Agent $\delta$,
breakable]
\textbf{Agent Under Test} \\
Model: \texttt{\{model\_name\}} \\
Evaluation Protocol: Closed-loop rollout with tool-calling, followed by rubric-based verification \\
Total Instances: \texttt{\{total\_instances\}} \quad Overall Pass Rate: \texttt{\{pass\_rate\}}

\medskip
\textbf{Error Distribution by Environment}

\texttt{\{error\_distribution\_table\}}

\medskip
\textbf{Failure Traces (per-task)}

For each failed instance, the following fields are provided:
\begin{itemize}[leftmargin=1.2em, itemsep=1pt]
    \item \textbf{Environment}: the domain/environment type
    \item \textbf{Task Description}: what the agent was asked to do
    \item \textbf{Rubrics}: the evaluation criteria (JSON with per-criterion checks)
    \item \textbf{Tool Schemas}: available tool interfaces
    \item \textbf{Agent Conversation}: the full multi-turn interaction trace
    \item \textbf{Evaluation Result}: per-criterion scores and failure reasons
    \item \textbf{Eval Stats}: steps taken, finish reason, per-tool success/fail counts
\end{itemize}

\texttt{--- BEGIN FAILURE TRACES ---} \\
\texttt{\{failure\_traces\}} \\
\texttt{--- END FAILURE TRACES ---}

\medskip
\phantomsection
\label{tab:diagnosis-output-schema}
\textbf{Output Requirement} \\
Produce a JSON diagnosis report with the schema defined in Table~\ref{tab:diagnosis-output-schema}. Focus on \textbf{actionable, specific diagnosis} grounded in evidence from the traces above. Avoid generic advice.
\end{tcolorbox}

\begingroup

\makeatletter
\@ifundefined{envtool}{\newcommand{\envtool}[1]{\texttt{\small #1}}}{}
\makeatother

\definecolor{EnvVisPrimary}{HTML}{1B4F72}
\definecolor{EnvVisLightBlue}{HTML}{D6EAF8}
\definecolor{EnvVisAccentTeal}{HTML}{0E6655}
\definecolor{EnvVisSlate}{HTML}{2C3E50}
\definecolor{EnvVisCodeBg}{HTML}{F7F9FB}
\definecolor{EnvVisPurple}{HTML}{6C3483}

\newtcolorbox{envviscapsule}[1][]{%
  enhanced, breakable,
  colback=EnvVisLightBlue!35!white,
  colframe=EnvVisPrimary,
  coltitle=white,
  fonttitle=\bfseries\Large,
  attach boxed title to top left={yshift=-3mm, xshift=5mm},
  boxed title style={colback=EnvVisPrimary, sharp corners},
  sharp corners=south,
  rounded corners=north,
  boxrule=0.9pt,
  top=5mm, left=5mm, right=5mm, bottom=5mm,
  title={#1}
}

\newtcolorbox{envvisinner}[1][]{%
  enhanced,
  colback=white,
  colframe=EnvVisAccentTeal!55!black,
  boxrule=0.45pt,
  sharp corners,
  left=3.5mm, right=3.5mm, top=2.5mm, bottom=2.5mm,
  title={\small\bfseries #1}
}

\lstdefinestyle{envvispycode}{%
  language=Python,
  basicstyle=\small\ttfamily\color{EnvVisSlate},
  backgroundcolor=\color{EnvVisCodeBg},
  frame=single,
  framerule=0.4pt,
  rulecolor=\color{EnvVisPrimary!40!white},
  xleftmargin=3mm,
  xrightmargin=3mm,
  aboveskip=6pt,
  belowskip=6pt,
  breaklines=true,
  columns=fullflexible,
  keepspaces=true,
  showstringspaces=false,
  keywordstyle=\color{EnvVisPrimary},
  commentstyle=\color{EnvVisAccentTeal!80!black},
  stringstyle=\color{EnvVisPurple},
  literate={—}{{---}}1
}

\section{Environment Visualizations}\label{sec:appendix-env-visualizations}

This section collects self-contained \texttt{tcolorbox} capsules for representative environments (same layout as standalone \texttt{appendix\_arxiv\_local\_env.tex}): environment summary, directory-style file map, tool list with intent, and one listing-style tool implementation.
Duplicate the \texttt{subsection} + \texttt{envviscapsule} pattern to add more domains.

\clearpage
\subsection{\texttt{Arxiv\_local}}

\begin{envviscapsule}[\faBook\; Environment: \texttt{Arxiv\_local}]

\section*{Environment}

This environment simulates a small, fully offline ``local arXiv'' library: each paper is represented by a Markdown metadata card on disk, plus a JSON manifest for fast title indexing. Agents never call the public arXiv API directly here; instead they use deterministic Python-backed tools that read \envtool{manifest.json} and files under \envtool{papers/}. The domain is training-friendly and self-contained: prompts, on-disk fixtures, reference answers, and verifiers can all assume these paths and schemas.

\section*{File summary}

\begin{envvisinner}[Directory layout (repository root)]
\begin{lstlisting}[style=envvispycode, language={}, morekeywords={}]
./
  manifest.json              -- JSON map: relative path -> canonical paper title
  papers/
    1706.03762.md            -- Metadata card (Attention Is All You Need)
    1810.04805.md            -- Metadata card (BERT)
    1512.03385.md            -- Metadata card (ResNet)
    1406.2661.md             -- Metadata card (GANs)
    1806.07366.md            -- Metadata card (Neural ODEs)
    2005.14165.md            -- Metadata card (GPT-3 / few-shot LMs)
    2102.12092.md            -- Metadata card (DALL-E / zero-shot T2I)
    2103.00020.md            -- Metadata card (CLIP)
    2003.08934.md            -- Metadata card (NeRF)
    1910.11333.md            -- Metadata card (quantum supremacy supplement)
\end{lstlisting}
\end{envvisinner}

\medskip
\noindent\textbf{Manifest and cards (what each file is for)}
\begin{itemize}[leftmargin=1.2em, itemsep=4pt]
  \item \textbf{\envtool{manifest.json}} --- Maps each \envtool{papers/<id>.md} path to the human-readable title; used for $O(1)$-ish listing and search without opening every card.
  \item \textbf{\envtool{papers/*.md}} --- One Markdown file per arXiv ID with lines such as \texttt{Authors:}, \texttt{Abstract:}, \texttt{Subjects:}, optional \texttt{Comments:}, \texttt{DOI:}, \texttt{URL:}, and when present \texttt{Journal reference:} / \texttt{Project page:}. The first \texttt{\#} heading is treated as the title line for parsers that only skim headings.
\end{itemize}

\section*{Tool list}

\noindent\textbf{Tools (each paired with intent)}
\begin{itemize}[leftmargin=1.2em, itemsep=6pt]
  \item \textbf{\envtool{list\_local\_papers}} --- Returns every \texttt{(paper\_id, title)} pair discovered via \envtool{manifest.json}; empty list if the manifest is missing so callers can distinguish ``no corpus'' from ``parse error''.
  \item \textbf{\envtool{get\_local\_paper\_metadata}} --- Given \texttt{paper\_id}, opens the corresponding Markdown card and returns a JSON-friendly dict of canonical fields (\texttt{title}, \texttt{authors}, \texttt{abstract}, \texttt{subjects}, \texttt{comments}, \texttt{doi}, \texttt{url}, \texttt{journal\_reference}, \texttt{project\_page}) with \texttt{None} for absent lines; raises if the file is missing.
  \item \textbf{\envtool{search\_local\_papers}} --- Case-insensitive substring search over \emph{titles only} (values in the manifest), returning matching id/title rows; cheap and deterministic, suitable for agent planning loops before paying the cost of opening cards.
\end{itemize}

\section*{Tool example}

Below is the reference implementation for \envtool{get\_local\_paper\_metadata}; the function name matches the exposed tool name and the body is intentionally boring, line-oriented parsing so it is easy to audit in reviews.

\begin{lstlisting}[style=envvispycode]
def get_local_paper_metadata(paper_id):
    import os
    md_path = os.path.join('papers', f'{paper_id}.md')
    if not os.path.isfile(md_path):
        raise FileNotFoundError(f'Paper metadata file not found: {md_path}')
    with open(md_path, 'r', encoding='utf-8') as f:
        lines = f.readlines()
    fields = {
        'paper_id': paper_id,
        'title': None,
        'authors': None,
        'abstract': None,
        'subjects': None,
        'comments': None,
        'doi': None,
        'url': None,
        'journal_reference': None,
        'project_page': None,
    }
    for line in lines:
        line = line.strip()
        if not line:
            continue
        if fields['title'] is None and line.startswith('#'):
            fields['title'] = line.lstrip('#').strip()
            continue
        low = line.lower()
        if low.startswith('authors:'):
            fields['authors'] = line.split(':', 1)[1].strip()
        elif low.startswith('abstract:'):
            fields['abstract'] = line.split(':', 1)[1].strip()
        elif low.startswith('subjects:'):
            fields['subjects'] = line.split(':', 1)[1].strip()
        elif low.startswith('comments:'):
            fields['comments'] = line.split(':', 1)[1].strip()
        elif low.startswith('doi:'):
            fields['doi'] = line.split(':', 1)[1].strip()
        elif low.startswith('url:'):
            fields['url'] = line.split(':', 1)[1].strip()
        elif low.startswith('journal reference:'):
            fields['journal_reference'] = line.split(':', 1)[1].strip()
        elif low.startswith('project page:'):
            fields['project_page'] = line.split(':', 1)[1].strip()
    return fields
\end{lstlisting}

\end{envviscapsule}

\clearpage
\subsection{\texttt{Emails}}

\begin{envviscapsule}[\faEnvelope\; Environment: \texttt{Emails}]

\section*{Environment}

This environment simulates a \textbf{file-backed mailbox} for training and evaluation: a large historical \textbf{Enron spam/ham} CSV supplies raw distribution statistics, while a compact \textbf{synthetic export} (\texttt{emails\_export.json}) holds 2{,}000 fully structured messages (folders, headers, bodies, labels, attachment filenames). Agents interact through \textbf{read-only JSON/CSV tools} rooted at the workspace working directory (e.g.\ \envtool{emails\_export.json}, \envtool{folders.json}, \envtool{mailbox\_stats.json}, \envtool{attachments/}, \envtool{enron\_spam\_data/}) rather than a live IMAP surface; the reference ``emails'' MCP catalog (25 tools such as \texttt{send\_email}, \texttt{move\_email}) is a conceptual superset, but the bundled verifier assumes the \textbf{local file toolset} below.

\section*{File summary}

\begin{envvisinner}[Directory layout (repository root)]
\begin{lstlisting}[style=envvispycode, language={}, morekeywords={}]
./
  emails_export.json         -- JSON array: 2000 synthetic emails (email_id, folder, from, to, cc, bcc,
                               --   subject, body, date, spam_ham, attachments[])
  folders.json               -- JSON array of folder names present in the export
  mailbox_stats.json         -- JSON: total_emails + per-folder counts
  enron_spam_data/
    enron_spam_data.csv      -- ~814k-row Enron Spam/Ham CSV (Message ID, Subject, Message, Spam/Ham, Date)
  attachments/
    attachment1.txt          -- Plain-text sample attachment for download tests
    attachment2.txt          -- Plain-text sample attachment for download tests
\end{lstlisting}
\end{envvisinner}

\medskip
\noindent\textbf{Data roles}
\begin{itemize}[leftmargin=1.2em, itemsep=4pt]
  \item \textbf{\envtool{emails\_export.json}} --- Primary corpus for listing, lookup, search, and spam/ham aggregation; each record mirrors a realistic MIME-like payload without requiring a network.
  \item \textbf{\envtool{folders.json}} / \textbf{\envtool{mailbox\_stats.json}} --- Lightweight indices so agents can validate folder vocabulary and global counts before expensive scans.
  \item \textbf{\envtool{enron\_spam\_data/enron\_spam\_data.csv}} --- Source-scale file for sampling and distributional questions; implementations should avoid loading the entire CSV into memory.
  \item \textbf{\envtool{attachments/*.txt}} --- On-disk payloads referenced when an email lists an attachment name; \envtool{get\_attachment\_content} joins export metadata with these files.
\end{itemize}

\section*{Tool list}

\noindent\textbf{Tools (each paired with intent)}
\begin{itemize}[leftmargin=1.2em, itemsep=6pt]
  \item \textbf{\envtool{list\_emails}} --- Paginated slice over \envtool{emails\_export.json} with optional \texttt{folder} filter; returns \texttt{total}, \texttt{page}, \texttt{page\_size}, and \texttt{emails}.
  \item \textbf{\envtool{get\_email\_by\_id}} --- Returns a single message dict by \texttt{email\_id}, or a structured \texttt{error} object if missing.
  \item \textbf{\envtool{get\_folders}} --- Reads \envtool{folders.json} and returns the folder name array (or an \texttt{error} wrapper).
  \item \textbf{\envtool{get\_mailbox\_stats}} --- Loads \envtool{mailbox\_stats.json}; optional \texttt{folder\_name} narrows to one bucket, otherwise returns the full summary object.
  \item \textbf{\envtool{search\_emails}} --- Case-insensitive substring match over \texttt{subject}, \texttt{body}, \texttt{from}, and \texttt{to} (list-aware), with optional folder scoping and pagination matching \envtool{list\_emails}.
  \item \textbf{\envtool{get\_attachment\_content}} --- Verifies the filename appears in the email's \texttt{attachments} list, then reads \envtool{attachments/<attachment\_filename>} as UTF-8 text.
  \item \textbf{\envtool{sample\_enron\_spam}} --- Reservoir-style sampling of up to \texttt{n} rows from the Enron CSV with optional \texttt{spam\_ham} $\in$ \{\texttt{spam}, \texttt{ham}, \texttt{both}\}.
  \item \textbf{\envtool{get\_email\_counts\_by\_folder\_and\_spam}} --- Derives per-folder spam/ham counts directly from the synthetic export for charting and consistency checks.
\end{itemize}

\section*{Tool example}

Reference implementation for \envtool{list\_emails} (paths resolved with \texttt{os.getcwd()}, matching the tool name).

\begin{lstlisting}[style=envvispycode]
def list_emails(folder=None, page=1, page_size=20):
    import json, os
    path = os.path.join(os.getcwd(), 'emails_export.json')
    try:
        with open(path, 'r', encoding='utf-8') as f:
            emails = json.load(f)
    except Exception as e:
        return {"error": f"Failed to load emails: {str(e)}"}
    if folder:
        emails = [e for e in emails if e.get('folder') == folder]
    total = len(emails)
    start = (page - 1) * page_size
    end = start + page_size
    paginated = emails[start:end]
    return {"total": total, "page": page, "page_size": page_size, "emails": paginated}
\end{lstlisting}

\end{envviscapsule}

\clearpage
\subsection{\texttt{Calendar}}

\begin{envviscapsule}[\faCalendar\; Environment: \texttt{Calendar}]

\section*{Environment}

This environment backs a \textbf{synthetic Calendar service} used in evaluation: all state lives under \envtool{/app/database/} as JSON plus a short \envtool{README.md}. \textbf{Users} (\envtool{users.json}) supply stable UUIDs, corporate-style emails, display names, and IANA time zones; \textbf{holidays} (\envtool{holidays.json}) enumerate U.S.\ all-day markers for 2025 with federal/regional typing; \textbf{events} (\envtool{events.json}) hold 34 timed meetings from January--June~2025 (all-hands, kickoff, eight weekly syncs, twenty one-on-ones, four client reviews) with UTC \texttt{Z} timestamps, attendee email lists drawn from the user roster, and string locations (rooms or Zoom/Teams/Meet). A richer production calendar MCP might expose \texttt{calendar\_create\_event}, \texttt{calendar\_get\_event}, \emph{etc.}; here the \textbf{read/query file toolset} below is what agents and verifiers assume.

\section*{File summary}

\begin{envvisinner}[Directory layout (\envtool{/app/database/})]
\begin{lstlisting}[style=envvispycode, language={}, morekeywords={}]
/app/database/
  README.md          -- Markdown: purpose of the directory + schema overview for the three JSON files
  users.json         -- JSON array: id (UUID), email, name, timezone (IANA string)
  holidays.json      -- JSON array: event_id, title, start_time/end_time (UTC all-day window),
                       --   attendees [], location null, type (Federal / Regional / Not a Public Holiday, ...)
  events.json        -- JSON array: event_id, title, start_time, end_time (UTC), attendees[], location
\end{lstlisting}
\end{envvisinner}

\medskip
\noindent\textbf{Data roles}
\begin{itemize}[leftmargin=1.2em, itemsep=4pt]
  \item \textbf{\envtool{users.json}} --- Canonical directory for resolving attendee emails to people and time zones; also supports standalone user lookups.
  \item \textbf{\envtool{holidays.json}} --- Public-holiday layer with empty attendee lists and \texttt{null} locations; range filters use the calendar date of \texttt{start\_time}.
  \item \textbf{\envtool{events.json}} --- Primary workload for scheduling queries: overlaps, per-user agendas, and title search all scan this array.
  \item \textbf{\envtool{README.md}} --- Human-readable contract tying on-disk layout to the intended calendar tool surface for developers.
\end{itemize}

\section*{Tool list}

\noindent\textbf{Tools (each paired with intent)}
\begin{itemize}[leftmargin=1.2em, itemsep=6pt]
  \item \textbf{\envtool{list\_all\_users}} --- Loads and returns the full \envtool{users.json} array.
  \item \textbf{\envtool{get\_user\_by\_email}} --- Case-insensitive email match; returns one user dict or \texttt{None}.
  \item \textbf{\envtool{get\_user\_by\_id}} --- UUID equality on \texttt{id}; returns one user dict or \texttt{None}.
  \item \textbf{\envtool{list\_all\_holidays}} --- Returns every holiday record from \envtool{holidays.json}.
  \item \textbf{\envtool{list\_holidays\_in\_range}} --- Inclusive \texttt{YYYY-MM-DD} window on the holiday's start date (UTC-derived calendar day).
  \item \textbf{\envtool{get\_holiday\_by\_id}} --- Lookup by \texttt{event\_id} in \envtool{holidays.json}.
  \item \textbf{\envtool{list\_all\_events}} --- Returns the full \envtool{events.json} array (all 34 events).
  \item \textbf{\envtool{list\_events\_in\_range}} --- Inclusive date filter on each event's \texttt{start\_time} calendar day in UTC.
  \item \textbf{\envtool{get\_event\_by\_id}} --- Lookup by \texttt{event\_id} in \envtool{events.json}.
  \item \textbf{\envtool{list\_user\_events\_on\_date}} --- Filters events where \texttt{user\_email} appears in \texttt{attendees} (case-insensitive) and the start date equals the given day.
  \item \textbf{\envtool{search\_events\_by\_title}} --- Case-insensitive substring filter over \texttt{title} only.
\end{itemize}

\section*{Tool example}

Reference implementation for \envtool{list\_events\_in\_range} (fixed \envtool{/app/database/events.json} path, matching the tool name).

\begin{lstlisting}[style=envvispycode]
def list_events_in_range(start_date, end_date):
    import json, pathlib, datetime
    start = datetime.date.fromisoformat(start_date)
    end = datetime.date.fromisoformat(end_date)
    path = pathlib.Path('/app/database/events.json')
    with path.open() as f:
        events = json.load(f)
    result = []
    for e in events:
        e_start = datetime.datetime.fromisoformat(
            e['start_time'].replace('Z', '+00:00')
        ).date()
        if start <= e_start <= end:
            result.append(e)
    return result
\end{lstlisting}

\end{envviscapsule}

\clearpage
\subsection{\texttt{Hotels}}

\begin{envviscapsule}[\faMapMarker\; Environment: \texttt{Hotels}]

\section*{Environment}

This environment backs a \textbf{Hotels service} used in evaluation: structured inventory and booking state live under \envtool{/app/database/}. A compact \textbf{Hangzhou-focused trio} in \envtool{hotels.json} / \envtool{hotel\_details.json} supports end-to-end drills (geocode $\rightarrow$ search $\rightarrow$ room details $\rightarrow$ book), while \textbf{\envtool{rates.json}} maps many numeric supplier hotel IDs to synthetic Standard/Deluxe/Suite rate rows, \textbf{\envtool{hotels\_sample.csv}} carries wide scraped-style listing rows (bilingual names, scores, embedded JSON facility blobs, media URLs), and \textbf{\envtool{world\_cities.csv}} supplies a global city/lat/lng/population reference. Production systems might expose dozens of OTA tools; here agents assume the \textbf{six-tool surface} below, with \envtool{book\_hotel} as the sole writer of persistent booking records.

\section*{File summary}

\begin{envvisinner}[Directory layout (\envtool{/app/database/})]
\begin{lstlisting}[style=envvispycode, language={}, morekeywords={}]
/app/database/
  bookings.json        -- JSON array (initially empty): booking_id, hotel_id, rate_id,
  --                     user_id, payment_link, status; appended by book_hotel
  facilities.json      -- JSON array: id (e.g. FREE_WIFI), name, description
  hotel_details.json   -- JSON array: hotel_id, description, rooms[] (rate_id, name,
  --                     price, currency, amenities[], cancellation_policy)
  hotels.json          -- JSON array: hotel_id, name, address, city, province,
  --                     star_rating, lat/lng, price_per_night, rating, facilities[]
  hotels_sample.csv    -- Wide CSV: many columns (names, geo, scores, nested JSON text, URLs)
  places.json          -- JSON array: query, name, latitude, longitude, country
  rates.json           -- JSON object: hotel_id (string key) -> array of rate rows
  world_cities.csv     -- CSV: city, lat, lng, country, population
\end{lstlisting}
\end{envvisinner}

\medskip
\noindent\textbf{Data roles}
\begin{itemize}[leftmargin=1.2em, itemsep=4pt]
  \item \textbf{\envtool{places.json}} --- Normalizes free-text locations (\texttt{Hangzhou}, \texttt{Paris}, \ldots) to coordinates for \envtool{find\_place} before \envtool{search\_hotels}.
  \item \textbf{\envtool{facilities.json}} --- Canonical amenity vocabulary: internal IDs (\texttt{FREE\_WIFI}, \texttt{GYM}, \ldots) with English labels used for search filters and room metadata.
  \item \textbf{\envtool{hotels.json}} --- Paged search index: summary rows with star rating, optional list price and guest score, and facility ID lists aligned with \envtool{facilities.json}.
  \item \textbf{\envtool{hotel\_details.json}} --- Per-hotel narrative blurb plus priced \texttt{rooms} with \texttt{rate\_id} values that \envtool{book\_hotel} must echo into the ledger.
  \item \textbf{\envtool{rates.json}} / \textbf{\envtool{hotels\_sample.csv}} / \textbf{\envtool{world\_cities.csv}} --- Heavier reference slices for tasks that join bulk pricing, listing exports, or city-level geography against the small curated hotel set.
  \item \textbf{\envtool{bookings.json}} --- Append-only booking store created at cold start as \texttt{[]}; each successful reservation adds an object with stable identifiers and a payment hand-off field.
\end{itemize}

\section*{Tool list}

\noindent\textbf{Tools (each paired with intent)}
\begin{itemize}[leftmargin=1.2em, itemsep=6pt]
  \item \textbf{\envtool{find\_place}} --- Maps a user query string plus optional language hint to a normalized place name and coordinates using \envtool{places.json}.
  \item \textbf{\envtool{search\_hotels}} --- Location-driven availability search: accepts latitude/longitude, optional name hint, stay dates (defaults \texttt{2025-06-25} / \texttt{2025-06-26}), and party size; returns a session handle for follow-on calls.
  \item \textbf{\envtool{load\_more\_hotels}} --- Pagination helper that continues a prior \envtool{search\_hotels} (or self) result stream given \texttt{session\_id}.
  \item \textbf{\envtool{get\_hotel\_details}} --- Requires \texttt{session\_id} and \texttt{hotel\_id}; returns long-form copy and room/rate rows from \envtool{hotel\_details.json} (or an equivalent merge with cached search context).
  \item \textbf{\envtool{book\_hotel}} --- Starts checkout for a chosen \texttt{rate\_id} under \texttt{hotel\_id}; persists a booking record into \envtool{bookings.json}.
  \item \textbf{\envtool{get\_facilities}} --- Returns the full facility dictionary for display or filter construction (optional \texttt{language} argument; evaluation fixtures are English-centric).
\end{itemize}

\section*{Tool example}

Reference implementation for \envtool{get\_facilities} (fixed \envtool{/app/database/facilities.json} path, matching the tool name).

\begin{lstlisting}[style=envvispycode]
def get_facilities(language='en'):
    import json, pathlib
    path = pathlib.Path('/app/database/facilities.json')
    with path.open(encoding='utf-8') as f:
        facilities = json.load(f)
    return {'language': language, 'facilities': facilities}
\end{lstlisting}

\end{envviscapsule}

\clearpage
\subsection{\texttt{App\_stores}}

\begin{envviscapsule}[\faGlobe\; Environment: \texttt{App\_stores}]

\section*{Environment}

This environment packages a \textbf{read-only, file-backed mirror} of public storefront metadata for three flagship social apps (Instagram, TikTok, WhatsApp): one \textbf{App Store} slice (search summary row plus full \texttt{app-store-details} style objects and recent Apple reviews per app) and one \textbf{Google Play} slice (details with embedded score/installs/summary, synthetic review threads, structured Data Safety objects, and permission string lists). All payloads live as static JSON under \envtool{/app/database/}; agents never call live App Store or Play APIs. Tools are thin \texttt{json.load} wrappers with \textbf{fixed absolute paths}, so tasks can require deterministic cross-platform joins (ratings, dates, permission overlap, data-sharing posture) without network variance.

\section*{File summary}

\begin{envvisinner}[Directory layout (\envtool{/app/database/})]
\begin{lstlisting}[style=envvispycode, language={}, morekeywords={}]
/app/database/
  app_store_search_results.json          -- Array: one row each for Instagram, TikTok, WhatsApp (id, appId, title, icon, url, price, currency, free, description, developer, ...)
  app_store_details_instagram.json       -- Full App Store metadata object for Instagram (devices, screenshots, ratings, version, advisories, ...)
  app_store_details_tiktok.json          -- Full App Store metadata object for TikTok
  app_store_details_whatsapp.json        -- Full App Store metadata object for WhatsApp
  app_store_reviews_instagram.json       -- Array of recent Apple reviews (id, userName, rating, title, text, date)
  app_store_reviews_tiktok.json          -- Array of recent Apple reviews for TikTok
  google_play_details_instagram.json     -- Play listing dict (title, appId, url, developer, score, installs, updated, dataSafety summary text, permissions[])
  google_play_details_tiktok.json        -- Play listing dict for TikTok
  google_play_details_whatsapp.json      -- Play listing dict for WhatsApp
  google_play_reviews_instagram.json      -- Array of Play-style reviews (id, userName, rating, title, text, date, version)
  google_play_reviews_tiktok.json
  google_play_reviews_whatsapp.json
  google_play_datasafety_instagram.json    -- Object: dataShared[], dataCollected[], securityPractices[], deletionRequest
  google_play_datasafety_tiktok.json
  google_play_datasafety_whatsapp.json
  google_play_permissions_instagram.json   -- JSON array of human-readable Android permission strings
  google_play_permissions_tiktok.json
  google_play_permissions_whatsapp.json
\end{lstlisting}
\end{envvisinner}

\medskip
\noindent\textbf{Data roles}
\begin{itemize}[leftmargin=1.2em, itemsep=4pt]
  \item \textbf{\envtool{app\_store\_search\_results.json}} --- Compact index of the three apps for list/compare tasks before opening heavy detail blobs.
  \item \textbf{\envtool{app\_store\_details\_*.json}} --- Canonical iOS-side fields (version strings, content advisories, language lists, rating counts) for rubrics that reference Apple-only semantics.
  \item \textbf{\envtool{app\_store\_reviews\_instagram.json}} / \textbf{\envtool{app\_store\_reviews\_tiktok.json}} --- Short recent-review windows for sentiment or moderation-themed queries on the Apple side (no separate WhatsApp App Store review file in this snapshot).
  \item \textbf{\envtool{google\_play\_details\_*.json}} --- Android-side headline metrics plus inline \texttt{dataSafety} prose and \texttt{permissions} arrays suitable for permission-set reasoning.
  \item \textbf{\envtool{google\_play\_datasafety\_*.json}} --- Structured sharing/collection/security labels that pair with the narrative \texttt{dataSafety} field in details when tasks require consistency checks.
  \item \textbf{\envtool{google\_play\_reviews\_*.json}} / \textbf{\envtool{google\_play\_permissions\_*.json}} --- Sidecar corpora for cross-store contrast (e.g., average rating vs.\ Play score, or permission deltas across messengers).
\end{itemize}

\section*{Tool list}

\noindent\textbf{Tools (each paired with intent)}
\begin{itemize}[leftmargin=1.2em, itemsep=6pt]
  \item \textbf{\envtool{get\_app\_store\_search\_results}} --- Returns the full search array (three apps) from \envtool{app\_store\_search\_results.json}.
  \item \textbf{\envtool{get\_app\_store\_details\_instagram}} --- Load full App Store metadata for Instagram.
  \item \textbf{\envtool{get\_app\_store\_details\_tiktok}} --- Load full App Store metadata for TikTok.
  \item \textbf{\envtool{get\_app\_store\_details\_whatsapp}} --- Load full App Store metadata for WhatsApp.
  \item \textbf{\envtool{get\_app\_store\_reviews\_instagram}} --- Load Apple-side reviews for Instagram.
  \item \textbf{\envtool{get\_app\_store\_reviews\_tiktok}} --- Load Apple-side reviews for TikTok.
  \item \textbf{\envtool{get\_google\_play\_details\_instagram}} --- Load Google Play listing dict for Instagram (scores, installs, summary, inline data-safety text, permissions).
  \item \textbf{\envtool{get\_google\_play\_details\_tiktok}} --- Load Google Play listing dict for TikTok.
  \item \textbf{\envtool{get\_google\_play\_details\_whatsapp}} --- Load Google Play listing dict for WhatsApp.
  \item \textbf{\envtool{get\_google\_play\_reviews\_instagram}} --- Load Play review array for Instagram.
  \item \textbf{\envtool{get\_google\_play\_reviews\_tiktok}} --- Load Play review array for TikTok.
  \item \textbf{\envtool{get\_google\_play\_reviews\_whatsapp}} --- Load Play review array for WhatsApp.
  \item \textbf{\envtool{get\_google\_play\_datasafety\_instagram}} --- Load structured Data Safety object for Instagram.
  \item \textbf{\envtool{get\_google\_play\_datasafety\_tiktok}} --- Load structured Data Safety object for TikTok.
  \item \textbf{\envtool{get\_google\_play\_datasafety\_whatsapp}} --- Load structured Data Safety object for WhatsApp.
  \item \textbf{\envtool{get\_google\_play\_permissions\_instagram}} --- Load Android permission strings for Instagram.
  \item \textbf{\envtool{get\_google\_play\_permissions\_tiktok}} --- Load Android permission strings for TikTok.
  \item \textbf{\envtool{get\_google\_play\_permissions\_whatsapp}} --- Load Android permission strings for WhatsApp.
\end{itemize}

\section*{Tool example}

Reference implementation for \envtool{get\_app\_store\_search\_results} (fixed path under \envtool{/app/database/}, matching the exposed tool name).

\begin{lstlisting}[style=envvispycode]
def get_app_store_search_results():
    '''Load App Store search results from JSON file.'''
    import json
    path = '/app/database/app_store_search_results.json'
    with open(path, 'r', encoding='utf-8') as f:
        return json.load(f)
\end{lstlisting}

\end{envviscapsule}

\clearpage
\subsection{\texttt{Food\_delivery}}

\begin{envviscapsule}[\faCoffee\; Environment: \texttt{Food\_delivery}]

\section*{Environment}

This environment simulates a \textbf{multi-city food-delivery storefront} backed entirely by JSON under \envtool{/app/database/}: users, restaurants (\texttt{stores}), menu items (\texttt{products}), order histories with delivery timing and status, a pre-materialized \textbf{user--store distance} table (Haversine meters), a \textbf{2025 China public-holiday} calendar, a compact \textbf{early-January~2025 weather} panel for five major cities, and per-user \textbf{behavior aggregates} (order counts and average spend on paid/delivered orders). Prices are in \textbf{RMB}. Tools are read-only queries and scans over these fixtures; there is no live ordering API.

\section*{File summary}

\begin{envvisinner}[Directory layout (\envtool{/app/database/})]
\begin{lstlisting}[style=envvispycode, language={}, morekeywords={}]
/app/database/
  users.json              -- user_id, name, address, lat/lon, phone, dietary_restrictions[], favorite_tags[], order_history[]
  stores.json             -- store_id, name, address, lat/lon, rating, tags[], product_ids[] (catalog pointers)
  products.json           -- product_id, name, price (RMB), rating, tags[], store_id
  orders.json             -- order_id, user_id, store_id, products[{product_id, quantity}], total_price, address, timestamps, distance_m, estimated_delivery_minutes, status, update_time
  distances.json          -- user_id, store_id, distance_m (precomputed for every user x store pair)
  holidays_2025.json      -- date (YYYY-MM-DD), name (Chinese public holidays for 2025)
  weather_2025.json       -- city, date, temperature_c, condition (Beijing, Shanghai, Guangzhou, Chengdu, Shenzhen; first week of Jan 2025)
  user_behaviors.json     -- user_id, home_address, dietary_restrictions[], favorite_tags[], order_count, average_spend
\end{lstlisting}
\end{envvisinner}

\medskip
\noindent\textbf{Data roles}
\begin{itemize}[leftmargin=1.2em, itemsep=4pt]
  \item \textbf{\envtool{users.json}} / \textbf{\envtool{stores.json}} / \textbf{\envtool{products.json}} --- Core dimensional tables for personalization, catalog filtering, and price/rating joins.
  \item \textbf{\envtool{orders.json}} --- Transactional facts for fulfillment timelines, status filters, and basket reconstruction.
  \item \textbf{\envtool{distances.json}} --- O(1)-style lookup for routing or SLA tasks without recomputing geodesics at runtime.
  \item \textbf{\envtool{holidays\_2025.json}} / \textbf{\envtool{weather\_2025.json}} --- Calendar and meteorology side channels for cross-domain prompts (e.g., holiday surcharges, snow-day delays).
  \item \textbf{\envtool{user\_behaviors.json}} --- Denormalized rollups aligned with user profiles for cohort-style questions without re-scanning all orders.
\end{itemize}

\section*{Tool list}

\noindent\textbf{Tools (each paired with intent)}
\begin{itemize}[leftmargin=1.2em, itemsep=6pt]
  \item \textbf{\envtool{get\_user\_by\_id}} --- Return one user dict by \texttt{user\_id}, or \texttt{None}.
  \item \textbf{\envtool{list\_users}} --- Return the full \envtool{users.json} array.
  \item \textbf{\envtool{get\_store\_by\_id}} --- Return one store dict by \texttt{store\_id}, or \texttt{None}.
  \item \textbf{\envtool{list\_stores}} --- Return the full \envtool{stores.json} array.
  \item \textbf{\envtool{get\_product\_by\_id}} --- Return one product dict by \texttt{product\_id}, or \texttt{None}.
  \item \textbf{\envtool{list\_products}} --- Filter \envtool{products.json} by optional \texttt{store\_id} and/or \texttt{tag} membership.
  \item \textbf{\envtool{get\_order\_by\_id}} --- Return one order dict by \texttt{order\_id}, or \texttt{None}.
  \item \textbf{\envtool{list\_user\_orders}} --- List orders for a \texttt{user\_id}, optionally restricted by \texttt{status} (\texttt{delivered}, \texttt{paid}, \texttt{unpaid}, \texttt{cancelled}, \ldots).
  \item \textbf{\envtool{get\_distance}} --- Return precomputed \texttt{distance\_m} for a \texttt{(user\_id, store\_id)} pair from \envtool{distances.json}.
  \item \textbf{\envtool{get\_holiday\_by\_date}} --- Return holiday name for a \texttt{YYYY-MM-DD} key in \envtool{holidays\_2025.json}, else \texttt{None}.
  \item \textbf{\envtool{get\_weather}} --- Return \texttt{\{temperature\_c, condition\}} for a \texttt{(city, date)} row in \envtool{weather\_2025.json}, else \texttt{None}.
  \item \textbf{\envtool{get\_user\_behavior}} --- Return the aggregated behavior row for \texttt{user\_id} from \envtool{user\_behaviors.json}, or \texttt{None}.
  \item \textbf{\envtool{search\_products\_by\_tag}} --- Return all products whose \texttt{tags} list contains the query tag.
  \item \textbf{\envtool{search\_stores\_by\_tag}} --- Return all stores whose \texttt{tags} list contains the query tag.
\end{itemize}

\section*{Tool example}

Reference implementation for \envtool{list\_users} (paths built with \texttt{os.path.join} under \envtool{/app/database/}).

\begin{lstlisting}[style=envvispycode]
def list_users():
    import json, os
    path = os.path.join('/app/database', 'users.json')
    with open(path, 'r', encoding='utf-8') as f:
        return json.load(f)
\end{lstlisting}

\end{envviscapsule}

\endgroup

\section{Case Study}\label{sec:case-study}

We present three representative trajectories with at least 7 interaction turns. For each case, we report the environment context, task requirement, tool inventory, rubric criteria, and an abbreviated interaction trajectory.

\begingroup
\fontfamily{ptm}\selectfont
\small

\clearpage
\begin{figure*}[p]
\subsection{Case 1: Ecomm MCP Server}
\begin{tcolorbox}[
    enhanced,
    colback=CaseGreenBg,
    colframe=CaseGreen,
    coltitle=white,
    fonttitle=\bfseries,
    attach boxed title to top left={yshift=-3mm, xshift=5mm},
    boxed title style={colback=CaseGreen, sharp corners},
    sharp corners=south, rounded corners=north,
    boxrule=0.8pt,
    top=4mm, left=4mm, right=4mm, bottom=3mm,
    breakable,
    title={Case 1: Ecomm MCP Server — Multi-Step Return Execution}
]

\textbf{Environment:} \texttt{ecomm\_mcp\_server} \hfill \textbf{Tools Used:} 4/17 \hfill \textbf{Steps:} 9

\smallskip
\textbf{Task Requirement.}\;
\textit{Handle a customer return request by completing identity verification, enumerating historical orders, identifying eligible delivered orders, collecting item-level return intent, confirming refund destination, and submitting the return action under policy constraints.}

\smallskip
\begin{tabularx}{\textwidth}{p{0.32\textwidth} >{\raggedright\arraybackslash}X}
\toprule
\textbf{Tool} & \textbf{Description} \\
\midrule
\texttt{find\_user\_id\_by\_name\_zip} & Resolve customer identifier from profile metadata \\
\texttt{get\_user\_details} & Retrieve account, payment method, and order identifiers \\
\texttt{get\_order\_details} & Retrieve per-order status and item-level fulfillment details \\
\texttt{return\_delivered\_order\_items} & Submit return request for selected delivered items \\
\bottomrule
\end{tabularx}

\smallskip
\textbf{Rubrics.}\;
\textit{C1}: Returned items must belong to a delivered order.\;
\textit{C2}: Refund method must follow user-confirmed payment path.\;
\textit{C3}: Order state transition must become \texttt{return requested}.\;
\textit{C4}: Selected returned item IDs must match user intent exactly.

\smallskip
\textbf{Trajectory.}

\begin{longtable}{@{}p{0.03\textwidth} >{\raggedright\arraybackslash}p{0.92\textwidth}@{}}
\toprule
\textbf{Turn} & \textbf{Interaction} \\
\midrule
1 & \textcolor{RoleUser}{\textbf{[User]}}\; Requests help returning multiple recent products (\textit{content abbreviated}). \\
\midrule
2 & \textcolor{RoleAssistant}{\textbf{[Assistant]}}\; Identity verification requested and resolved via name+zip \newline
\textcolor{ActionBlue}{\texttt{find\_user\_id\_by\_name\_zip(first\_name="Anya", last\_name="Sanchez", zip="43171")}} \\
\midrule
3 & \textcolor{RoleTool}{\textbf{[Tool]}}\; \texttt{anya\_sanchez\_9707} \newline
\textcolor{RoleAssistant}{\textbf{[Assistant]}}\; \textcolor{ActionBlue}{\texttt{get\_user\_details(user\_id="anya\_sanchez\_9707")}} \\
\midrule
4--6 & \textcolor{RoleAssistant}{\textbf{[Assistant]}}\; Iteratively checks four orders for eligibility \newline
\textcolor{ActionBlue}{\texttt{get\_order\_details(order\_id="\#W5402785")}} \newline
\textcolor{ActionBlue}{\texttt{get\_order\_details(order\_id="\#W2136962")}} \newline
\textcolor{ActionBlue}{\texttt{get\_order\_details(order\_id="\#W4442043")}} \newline
\textcolor{ActionBlue}{\texttt{get\_order\_details(order\_id="\#W6002958")}} \\
\midrule
7 & \textcolor{RoleAssistant}{\textbf{[Assistant]}}\; Presents eligible delivered order \texttt{\#W4442043}, collects item selection and refund confirmation (\textit{content abbreviated}). \\
\midrule
8 & \textcolor{RoleAssistant}{\textbf{[Assistant]}}\; Executes return submission \newline
\textcolor{ActionBlue}{\texttt{return\_delivered\_order\_items(order\_id="\#W4442043", item\_ids=[6697922351,7758198585,3187628796], payment\_method\_id="paypal\_1191071")}} \\
\midrule
9 & \textcolor{RoleTool}{\textbf{[Tool]}}\; \texttt{\{status: "return requested", return\_items: [3187628796, 6697922351, 7758198585]\}} \newline
\textcolor{RoleAssistant}{\textbf{[Assistant]}}\; Confirms completed return and refund processing window. \\
\bottomrule
\end{longtable}
\end{tcolorbox}
\vspace{-0.3cm}
\caption{Case 1: The agent executes a complete e-commerce return transaction under policy constraints with iterative order-state verification and user-confirmed actuation.}
\label{fig:case_study_1}
\end{figure*}


\begin{figure*}[p]
\subsection{Case 2: Slack Workspace Automation Server}
\begin{tcolorbox}[
    enhanced,
    colback=CaseBlueBg,
    colframe=CaseBlue,
    coltitle=white,
    fonttitle=\bfseries,
    attach boxed title to top left={yshift=-3mm, xshift=5mm},
    boxed title style={colback=CaseBlue, sharp corners},
    sharp corners=south, rounded corners=north,
    boxrule=0.8pt,
    top=4mm, left=4mm, right=4mm, bottom=3mm,
    breakable,
    title={Case 2: Slack Workspace Automation Server — Compliance Triage}
]

\textbf{Environment:} \texttt{slack\_workspace\_automation\_server} \hfill \textbf{Tools Used:} 5/18 \hfill \textbf{Steps:} 7

\smallskip
\textbf{Task Requirement.}\;
\textit{Conduct Slack compliance triage for a reported message by identifying sender attributes, computing text similarity and temporal proximity against a target user's CSV history, checking login state and contact graph dependency, and generating a deterministic triage comment payload for issue synchronization.}

\smallskip
\begin{tabularx}{\textwidth}{p{0.30\textwidth} >{\raggedright\arraybackslash}X}
\toprule
\textbf{Tool} & \textbf{Description} \\
\midrule
\texttt{get\_message\_by\_ts} & Retrieve reported message by timestamp \\
\texttt{get\_user\_by\_id} & Retrieve sender profile / existence signal \\
\texttt{filter\_message\_csv\_by\_user} & Retrieve historical CSV utterances for comparison target \\
\texttt{get\_login\_status} & Retrieve login state for workflow dependency user \\
\texttt{list\_contacts} & Retrieve contact graph to determine add-contact necessity \\
\bottomrule
\end{tabularx}

\smallskip
\textbf{Rubrics.}\;
\textit{C1}: Required JSON fields and value types must be correct.\;
\textit{C2}: Similarity and time-difference computations must be reproducible.\;
\textit{C3}: \texttt{need\_add\_contact} must reflect login + contact-state logic.\;
\textit{C4}: Comment body must include branch name, score, time gap, level, and contact conclusion.

\smallskip
\textbf{Trajectory.}

\begin{longtable}{@{}p{0.03\textwidth} >{\raggedright\arraybackslash}p{0.92\textwidth}@{}}
\toprule
\textbf{Turn} & \textbf{Interaction} \\
\midrule
1 & \textcolor{RoleUser}{\textbf{[User]}}\; Requests compliance triage JSON for reported Slack message with fixed output schema (\textit{content abbreviated}). \\
\midrule
2 & \textcolor{RoleAssistant}{\textbf{[Assistant]}}\; \textcolor{ActionBlue}{\texttt{get\_message\_by\_ts(ts="1512104434.000490")}} \\
\midrule
3 & \textcolor{RoleTool}{\textbf{[Tool]}}\; \texttt{\{user: "U222BBB222", text: "What, you want to smell my shoes better?"\}} \newline
\textcolor{RoleAssistant}{\textbf{[Assistant]}}\; \textcolor{ActionBlue}{\texttt{get\_user\_by\_id(user\_id="U222BBB222")}} \\
\midrule
4 & \textcolor{RoleTool}{\textbf{[Tool]}}\; \texttt{\{status: "not\_found"\}} \newline
\textcolor{RoleAssistant}{\textbf{[Assistant]}}\; \textcolor{ActionBlue}{\texttt{filter\_message\_csv\_by\_user(user\_id="U123ABC456")}} \\
\midrule
5 & \textcolor{RoleTool}{\textbf{[Tool]}}\; Returns one CSV utterance at \texttt{1512085950.000216} \newline
\textcolor{RoleAssistant}{\textbf{[Assistant]}}\; \textcolor{ActionBlue}{\texttt{get\_login\_status(user\_id="W012A3CDE")}} \\
\midrule
6 & \textcolor{RoleTool}{\textbf{[Tool]}}\; \texttt{\{is\_logged\_in: true\}} \newline
\textcolor{RoleAssistant}{\textbf{[Assistant]}}\; \textcolor{ActionBlue}{\texttt{list\_contacts()}} \\
\midrule
7 & \textcolor{RoleTool}{\textbf{[Tool]}}\; Contact \texttt{W07QCRPA4} already exists \newline
\textcolor{RoleAssistant}{\textbf{[Assistant]}}\; Final JSON: \texttt{similarity\_score = 0.18}, \texttt{time\_diff\_seconds = 18484}, \texttt{need\_add\_contact = false}, triage level \texttt{ignore}. \\
\bottomrule
\end{longtable}
\end{tcolorbox}
\vspace{-0.3cm}
\caption{Case 2: The agent combines message forensics, lexical similarity scoring, temporal reasoning, and contact-graph constraints into a deterministic Slack triage decision.}
\label{fig:case_study_2}
\end{figure*}


\begin{figure*}[p]
\subsection{Case 3: Population Data Server}
\begin{tcolorbox}[
    enhanced,
    colback=CaseGreenBg,
    colframe=CaseGreen,
    coltitle=white,
    fonttitle=\bfseries,
    attach boxed title to top left={yshift=-3mm, xshift=5mm},
    boxed title style={colback=CaseGreen, sharp corners},
    sharp corners=south, rounded corners=north,
    boxrule=0.8pt,
    top=4mm, left=4mm, right=4mm, bottom=3mm,
    breakable,
    title={Case 3: Population Data Server — Quantitative Ranking Under Tool Constraints}
]

\textbf{Environment:} \texttt{population\_data\_server} \hfill \textbf{Tools Used:} 5/11 \hfill \textbf{Steps:} 10

\smallskip
\textbf{Task Requirement.}\;
\textit{Construct a city-influence ranking using top-10 population cities plus Karachi, apply growth-adjusted scoring for Karachi, and output winner plus transformed magnitude metric. The workflow must integrate sampled population data, category-specific percentage lookups, and logarithm constants under available tool precision limits.}

\smallskip
\begin{tabularx}{\textwidth}{p{0.30\textwidth} >{\raggedright\arraybackslash}X}
\toprule
\textbf{Tool} & \textbf{Description} \\
\midrule
\texttt{list\_top\_cities\_by\_population} & Retrieve ranked top-\(N\) city populations \\
\texttt{get\_city\_population} & Retrieve city-level population and growth metadata \\
\texttt{get\_city\_population\_percentage} & Retrieve global listed-population percentage for target city \\
\texttt{get\_logarithms\_for\_value} & Retrieve \(\log_{10}\), \(\log_2\), and \(\ln\) from lookup table \\
\texttt{get\_precision\_constant} & Retrieve high-precision constants (with fallback when unavailable) \\
\bottomrule
\end{tabularx}

\smallskip
\textbf{Rubrics.}\;
\textit{C1}: Candidate pool must include top-10 cities plus Karachi.\;
\textit{C2}: Scoring and Karachi growth adjustment must follow declared formula.\;
\textit{C3}: Tie-break logic must be deterministic and reproducible.\;
\textit{C4}: Output fields and numeric precision must match schema.

\smallskip
\textbf{Trajectory.}

\begin{longtable}{@{}p{0.03\textwidth} >{\raggedright\arraybackslash}p{0.92\textwidth}@{}}
\toprule
\textbf{Turn} & \textbf{Interaction} \\
\midrule
1 & \textcolor{RoleUser}{\textbf{[User]}}\; Requests winner-city computation with explicit logarithmic and tie-break rules (\textit{content abbreviated}). \\
\midrule
2 & \textcolor{RoleAssistant}{\textbf{[Assistant]}}\; \textcolor{ActionBlue}{\texttt{list\_top\_cities\_by\_population(limit=10)}} \\
\midrule
3 & \textcolor{RoleTool}{\textbf{[Tool]}}\; Returns top-10 list headed by Tokyo \newline
\textcolor{RoleAssistant}{\textbf{[Assistant]}}\; \textcolor{ActionBlue}{\texttt{get\_city\_population(city="Karachi", country="Pakistan")}} \\
\midrule
4 & \textcolor{RoleTool}{\textbf{[Tool]}}\; \texttt{\{Pop2023: 17236230, growthRate: 0.0235\}} \newline
\textcolor{RoleAssistant}{\textbf{[Assistant]}}\; \textcolor{ActionBlue}{\texttt{get\_city\_population\_percentage(city="Tokyo")}} \newline
\textcolor{ActionBlue}{\texttt{get\_city\_population\_percentage(city="Mumbai")}} \\
\midrule
5 & \textcolor{RoleTool}{\textbf{[Tool]}}\; Returns percentages for Tokyo and Mumbai \newline
\textcolor{RoleAssistant}{\textbf{[Assistant]}}\; \textcolor{ActionBlue}{\texttt{get\_logarithms\_for\_value(x=40)}} \newline
\textcolor{ActionBlue}{\texttt{get\_precision\_constant(constant\_name="log2\_10")}} \\
\midrule
6 & \textcolor{RoleTool}{\textbf{[Tool]}}\; \texttt{get\_precision\_constant} returns \texttt{None}; assistant triggers fallback via \texttt{get\_logarithms\_for\_value(x=10)} and additional checks (\textit{content abbreviated}). \\
\midrule
7--9 & \textcolor{RoleAssistant}{\textbf{[Assistant]}}\; \textcolor{ThinkGray}{\texttt{<think>} Resolves constant fallback, computes score ordering, and confirms Tokyo as winner after growth-adjusted Karachi comparison \ldots \texttt{</think>}} \\
\midrule
10 & \textcolor{RoleAssistant}{\textbf{[Assistant]}}\; Final JSON: \texttt{winner\_city = "Tokyo"}, \texttt{winner\_country = "Japan"}, \texttt{score\_S = 13.7423}, \texttt{magnitude\_M = 1.1384}. \\
\bottomrule
\end{longtable}
\end{tcolorbox}
\vspace{-0.3cm}
\caption{Case 3: The agent solves a constrained quantitative ranking problem with explicit fallback reasoning when preferred constants are unavailable.}
\label{fig:case_study_3}
\end{figure*}

\normalsize

\endgroup

\end{document}